\documentclass{article}

\usepackage[preprint]{icml2026}

\usepackage[utf8]{inputenc}
\usepackage[T1]{fontenc}
\usepackage{hyperref}
\usepackage{url}
\usepackage{booktabs}
\usepackage{array}
\usepackage{amsfonts, amsmath, amssymb}
\usepackage{nicefrac}
\usepackage{microtype}
\usepackage{xcolor}
\usepackage{graphicx}
\usepackage{tikz}
\usetikzlibrary{positioning, arrows.meta, calc}
\usepackage{caption}
\usepackage{enumitem}

\newcommand{\todo}[1]{{\color{red!70!black}\textbf{[TODO:} #1\textbf{]}}}
\newcommand{\physicl}{PhysICL}
\newcommand{\tficl}{\ensuremath{\mathrm{TF}_\mathrm{icl}}}
\newcommand{\tfcol}{\ensuremath{\mathrm{TF}_\mathrm{col}}}
\newcommand{\tfrow}{\ensuremath{\mathrm{TF}_\mathrm{row}}}
\newcommand{\R}{\mathbb{R}}

\newcommand{\journalIntegralUniform}{0.039}
\newcommand{\journalIntegralSobol}{0.007}

\IfFileExists{results/ahmed/plaid.tex}{\providecommand{\plaid}[3]{\ifcsname plaid@#1@#2@#3\endcsname\csname plaid@#1@#2@#3\endcsname\else\todo{#1 #2 #3}\fi}
\expandafter\def\csname plaid@AhmedML@8@physicl\endcsname{0.183}
\expandafter\def\csname plaid@AhmedML@8@nn\endcsname{0.601}
\expandafter\def\csname plaid@AhmedML@8@spread\endcsname{--}
\expandafter\def\csname plaid@AhmedML@8@spreadpct\endcsname{--}
\expandafter\def\csname plaid@AhmedML@8@draws\endcsname{1}
\expandafter\def\csname plaid@AhmedML@8@mean\endcsname{1.06}
\expandafter\def\csname plaid@AhmedML@8@best\endcsname{0.342}
\expandafter\def\csname plaid@AhmedML@8@bestname\endcsname{probe\_kumo[large,e8,v0.9999]}
\expandafter\def\csname plaid@AhmedML@8@floor\endcsname{0.181}
\expandafter\def\csname plaid@AhmedML@8@ratio\endcsname{0.536}
\expandafter\def\csname plaid@AhmedML@8@onfront\endcsname{yes}
\expandafter\def\csname plaid@AhmedML@8@onfronttotal\endcsname{yes}
\expandafter\def\csname plaid@AhmedML@8@onfrontqten\endcsname{yes}
\expandafter\def\csname plaid@AhmedML@8@onfrontqhundred\endcsname{yes}
\expandafter\def\csname plaid@AhmedML@8@beatsnn\endcsname{yes}
\expandafter\def\csname plaid@AhmedML@8@mostaccurate\endcsname{yes}
\expandafter\def\csname plaid@AhmedML@8@fit\endcsname{31.9}
\expandafter\def\csname plaid@AhmedML@8@query\endcsname{27.7}
\expandafter\def\csname plaid@AhmedML@8@bestfit\endcsname{74.6}
\expandafter\def\csname plaid@AhmedML@8@bestquery\endcsname{53.5}
\expandafter\def\csname plaid@AhmedML@8@coverage\endcsname{0.951}
\expandafter\def\csname plaid@AhmedML@8@scalars\endcsname{4}
\expandafter\def\csname plaid@AhmedML@8@covfmin\endcsname{0.925}
\expandafter\def\csname plaid@AhmedML@8@covfmax\endcsname{0.967}
\expandafter\def\csname plaid@AhmedML@8@r2\endcsname{0.136}
\expandafter\def\csname plaid@AhmedML@8@bestr2\endcsname{0.156}
\expandafter\def\csname plaid@AhmedML@8@r2ahead\endcsname{3}
\expandafter\def\csname plaid@AhmedML@8@knodes\endcsname{1108}
\expandafter\def\csname plaid@AhmedML@8@peak\endcsname{2.19}
\expandafter\def\csname plaid@all@8@onfront\endcsname{1}
\expandafter\def\csname plaid@all@8@onfronttotal\endcsname{1}
\expandafter\def\csname plaid@all@8@onfrontqten\endcsname{1}
\expandafter\def\csname plaid@all@8@onfrontqhundred\endcsname{1}
\expandafter\def\csname plaid@all@8@mostaccurate\endcsname{1}
\expandafter\def\csname plaid@all@8@beatsnn\endcsname{1}
\expandafter\def\csname plaid@all@8@datasets\endcsname{1}
\expandafter\def\csname plaid@all@8@scalars\endcsname{4}
\expandafter\def\csname plaid@all@8@r2ahead\endcsname{3}
\expandafter\def\csname plaid@all@8@covmin\endcsname{0.951}
\expandafter\def\csname plaid@all@8@covmax\endcsname{0.951}
\expandafter\def\csname plaid@all@8@covfieldmin\endcsname{0.925}
\expandafter\def\csname plaid@all@8@covfieldmax\endcsname{0.967}
\expandafter\def\csname plaid@AhmedML@16@physicl\endcsname{0.0978}
\expandafter\def\csname plaid@AhmedML@16@nn\endcsname{0.392}
\expandafter\def\csname plaid@AhmedML@16@spread\endcsname{0.00467}
\expandafter\def\csname plaid@AhmedML@16@spreadpct\endcsname{5}
\expandafter\def\csname plaid@AhmedML@16@draws\endcsname{3}
\expandafter\def\csname plaid@AhmedML@16@mean\endcsname{1.02}
\expandafter\def\csname plaid@AhmedML@16@best\endcsname{0.184}
\expandafter\def\csname plaid@AhmedML@16@bestname\endcsname{probe\_kumo[large,e8,v0.9999]}
\expandafter\def\csname plaid@AhmedML@16@floor\endcsname{0.101}
\expandafter\def\csname plaid@AhmedML@16@ratio\endcsname{0.531}
\expandafter\def\csname plaid@AhmedML@16@onfront\endcsname{yes}
\expandafter\def\csname plaid@AhmedML@16@onfronttotal\endcsname{yes}
\expandafter\def\csname plaid@AhmedML@16@onfrontqten\endcsname{yes}
\expandafter\def\csname plaid@AhmedML@16@onfrontqhundred\endcsname{yes}
\expandafter\def\csname plaid@AhmedML@16@beatsnn\endcsname{yes}
\expandafter\def\csname plaid@AhmedML@16@mostaccurate\endcsname{yes}
\expandafter\def\csname plaid@AhmedML@16@fit\endcsname{35.1}
\expandafter\def\csname plaid@AhmedML@16@query\endcsname{28.2}
\expandafter\def\csname plaid@AhmedML@16@bestfit\endcsname{88.6}
\expandafter\def\csname plaid@AhmedML@16@bestquery\endcsname{49.2}
\expandafter\def\csname plaid@AhmedML@16@coverage\endcsname{0.906}
\expandafter\def\csname plaid@AhmedML@16@scalars\endcsname{4}
\expandafter\def\csname plaid@AhmedML@16@covfmin\endcsname{0.841}
\expandafter\def\csname plaid@AhmedML@16@covfmax\endcsname{0.937}
\expandafter\def\csname plaid@AhmedML@16@r2\endcsname{0.375}
\expandafter\def\csname plaid@AhmedML@16@bestr2\endcsname{0.444}
\expandafter\def\csname plaid@AhmedML@16@r2ahead\endcsname{0}
\expandafter\def\csname plaid@AhmedML@16@knodes\endcsname{1108}
\expandafter\def\csname plaid@AhmedML@16@peak\endcsname{3.12}
\expandafter\def\csname plaid@all@16@onfront\endcsname{1}
\expandafter\def\csname plaid@all@16@onfronttotal\endcsname{1}
\expandafter\def\csname plaid@all@16@onfrontqten\endcsname{1}
\expandafter\def\csname plaid@all@16@onfrontqhundred\endcsname{1}
\expandafter\def\csname plaid@all@16@mostaccurate\endcsname{1}
\expandafter\def\csname plaid@all@16@beatsnn\endcsname{1}
\expandafter\def\csname plaid@all@16@datasets\endcsname{1}
\expandafter\def\csname plaid@all@16@scalars\endcsname{4}
\expandafter\def\csname plaid@all@16@r2ahead\endcsname{0}
\expandafter\def\csname plaid@all@16@covmin\endcsname{0.906}
\expandafter\def\csname plaid@all@16@covmax\endcsname{0.906}
\expandafter\def\csname plaid@all@16@covfieldmin\endcsname{0.841}
\expandafter\def\csname plaid@all@16@covfieldmax\endcsname{0.937}
\expandafter\def\csname plaid@AhmedML@32@physicl\endcsname{0.0672}
\expandafter\def\csname plaid@AhmedML@32@nn\endcsname{0.333}
\expandafter\def\csname plaid@AhmedML@32@spread\endcsname{--}
\expandafter\def\csname plaid@AhmedML@32@spreadpct\endcsname{--}
\expandafter\def\csname plaid@AhmedML@32@draws\endcsname{1}
\expandafter\def\csname plaid@AhmedML@32@mean\endcsname{1.01}
\expandafter\def\csname plaid@AhmedML@32@best\endcsname{0.124}
\expandafter\def\csname plaid@AhmedML@32@bestname\endcsname{probe\_kumo[large,e8,v0.9999]}
\expandafter\def\csname plaid@AhmedML@32@floor\endcsname{0.0631}
\expandafter\def\csname plaid@AhmedML@32@ratio\endcsname{0.543}
\expandafter\def\csname plaid@AhmedML@32@onfront\endcsname{yes}
\expandafter\def\csname plaid@AhmedML@32@onfronttotal\endcsname{yes}
\expandafter\def\csname plaid@AhmedML@32@onfrontqten\endcsname{yes}
\expandafter\def\csname plaid@AhmedML@32@onfrontqhundred\endcsname{yes}
\expandafter\def\csname plaid@AhmedML@32@beatsnn\endcsname{yes}
\expandafter\def\csname plaid@AhmedML@32@mostaccurate\endcsname{yes}
\expandafter\def\csname plaid@AhmedML@32@fit\endcsname{49.2}
\expandafter\def\csname plaid@AhmedML@32@query\endcsname{26.8}
\expandafter\def\csname plaid@AhmedML@32@bestfit\endcsname{122}
\expandafter\def\csname plaid@AhmedML@32@bestquery\endcsname{54.5}
\expandafter\def\csname plaid@AhmedML@32@coverage\endcsname{0.919}
\expandafter\def\csname plaid@AhmedML@32@scalars\endcsname{4}
\expandafter\def\csname plaid@AhmedML@32@covfmin\endcsname{0.898}
\expandafter\def\csname plaid@AhmedML@32@covfmax\endcsname{0.943}
\expandafter\def\csname plaid@AhmedML@32@r2\endcsname{0.697}
\expandafter\def\csname plaid@AhmedML@32@bestr2\endcsname{0.714}
\expandafter\def\csname plaid@AhmedML@32@r2ahead\endcsname{2}
\expandafter\def\csname plaid@AhmedML@32@knodes\endcsname{1108}
\expandafter\def\csname plaid@AhmedML@32@peak\endcsname{3.36}
\expandafter\def\csname plaid@all@32@onfront\endcsname{1}
\expandafter\def\csname plaid@all@32@onfronttotal\endcsname{1}
\expandafter\def\csname plaid@all@32@onfrontqten\endcsname{1}
\expandafter\def\csname plaid@all@32@onfrontqhundred\endcsname{1}
\expandafter\def\csname plaid@all@32@mostaccurate\endcsname{1}
\expandafter\def\csname plaid@all@32@beatsnn\endcsname{1}
\expandafter\def\csname plaid@all@32@datasets\endcsname{1}
\expandafter\def\csname plaid@all@32@scalars\endcsname{4}
\expandafter\def\csname plaid@all@32@r2ahead\endcsname{2}
\expandafter\def\csname plaid@all@32@covmin\endcsname{0.919}
\expandafter\def\csname plaid@all@32@covmax\endcsname{0.919}
\expandafter\def\csname plaid@all@32@covfieldmin\endcsname{0.898}
\expandafter\def\csname plaid@all@32@covfieldmax\endcsname{0.943}
\expandafter\def\csname plaid@AhmedML@64@physicl\endcsname{0.0542}
\expandafter\def\csname plaid@AhmedML@64@nn\endcsname{0.196}
\expandafter\def\csname plaid@AhmedML@64@spread\endcsname{0.00282}
\expandafter\def\csname plaid@AhmedML@64@spreadpct\endcsname{5}
\expandafter\def\csname plaid@AhmedML@64@draws\endcsname{3}
\expandafter\def\csname plaid@AhmedML@64@mean\endcsname{1.02}
\expandafter\def\csname plaid@AhmedML@64@best\endcsname{0.0725}
\expandafter\def\csname plaid@AhmedML@64@bestname\endcsname{probe\_kumo[large,e8,v0.9999]}
\expandafter\def\csname plaid@AhmedML@64@floor\endcsname{0.0351}
\expandafter\def\csname plaid@AhmedML@64@ratio\endcsname{0.748}
\expandafter\def\csname plaid@AhmedML@64@onfront\endcsname{yes}
\expandafter\def\csname plaid@AhmedML@64@onfronttotal\endcsname{yes}
\expandafter\def\csname plaid@AhmedML@64@onfrontqten\endcsname{yes}
\expandafter\def\csname plaid@AhmedML@64@onfrontqhundred\endcsname{yes}
\expandafter\def\csname plaid@AhmedML@64@beatsnn\endcsname{yes}
\expandafter\def\csname plaid@AhmedML@64@mostaccurate\endcsname{yes}
\expandafter\def\csname plaid@AhmedML@64@fit\endcsname{79.5}
\expandafter\def\csname plaid@AhmedML@64@query\endcsname{28.5}
\expandafter\def\csname plaid@AhmedML@64@bestfit\endcsname{170}
\expandafter\def\csname plaid@AhmedML@64@bestquery\endcsname{63}
\expandafter\def\csname plaid@AhmedML@64@coverage\endcsname{0.919}
\expandafter\def\csname plaid@AhmedML@64@scalars\endcsname{4}
\expandafter\def\csname plaid@AhmedML@64@covfmin\endcsname{0.902}
\expandafter\def\csname plaid@AhmedML@64@covfmax\endcsname{0.948}
\expandafter\def\csname plaid@AhmedML@64@r2\endcsname{0.781}
\expandafter\def\csname plaid@AhmedML@64@bestr2\endcsname{0.755}
\expandafter\def\csname plaid@AhmedML@64@r2ahead\endcsname{4}
\expandafter\def\csname plaid@AhmedML@64@knodes\endcsname{1108}
\expandafter\def\csname plaid@AhmedML@64@peak\endcsname{4.27}
\expandafter\def\csname plaid@all@64@onfront\endcsname{1}
\expandafter\def\csname plaid@all@64@onfronttotal\endcsname{1}
\expandafter\def\csname plaid@all@64@onfrontqten\endcsname{1}
\expandafter\def\csname plaid@all@64@onfrontqhundred\endcsname{1}
\expandafter\def\csname plaid@all@64@mostaccurate\endcsname{1}
\expandafter\def\csname plaid@all@64@beatsnn\endcsname{1}
\expandafter\def\csname plaid@all@64@datasets\endcsname{1}
\expandafter\def\csname plaid@all@64@scalars\endcsname{4}
\expandafter\def\csname plaid@all@64@r2ahead\endcsname{4}
\expandafter\def\csname plaid@all@64@covmin\endcsname{0.919}
\expandafter\def\csname plaid@all@64@covmax\endcsname{0.919}
\expandafter\def\csname plaid@all@64@covfieldmin\endcsname{0.902}
\expandafter\def\csname plaid@all@64@covfieldmax\endcsname{0.948}
\expandafter\def\csname plaid@AhmedML@125@physicl\endcsname{0.0454}
\expandafter\def\csname plaid@AhmedML@125@nn\endcsname{0.161}
\expandafter\def\csname plaid@AhmedML@125@spread\endcsname{--}
\expandafter\def\csname plaid@AhmedML@125@spreadpct\endcsname{--}
\expandafter\def\csname plaid@AhmedML@125@draws\endcsname{1}
\expandafter\def\csname plaid@AhmedML@125@mean\endcsname{1}
\expandafter\def\csname plaid@AhmedML@125@best\endcsname{0.0457}
\expandafter\def\csname plaid@AhmedML@125@bestname\endcsname{probe\_kumo[large,e8,v0.9999]}
\expandafter\def\csname plaid@AhmedML@125@floor\endcsname{0.0217}
\expandafter\def\csname plaid@AhmedML@125@ratio\endcsname{0.993}
\expandafter\def\csname plaid@AhmedML@125@onfront\endcsname{yes}
\expandafter\def\csname plaid@AhmedML@125@onfronttotal\endcsname{yes}
\expandafter\def\csname plaid@AhmedML@125@onfrontqten\endcsname{yes}
\expandafter\def\csname plaid@AhmedML@125@onfrontqhundred\endcsname{yes}
\expandafter\def\csname plaid@AhmedML@125@beatsnn\endcsname{yes}
\expandafter\def\csname plaid@AhmedML@125@mostaccurate\endcsname{yes}
\expandafter\def\csname plaid@AhmedML@125@fit\endcsname{136}
\expandafter\def\csname plaid@AhmedML@125@query\endcsname{28.3}
\expandafter\def\csname plaid@AhmedML@125@bestfit\endcsname{383}
\expandafter\def\csname plaid@AhmedML@125@bestquery\endcsname{83.1}
\expandafter\def\csname plaid@AhmedML@125@coverage\endcsname{0.929}
\expandafter\def\csname plaid@AhmedML@125@scalars\endcsname{4}
\expandafter\def\csname plaid@AhmedML@125@covfmin\endcsname{0.909}
\expandafter\def\csname plaid@AhmedML@125@covfmax\endcsname{0.953}
\expandafter\def\csname plaid@AhmedML@125@r2\endcsname{0.826}
\expandafter\def\csname plaid@AhmedML@125@bestr2\endcsname{0.823}
\expandafter\def\csname plaid@AhmedML@125@r2ahead\endcsname{4}
\expandafter\def\csname plaid@AhmedML@125@knodes\endcsname{1108}
\expandafter\def\csname plaid@AhmedML@125@peak\endcsname{7.06}
\expandafter\def\csname plaid@all@125@onfront\endcsname{1}
\expandafter\def\csname plaid@all@125@onfronttotal\endcsname{1}
\expandafter\def\csname plaid@all@125@onfrontqten\endcsname{1}
\expandafter\def\csname plaid@all@125@onfrontqhundred\endcsname{1}
\expandafter\def\csname plaid@all@125@mostaccurate\endcsname{1}
\expandafter\def\csname plaid@all@125@beatsnn\endcsname{1}
\expandafter\def\csname plaid@all@125@datasets\endcsname{1}
\expandafter\def\csname plaid@all@125@scalars\endcsname{4}
\expandafter\def\csname plaid@all@125@r2ahead\endcsname{4}
\expandafter\def\csname plaid@all@125@covmin\endcsname{0.929}
\expandafter\def\csname plaid@all@125@covmax\endcsname{0.929}
\expandafter\def\csname plaid@all@125@covfieldmin\endcsname{0.909}
\expandafter\def\csname plaid@all@125@covfieldmax\endcsname{0.953}
\expandafter\def\csname plaid@AhmedML@250@physicl\endcsname{0.042}
\expandafter\def\csname plaid@AhmedML@250@nn\endcsname{0.146}
\expandafter\def\csname plaid@AhmedML@250@spread\endcsname{0.00151}
\expandafter\def\csname plaid@AhmedML@250@spreadpct\endcsname{4}
\expandafter\def\csname plaid@AhmedML@250@draws\endcsname{3}
\expandafter\def\csname plaid@AhmedML@250@mean\endcsname{1}
\expandafter\def\csname plaid@AhmedML@250@best\endcsname{0.0331}
\expandafter\def\csname plaid@AhmedML@250@bestname\endcsname{probe\_kumo[large,e8,v0.9999]}
\expandafter\def\csname plaid@AhmedML@250@floor\endcsname{0.0155}
\expandafter\def\csname plaid@AhmedML@250@ratio\endcsname{1.27}
\expandafter\def\csname plaid@AhmedML@250@onfront\endcsname{yes}
\expandafter\def\csname plaid@AhmedML@250@onfronttotal\endcsname{yes}
\expandafter\def\csname plaid@AhmedML@250@onfrontqten\endcsname{yes}
\expandafter\def\csname plaid@AhmedML@250@onfrontqhundred\endcsname{yes}
\expandafter\def\csname plaid@AhmedML@250@beatsnn\endcsname{yes}
\expandafter\def\csname plaid@AhmedML@250@mostaccurate\endcsname{no}
\expandafter\def\csname plaid@AhmedML@250@fit\endcsname{229}
\expandafter\def\csname plaid@AhmedML@250@query\endcsname{31}
\expandafter\def\csname plaid@AhmedML@250@bestfit\endcsname{1{,}080}
\expandafter\def\csname plaid@AhmedML@250@bestquery\endcsname{134}
\expandafter\def\csname plaid@AhmedML@250@coverage\endcsname{0.93}
\expandafter\def\csname plaid@AhmedML@250@scalars\endcsname{4}
\expandafter\def\csname plaid@AhmedML@250@covfmin\endcsname{0.907}
\expandafter\def\csname plaid@AhmedML@250@covfmax\endcsname{0.961}
\expandafter\def\csname plaid@AhmedML@250@r2\endcsname{0.88}
\expandafter\def\csname plaid@AhmedML@250@bestr2\endcsname{0.882}
\expandafter\def\csname plaid@AhmedML@250@r2ahead\endcsname{1}
\expandafter\def\csname plaid@AhmedML@250@knodes\endcsname{1108}
\expandafter\def\csname plaid@AhmedML@250@peak\endcsname{13.4}
\expandafter\def\csname plaid@all@250@onfront\endcsname{1}
\expandafter\def\csname plaid@all@250@onfronttotal\endcsname{1}
\expandafter\def\csname plaid@all@250@onfrontqten\endcsname{1}
\expandafter\def\csname plaid@all@250@onfrontqhundred\endcsname{1}
\expandafter\def\csname plaid@all@250@mostaccurate\endcsname{0}
\expandafter\def\csname plaid@all@250@beatsnn\endcsname{1}
\expandafter\def\csname plaid@all@250@datasets\endcsname{1}
\expandafter\def\csname plaid@all@250@scalars\endcsname{4}
\expandafter\def\csname plaid@all@250@r2ahead\endcsname{1}
\expandafter\def\csname plaid@all@250@covmin\endcsname{0.93}
\expandafter\def\csname plaid@all@250@covmax\endcsname{0.93}
\expandafter\def\csname plaid@all@250@covfieldmin\endcsname{0.907}
\expandafter\def\csname plaid@all@250@covfieldmax\endcsname{0.961}
\expandafter\def\csname plaid@all@64@decreasing\endcsname{1}
}{}
\IfFileExists{results/plaid.tex}{\providecommand{\plaid}[3]{\ifcsname plaid@#1@#2@#3\endcsname\csname plaid@#1@#2@#3\endcsname\else\todo{#1 #2 #3}\fi}
\expandafter\def\csname plaid@2D_profile@8@physicl\endcsname{0.158}
\expandafter\def\csname plaid@2D_profile@8@nn\endcsname{0.285}
\expandafter\def\csname plaid@2D_profile@8@spread\endcsname{0.0233}
\expandafter\def\csname plaid@2D_profile@8@spreadpct\endcsname{15}
\expandafter\def\csname plaid@2D_profile@8@draws\endcsname{3}
\expandafter\def\csname plaid@2D_profile@8@mean\endcsname{0.982}
\expandafter\def\csname plaid@2D_profile@8@best\endcsname{0.188}
\expandafter\def\csname plaid@2D_profile@8@bestname\endcsname{probe\_kumo[medium,e8,v0.99]}
\expandafter\def\csname plaid@2D_profile@8@floor\endcsname{0.147}
\expandafter\def\csname plaid@2D_profile@8@ratio\endcsname{0.841}
\expandafter\def\csname plaid@2D_profile@8@onfront\endcsname{yes}
\expandafter\def\csname plaid@2D_profile@8@onfronttotal\endcsname{yes}
\expandafter\def\csname plaid@2D_profile@8@onfrontqten\endcsname{yes}
\expandafter\def\csname plaid@2D_profile@8@onfrontqhundred\endcsname{yes}
\expandafter\def\csname plaid@2D_profile@8@beatsnn\endcsname{yes}
\expandafter\def\csname plaid@2D_profile@8@mostaccurate\endcsname{yes}
\expandafter\def\csname plaid@2D_profile@8@fit\endcsname{14.1}
\expandafter\def\csname plaid@2D_profile@8@query\endcsname{0.9}
\expandafter\def\csname plaid@2D_profile@8@bestfit\endcsname{4.54}
\expandafter\def\csname plaid@2D_profile@8@bestquery\endcsname{1.55}
\expandafter\def\csname plaid@2D_profile@8@coverage\endcsname{0.865}
\expandafter\def\csname plaid@2D_profile@8@scalars\endcsname{0}
\expandafter\def\csname plaid@2D_profile@8@covfmin\endcsname{0.838}
\expandafter\def\csname plaid@2D_profile@8@covfmax\endcsname{0.881}
\expandafter\def\csname plaid@2D_profile@8@r2\endcsname{--}
\expandafter\def\csname plaid@2D_profile@8@bestr2\endcsname{--}
\expandafter\def\csname plaid@2D_profile@8@r2ahead\endcsname{0}
\expandafter\def\csname plaid@2D_profile@8@knodes\endcsname{37}
\expandafter\def\csname plaid@2D_profile@8@peak\endcsname{1.59}
\expandafter\def\csname plaid@2D_MSHE@8@physicl\endcsname{0.555}
\expandafter\def\csname plaid@2D_MSHE@8@nn\endcsname{1.34}
\expandafter\def\csname plaid@2D_MSHE@8@spread\endcsname{0.007}
\expandafter\def\csname plaid@2D_MSHE@8@spreadpct\endcsname{1}
\expandafter\def\csname plaid@2D_MSHE@8@draws\endcsname{3}
\expandafter\def\csname plaid@2D_MSHE@8@mean\endcsname{0.861}
\expandafter\def\csname plaid@2D_MSHE@8@best\endcsname{0.705}
\expandafter\def\csname plaid@2D_MSHE@8@bestname\endcsname{probe\_gp[small,v0.9999]}
\expandafter\def\csname plaid@2D_MSHE@8@floor\endcsname{0.615}
\expandafter\def\csname plaid@2D_MSHE@8@ratio\endcsname{0.788}
\expandafter\def\csname plaid@2D_MSHE@8@onfront\endcsname{yes}
\expandafter\def\csname plaid@2D_MSHE@8@onfronttotal\endcsname{yes}
\expandafter\def\csname plaid@2D_MSHE@8@onfrontqten\endcsname{yes}
\expandafter\def\csname plaid@2D_MSHE@8@onfrontqhundred\endcsname{yes}
\expandafter\def\csname plaid@2D_MSHE@8@beatsnn\endcsname{yes}
\expandafter\def\csname plaid@2D_MSHE@8@mostaccurate\endcsname{yes}
\expandafter\def\csname plaid@2D_MSHE@8@fit\endcsname{13.9}
\expandafter\def\csname plaid@2D_MSHE@8@query\endcsname{0.805}
\expandafter\def\csname plaid@2D_MSHE@8@bestfit\endcsname{7.93}
\expandafter\def\csname plaid@2D_MSHE@8@bestquery\endcsname{0.0372}
\expandafter\def\csname plaid@2D_MSHE@8@coverage\endcsname{0.888}
\expandafter\def\csname plaid@2D_MSHE@8@scalars\endcsname{1}
\expandafter\def\csname plaid@2D_MSHE@8@covfmin\endcsname{0.786}
\expandafter\def\csname plaid@2D_MSHE@8@covfmax\endcsname{0.981}
\expandafter\def\csname plaid@2D_MSHE@8@r2\endcsname{-1.6}
\expandafter\def\csname plaid@2D_MSHE@8@bestr2\endcsname{-1.41}
\expandafter\def\csname plaid@2D_MSHE@8@r2ahead\endcsname{0}
\expandafter\def\csname plaid@2D_MSHE@8@knodes\endcsname{6}
\expandafter\def\csname plaid@2D_MSHE@8@peak\endcsname{1.23}
\expandafter\def\csname plaid@Tensile2d@8@physicl\endcsname{0.00863}
\expandafter\def\csname plaid@Tensile2d@8@nn\endcsname{0.0547}
\expandafter\def\csname plaid@Tensile2d@8@spread\endcsname{0.00607}
\expandafter\def\csname plaid@Tensile2d@8@spreadpct\endcsname{70}
\expandafter\def\csname plaid@Tensile2d@8@draws\endcsname{3}
\expandafter\def\csname plaid@Tensile2d@8@mean\endcsname{0.573}
\expandafter\def\csname plaid@Tensile2d@8@best\endcsname{0.00728}
\expandafter\def\csname plaid@Tensile2d@8@bestname\endcsname{probe\_gp[medium,v0.99]}
\expandafter\def\csname plaid@Tensile2d@8@floor\endcsname{0.000615}
\expandafter\def\csname plaid@Tensile2d@8@ratio\endcsname{1.19}
\expandafter\def\csname plaid@Tensile2d@8@onfront\endcsname{no}
\expandafter\def\csname plaid@Tensile2d@8@onfronttotal\endcsname{no}
\expandafter\def\csname plaid@Tensile2d@8@onfrontqten\endcsname{no}
\expandafter\def\csname plaid@Tensile2d@8@onfrontqhundred\endcsname{no}
\expandafter\def\csname plaid@Tensile2d@8@beatsnn\endcsname{yes}
\expandafter\def\csname plaid@Tensile2d@8@mostaccurate\endcsname{no}
\expandafter\def\csname plaid@Tensile2d@8@fit\endcsname{14}
\expandafter\def\csname plaid@Tensile2d@8@query\endcsname{0.9}
\expandafter\def\csname plaid@Tensile2d@8@bestfit\endcsname{6.9}
\expandafter\def\csname plaid@Tensile2d@8@bestquery\endcsname{0.036}
\expandafter\def\csname plaid@Tensile2d@8@coverage\endcsname{0.996}
\expandafter\def\csname plaid@Tensile2d@8@scalars\endcsname{3}
\expandafter\def\csname plaid@Tensile2d@8@covfmin\endcsname{0.991}
\expandafter\def\csname plaid@Tensile2d@8@covfmax\endcsname{0.999}
\expandafter\def\csname plaid@Tensile2d@8@r2\endcsname{0.833}
\expandafter\def\csname plaid@Tensile2d@8@bestr2\endcsname{0.979}
\expandafter\def\csname plaid@Tensile2d@8@r2ahead\endcsname{0}
\expandafter\def\csname plaid@Tensile2d@8@knodes\endcsname{10}
\expandafter\def\csname plaid@Tensile2d@8@peak\endcsname{1.35}
\expandafter\def\csname plaid@VKI-LS59@8@physicl\endcsname{0.113}
\expandafter\def\csname plaid@VKI-LS59@8@nn\endcsname{0.122}
\expandafter\def\csname plaid@VKI-LS59@8@spread\endcsname{0.143}
\expandafter\def\csname plaid@VKI-LS59@8@spreadpct\endcsname{126}
\expandafter\def\csname plaid@VKI-LS59@8@draws\endcsname{3}
\expandafter\def\csname plaid@VKI-LS59@8@mean\endcsname{0.493}
\expandafter\def\csname plaid@VKI-LS59@8@best\endcsname{0.122}
\expandafter\def\csname plaid@VKI-LS59@8@bestname\endcsname{pca\_kumo[large,e4,v0.99]}
\expandafter\def\csname plaid@VKI-LS59@8@floor\endcsname{0.0341}
\expandafter\def\csname plaid@VKI-LS59@8@ratio\endcsname{0.927}
\expandafter\def\csname plaid@VKI-LS59@8@onfront\endcsname{yes}
\expandafter\def\csname plaid@VKI-LS59@8@onfronttotal\endcsname{yes}
\expandafter\def\csname plaid@VKI-LS59@8@onfrontqten\endcsname{yes}
\expandafter\def\csname plaid@VKI-LS59@8@onfrontqhundred\endcsname{yes}
\expandafter\def\csname plaid@VKI-LS59@8@beatsnn\endcsname{yes}
\expandafter\def\csname plaid@VKI-LS59@8@mostaccurate\endcsname{yes}
\expandafter\def\csname plaid@VKI-LS59@8@fit\endcsname{15.9}
\expandafter\def\csname plaid@VKI-LS59@8@query\endcsname{0.892}
\expandafter\def\csname plaid@VKI-LS59@8@bestfit\endcsname{1.58}
\expandafter\def\csname plaid@VKI-LS59@8@bestquery\endcsname{0.456}
\expandafter\def\csname plaid@VKI-LS59@8@coverage\endcsname{0.825}
\expandafter\def\csname plaid@VKI-LS59@8@scalars\endcsname{6}
\expandafter\def\csname plaid@VKI-LS59@8@covfmin\endcsname{0.672}
\expandafter\def\csname plaid@VKI-LS59@8@covfmax\endcsname{0.979}
\expandafter\def\csname plaid@VKI-LS59@8@r2\endcsname{-0.11}
\expandafter\def\csname plaid@VKI-LS59@8@bestr2\endcsname{0.423}
\expandafter\def\csname plaid@VKI-LS59@8@r2ahead\endcsname{0}
\expandafter\def\csname plaid@VKI-LS59@8@knodes\endcsname{0}
\expandafter\def\csname plaid@VKI-LS59@8@peak\endcsname{1.56}
\expandafter\def\csname plaid@Rotor37@8@physicl\endcsname{0.0482}
\expandafter\def\csname plaid@Rotor37@8@nn\endcsname{0.0357}
\expandafter\def\csname plaid@Rotor37@8@spread\endcsname{0.00268}
\expandafter\def\csname plaid@Rotor37@8@spreadpct\endcsname{6}
\expandafter\def\csname plaid@Rotor37@8@draws\endcsname{3}
\expandafter\def\csname plaid@Rotor37@8@mean\endcsname{0.981}
\expandafter\def\csname plaid@Rotor37@8@best\endcsname{0.00907}
\expandafter\def\csname plaid@Rotor37@8@bestname\endcsname{pca\_kumo[medium,e1,v0.99]}
\expandafter\def\csname plaid@Rotor37@8@floor\endcsname{0.00523}
\expandafter\def\csname plaid@Rotor37@8@ratio\endcsname{5.31}
\expandafter\def\csname plaid@Rotor37@8@onfront\endcsname{no}
\expandafter\def\csname plaid@Rotor37@8@onfronttotal\endcsname{no}
\expandafter\def\csname plaid@Rotor37@8@onfrontqten\endcsname{no}
\expandafter\def\csname plaid@Rotor37@8@onfrontqhundred\endcsname{no}
\expandafter\def\csname plaid@Rotor37@8@beatsnn\endcsname{no}
\expandafter\def\csname plaid@Rotor37@8@mostaccurate\endcsname{no}
\expandafter\def\csname plaid@Rotor37@8@fit\endcsname{13.3}
\expandafter\def\csname plaid@Rotor37@8@query\endcsname{1.07}
\expandafter\def\csname plaid@Rotor37@8@bestfit\endcsname{1.46}
\expandafter\def\csname plaid@Rotor37@8@bestquery\endcsname{0.613}
\expandafter\def\csname plaid@Rotor37@8@coverage\endcsname{0.926}
\expandafter\def\csname plaid@Rotor37@8@scalars\endcsname{3}
\expandafter\def\csname plaid@Rotor37@8@covfmin\endcsname{0.912}
\expandafter\def\csname plaid@Rotor37@8@covfmax\endcsname{0.933}
\expandafter\def\csname plaid@Rotor37@8@r2\endcsname{0.573}
\expandafter\def\csname plaid@Rotor37@8@bestr2\endcsname{0.717}
\expandafter\def\csname plaid@Rotor37@8@r2ahead\endcsname{0}
\expandafter\def\csname plaid@Rotor37@8@knodes\endcsname{30}
\expandafter\def\csname plaid@Rotor37@8@peak\endcsname{1.58}
\expandafter\def\csname plaid@all@8@onfront\endcsname{3}
\expandafter\def\csname plaid@all@8@onfronttotal\endcsname{3}
\expandafter\def\csname plaid@all@8@onfrontqten\endcsname{3}
\expandafter\def\csname plaid@all@8@onfrontqhundred\endcsname{3}
\expandafter\def\csname plaid@all@8@mostaccurate\endcsname{3}
\expandafter\def\csname plaid@all@8@beatsnn\endcsname{4}
\expandafter\def\csname plaid@all@8@datasets\endcsname{5}
\expandafter\def\csname plaid@all@8@scalars\endcsname{13}
\expandafter\def\csname plaid@all@8@r2ahead\endcsname{0}
\expandafter\def\csname plaid@all@8@covmin\endcsname{0.825}
\expandafter\def\csname plaid@all@8@covmax\endcsname{0.996}
\expandafter\def\csname plaid@all@8@covfieldmin\endcsname{0.672}
\expandafter\def\csname plaid@all@8@covfieldmax\endcsname{0.999}
\expandafter\def\csname plaid@2D_profile@16@physicl\endcsname{0.116}
\expandafter\def\csname plaid@2D_profile@16@nn\endcsname{0.228}
\expandafter\def\csname plaid@2D_profile@16@spread\endcsname{0.0088}
\expandafter\def\csname plaid@2D_profile@16@spreadpct\endcsname{8}
\expandafter\def\csname plaid@2D_profile@16@draws\endcsname{3}
\expandafter\def\csname plaid@2D_profile@16@mean\endcsname{0.998}
\expandafter\def\csname plaid@2D_profile@16@best\endcsname{0.129}
\expandafter\def\csname plaid@2D_profile@16@bestname\endcsname{probe\_kumo[medium,e8,v0.99]}
\expandafter\def\csname plaid@2D_profile@16@floor\endcsname{0.101}
\expandafter\def\csname plaid@2D_profile@16@ratio\endcsname{0.904}
\expandafter\def\csname plaid@2D_profile@16@onfront\endcsname{yes}
\expandafter\def\csname plaid@2D_profile@16@onfronttotal\endcsname{yes}
\expandafter\def\csname plaid@2D_profile@16@onfrontqten\endcsname{yes}
\expandafter\def\csname plaid@2D_profile@16@onfrontqhundred\endcsname{yes}
\expandafter\def\csname plaid@2D_profile@16@beatsnn\endcsname{yes}
\expandafter\def\csname plaid@2D_profile@16@mostaccurate\endcsname{yes}
\expandafter\def\csname plaid@2D_profile@16@fit\endcsname{15.5}
\expandafter\def\csname plaid@2D_profile@16@query\endcsname{0.95}
\expandafter\def\csname plaid@2D_profile@16@bestfit\endcsname{7.88}
\expandafter\def\csname plaid@2D_profile@16@bestquery\endcsname{3.78}
\expandafter\def\csname plaid@2D_profile@16@coverage\endcsname{0.892}
\expandafter\def\csname plaid@2D_profile@16@scalars\endcsname{0}
\expandafter\def\csname plaid@2D_profile@16@covfmin\endcsname{0.864}
\expandafter\def\csname plaid@2D_profile@16@covfmax\endcsname{0.921}
\expandafter\def\csname plaid@2D_profile@16@r2\endcsname{--}
\expandafter\def\csname plaid@2D_profile@16@bestr2\endcsname{--}
\expandafter\def\csname plaid@2D_profile@16@r2ahead\endcsname{0}
\expandafter\def\csname plaid@2D_profile@16@knodes\endcsname{37}
\expandafter\def\csname plaid@2D_profile@16@peak\endcsname{2.82}
\expandafter\def\csname plaid@2D_MSHE@16@physicl\endcsname{0.502}
\expandafter\def\csname plaid@2D_MSHE@16@nn\endcsname{1.23}
\expandafter\def\csname plaid@2D_MSHE@16@spread\endcsname{0.019}
\expandafter\def\csname plaid@2D_MSHE@16@spreadpct\endcsname{4}
\expandafter\def\csname plaid@2D_MSHE@16@draws\endcsname{3}
\expandafter\def\csname plaid@2D_MSHE@16@mean\endcsname{0.89}
\expandafter\def\csname plaid@2D_MSHE@16@best\endcsname{0.628}
\expandafter\def\csname plaid@2D_MSHE@16@bestname\endcsname{probe\_kumo[large,e4,v0.99]}
\expandafter\def\csname plaid@2D_MSHE@16@floor\endcsname{0.505}
\expandafter\def\csname plaid@2D_MSHE@16@ratio\endcsname{0.8}
\expandafter\def\csname plaid@2D_MSHE@16@onfront\endcsname{yes}
\expandafter\def\csname plaid@2D_MSHE@16@onfronttotal\endcsname{yes}
\expandafter\def\csname plaid@2D_MSHE@16@onfrontqten\endcsname{yes}
\expandafter\def\csname plaid@2D_MSHE@16@onfrontqhundred\endcsname{yes}
\expandafter\def\csname plaid@2D_MSHE@16@beatsnn\endcsname{yes}
\expandafter\def\csname plaid@2D_MSHE@16@mostaccurate\endcsname{yes}
\expandafter\def\csname plaid@2D_MSHE@16@fit\endcsname{15.5}
\expandafter\def\csname plaid@2D_MSHE@16@query\endcsname{1.37}
\expandafter\def\csname plaid@2D_MSHE@16@bestfit\endcsname{11.2}
\expandafter\def\csname plaid@2D_MSHE@16@bestquery\endcsname{3.57}
\expandafter\def\csname plaid@2D_MSHE@16@coverage\endcsname{0.899}
\expandafter\def\csname plaid@2D_MSHE@16@scalars\endcsname{1}
\expandafter\def\csname plaid@2D_MSHE@16@covfmin\endcsname{0.831}
\expandafter\def\csname plaid@2D_MSHE@16@covfmax\endcsname{0.953}
\expandafter\def\csname plaid@2D_MSHE@16@r2\endcsname{-0.162}
\expandafter\def\csname plaid@2D_MSHE@16@bestr2\endcsname{0.0408}
\expandafter\def\csname plaid@2D_MSHE@16@r2ahead\endcsname{0}
\expandafter\def\csname plaid@2D_MSHE@16@knodes\endcsname{6}
\expandafter\def\csname plaid@2D_MSHE@16@peak\endcsname{2.52}
\expandafter\def\csname plaid@Tensile2d@16@physicl\endcsname{0.00249}
\expandafter\def\csname plaid@Tensile2d@16@nn\endcsname{0.0308}
\expandafter\def\csname plaid@Tensile2d@16@spread\endcsname{0.000669}
\expandafter\def\csname plaid@Tensile2d@16@spreadpct\endcsname{27}
\expandafter\def\csname plaid@Tensile2d@16@draws\endcsname{3}
\expandafter\def\csname plaid@Tensile2d@16@mean\endcsname{0.526}
\expandafter\def\csname plaid@Tensile2d@16@best\endcsname{0.000534}
\expandafter\def\csname plaid@Tensile2d@16@bestname\endcsname{probe\_kumo[medium,e1,v0.9999]}
\expandafter\def\csname plaid@Tensile2d@16@floor\endcsname{0.000122}
\expandafter\def\csname plaid@Tensile2d@16@ratio\endcsname{4.67}
\expandafter\def\csname plaid@Tensile2d@16@onfront\endcsname{no}
\expandafter\def\csname plaid@Tensile2d@16@onfronttotal\endcsname{no}
\expandafter\def\csname plaid@Tensile2d@16@onfrontqten\endcsname{no}
\expandafter\def\csname plaid@Tensile2d@16@onfrontqhundred\endcsname{no}
\expandafter\def\csname plaid@Tensile2d@16@beatsnn\endcsname{yes}
\expandafter\def\csname plaid@Tensile2d@16@mostaccurate\endcsname{no}
\expandafter\def\csname plaid@Tensile2d@16@fit\endcsname{16.1}
\expandafter\def\csname plaid@Tensile2d@16@query\endcsname{1.18}
\expandafter\def\csname plaid@Tensile2d@16@bestfit\endcsname{4.96}
\expandafter\def\csname plaid@Tensile2d@16@bestquery\endcsname{0.913}
\expandafter\def\csname plaid@Tensile2d@16@coverage\endcsname{0.974}
\expandafter\def\csname plaid@Tensile2d@16@scalars\endcsname{3}
\expandafter\def\csname plaid@Tensile2d@16@covfmin\endcsname{0.966}
\expandafter\def\csname plaid@Tensile2d@16@covfmax\endcsname{0.986}
\expandafter\def\csname plaid@Tensile2d@16@r2\endcsname{0.898}
\expandafter\def\csname plaid@Tensile2d@16@bestr2\endcsname{0.996}
\expandafter\def\csname plaid@Tensile2d@16@r2ahead\endcsname{0}
\expandafter\def\csname plaid@Tensile2d@16@knodes\endcsname{10}
\expandafter\def\csname plaid@Tensile2d@16@peak\endcsname{2.6}
\expandafter\def\csname plaid@VKI-LS59@16@physicl\endcsname{0.0794}
\expandafter\def\csname plaid@VKI-LS59@16@nn\endcsname{0.1}
\expandafter\def\csname plaid@VKI-LS59@16@spread\endcsname{0.0828}
\expandafter\def\csname plaid@VKI-LS59@16@spreadpct\endcsname{104}
\expandafter\def\csname plaid@VKI-LS59@16@draws\endcsname{3}
\expandafter\def\csname plaid@VKI-LS59@16@mean\endcsname{0.485}
\expandafter\def\csname plaid@VKI-LS59@16@best\endcsname{0.0702}
\expandafter\def\csname plaid@VKI-LS59@16@bestname\endcsname{pca\_kumo[medium,e1,v0.99]}
\expandafter\def\csname plaid@VKI-LS59@16@floor\endcsname{0.00917}
\expandafter\def\csname plaid@VKI-LS59@16@ratio\endcsname{1.13}
\expandafter\def\csname plaid@VKI-LS59@16@onfront\endcsname{no}
\expandafter\def\csname plaid@VKI-LS59@16@onfronttotal\endcsname{no}
\expandafter\def\csname plaid@VKI-LS59@16@onfrontqten\endcsname{no}
\expandafter\def\csname plaid@VKI-LS59@16@onfrontqhundred\endcsname{no}
\expandafter\def\csname plaid@VKI-LS59@16@beatsnn\endcsname{yes}
\expandafter\def\csname plaid@VKI-LS59@16@mostaccurate\endcsname{no}
\expandafter\def\csname plaid@VKI-LS59@16@fit\endcsname{16.3}
\expandafter\def\csname plaid@VKI-LS59@16@query\endcsname{0.999}
\expandafter\def\csname plaid@VKI-LS59@16@bestfit\endcsname{1.6}
\expandafter\def\csname plaid@VKI-LS59@16@bestquery\endcsname{0.478}
\expandafter\def\csname plaid@VKI-LS59@16@coverage\endcsname{0.903}
\expandafter\def\csname plaid@VKI-LS59@16@scalars\endcsname{6}
\expandafter\def\csname plaid@VKI-LS59@16@covfmin\endcsname{0.857}
\expandafter\def\csname plaid@VKI-LS59@16@covfmax\endcsname{0.949}
\expandafter\def\csname plaid@VKI-LS59@16@r2\endcsname{0.0804}
\expandafter\def\csname plaid@VKI-LS59@16@bestr2\endcsname{0.761}
\expandafter\def\csname plaid@VKI-LS59@16@r2ahead\endcsname{0}
\expandafter\def\csname plaid@VKI-LS59@16@knodes\endcsname{0}
\expandafter\def\csname plaid@VKI-LS59@16@peak\endcsname{2.76}
\expandafter\def\csname plaid@Rotor37@16@physicl\endcsname{0.0336}
\expandafter\def\csname plaid@Rotor37@16@nn\endcsname{0.0383}
\expandafter\def\csname plaid@Rotor37@16@spread\endcsname{0.00152}
\expandafter\def\csname plaid@Rotor37@16@spreadpct\endcsname{5}
\expandafter\def\csname plaid@Rotor37@16@draws\endcsname{3}
\expandafter\def\csname plaid@Rotor37@16@mean\endcsname{0.998}
\expandafter\def\csname plaid@Rotor37@16@best\endcsname{0.00761}
\expandafter\def\csname plaid@Rotor37@16@bestname\endcsname{pca\_kumo[large,e4,v0.999]}
\expandafter\def\csname plaid@Rotor37@16@floor\endcsname{0.00282}
\expandafter\def\csname plaid@Rotor37@16@ratio\endcsname{4.42}
\expandafter\def\csname plaid@Rotor37@16@onfront\endcsname{no}
\expandafter\def\csname plaid@Rotor37@16@onfronttotal\endcsname{no}
\expandafter\def\csname plaid@Rotor37@16@onfrontqten\endcsname{no}
\expandafter\def\csname plaid@Rotor37@16@onfrontqhundred\endcsname{no}
\expandafter\def\csname plaid@Rotor37@16@beatsnn\endcsname{yes}
\expandafter\def\csname plaid@Rotor37@16@mostaccurate\endcsname{no}
\expandafter\def\csname plaid@Rotor37@16@fit\endcsname{16.2}
\expandafter\def\csname plaid@Rotor37@16@query\endcsname{0.906}
\expandafter\def\csname plaid@Rotor37@16@bestfit\endcsname{4.32}
\expandafter\def\csname plaid@Rotor37@16@bestquery\endcsname{2.15}
\expandafter\def\csname plaid@Rotor37@16@coverage\endcsname{0.915}
\expandafter\def\csname plaid@Rotor37@16@scalars\endcsname{3}
\expandafter\def\csname plaid@Rotor37@16@covfmin\endcsname{0.903}
\expandafter\def\csname plaid@Rotor37@16@covfmax\endcsname{0.921}
\expandafter\def\csname plaid@Rotor37@16@r2\endcsname{0.691}
\expandafter\def\csname plaid@Rotor37@16@bestr2\endcsname{0.791}
\expandafter\def\csname plaid@Rotor37@16@r2ahead\endcsname{0}
\expandafter\def\csname plaid@Rotor37@16@knodes\endcsname{30}
\expandafter\def\csname plaid@Rotor37@16@peak\endcsname{2.79}
\expandafter\def\csname plaid@all@16@onfront\endcsname{2}
\expandafter\def\csname plaid@all@16@onfronttotal\endcsname{2}
\expandafter\def\csname plaid@all@16@onfrontqten\endcsname{2}
\expandafter\def\csname plaid@all@16@onfrontqhundred\endcsname{2}
\expandafter\def\csname plaid@all@16@mostaccurate\endcsname{2}
\expandafter\def\csname plaid@all@16@beatsnn\endcsname{5}
\expandafter\def\csname plaid@all@16@datasets\endcsname{5}
\expandafter\def\csname plaid@all@16@scalars\endcsname{13}
\expandafter\def\csname plaid@all@16@r2ahead\endcsname{0}
\expandafter\def\csname plaid@all@16@covmin\endcsname{0.892}
\expandafter\def\csname plaid@all@16@covmax\endcsname{0.974}
\expandafter\def\csname plaid@all@16@covfieldmin\endcsname{0.831}
\expandafter\def\csname plaid@all@16@covfieldmax\endcsname{0.986}
\expandafter\def\csname plaid@2D_profile@32@physicl\endcsname{0.0842}
\expandafter\def\csname plaid@2D_profile@32@nn\endcsname{0.179}
\expandafter\def\csname plaid@2D_profile@32@spread\endcsname{0.0117}
\expandafter\def\csname plaid@2D_profile@32@spreadpct\endcsname{14}
\expandafter\def\csname plaid@2D_profile@32@draws\endcsname{3}
\expandafter\def\csname plaid@2D_profile@32@mean\endcsname{0.976}
\expandafter\def\csname plaid@2D_profile@32@best\endcsname{0.0938}
\expandafter\def\csname plaid@2D_profile@32@bestname\endcsname{probe\_kumo[medium,e4,v0.99]}
\expandafter\def\csname plaid@2D_profile@32@floor\endcsname{0.0604}
\expandafter\def\csname plaid@2D_profile@32@ratio\endcsname{0.898}
\expandafter\def\csname plaid@2D_profile@32@onfront\endcsname{yes}
\expandafter\def\csname plaid@2D_profile@32@onfronttotal\endcsname{yes}
\expandafter\def\csname plaid@2D_profile@32@onfrontqten\endcsname{yes}
\expandafter\def\csname plaid@2D_profile@32@onfrontqhundred\endcsname{yes}
\expandafter\def\csname plaid@2D_profile@32@beatsnn\endcsname{yes}
\expandafter\def\csname plaid@2D_profile@32@mostaccurate\endcsname{yes}
\expandafter\def\csname plaid@2D_profile@32@fit\endcsname{17.1}
\expandafter\def\csname plaid@2D_profile@32@query\endcsname{0.892}
\expandafter\def\csname plaid@2D_profile@32@bestfit\endcsname{11.7}
\expandafter\def\csname plaid@2D_profile@32@bestquery\endcsname{4.95}
\expandafter\def\csname plaid@2D_profile@32@coverage\endcsname{0.917}
\expandafter\def\csname plaid@2D_profile@32@scalars\endcsname{0}
\expandafter\def\csname plaid@2D_profile@32@covfmin\endcsname{0.883}
\expandafter\def\csname plaid@2D_profile@32@covfmax\endcsname{0.943}
\expandafter\def\csname plaid@2D_profile@32@r2\endcsname{--}
\expandafter\def\csname plaid@2D_profile@32@bestr2\endcsname{--}
\expandafter\def\csname plaid@2D_profile@32@r2ahead\endcsname{0}
\expandafter\def\csname plaid@2D_profile@32@knodes\endcsname{37}
\expandafter\def\csname plaid@2D_profile@32@peak\endcsname{2.99}
\expandafter\def\csname plaid@2D_MSHE@32@physicl\endcsname{0.457}
\expandafter\def\csname plaid@2D_MSHE@32@nn\endcsname{1.19}
\expandafter\def\csname plaid@2D_MSHE@32@spread\endcsname{0.0229}
\expandafter\def\csname plaid@2D_MSHE@32@spreadpct\endcsname{5}
\expandafter\def\csname plaid@2D_MSHE@32@draws\endcsname{3}
\expandafter\def\csname plaid@2D_MSHE@32@mean\endcsname{0.901}
\expandafter\def\csname plaid@2D_MSHE@32@best\endcsname{0.553}
\expandafter\def\csname plaid@2D_MSHE@32@bestname\endcsname{probe\_kumo[large,e1,v0.99]}
\expandafter\def\csname plaid@2D_MSHE@32@floor\endcsname{0.414}
\expandafter\def\csname plaid@2D_MSHE@32@ratio\endcsname{0.826}
\expandafter\def\csname plaid@2D_MSHE@32@onfront\endcsname{yes}
\expandafter\def\csname plaid@2D_MSHE@32@onfronttotal\endcsname{yes}
\expandafter\def\csname plaid@2D_MSHE@32@onfrontqten\endcsname{yes}
\expandafter\def\csname plaid@2D_MSHE@32@onfrontqhundred\endcsname{yes}
\expandafter\def\csname plaid@2D_MSHE@32@beatsnn\endcsname{yes}
\expandafter\def\csname plaid@2D_MSHE@32@mostaccurate\endcsname{yes}
\expandafter\def\csname plaid@2D_MSHE@32@fit\endcsname{15.9}
\expandafter\def\csname plaid@2D_MSHE@32@query\endcsname{0.752}
\expandafter\def\csname plaid@2D_MSHE@32@bestfit\endcsname{14.2}
\expandafter\def\csname plaid@2D_MSHE@32@bestquery\endcsname{5.76}
\expandafter\def\csname plaid@2D_MSHE@32@coverage\endcsname{0.895}
\expandafter\def\csname plaid@2D_MSHE@32@scalars\endcsname{1}
\expandafter\def\csname plaid@2D_MSHE@32@covfmin\endcsname{0.829}
\expandafter\def\csname plaid@2D_MSHE@32@covfmax\endcsname{0.942}
\expandafter\def\csname plaid@2D_MSHE@32@r2\endcsname{0.366}
\expandafter\def\csname plaid@2D_MSHE@32@bestr2\endcsname{0.783}
\expandafter\def\csname plaid@2D_MSHE@32@r2ahead\endcsname{0}
\expandafter\def\csname plaid@2D_MSHE@32@knodes\endcsname{6}
\expandafter\def\csname plaid@2D_MSHE@32@peak\endcsname{2.78}
\expandafter\def\csname plaid@Tensile2d@32@physicl\endcsname{0.000874}
\expandafter\def\csname plaid@Tensile2d@32@nn\endcsname{0.023}
\expandafter\def\csname plaid@Tensile2d@32@spread\endcsname{0.000166}
\expandafter\def\csname plaid@Tensile2d@32@spreadpct\endcsname{19}
\expandafter\def\csname plaid@Tensile2d@32@draws\endcsname{3}
\expandafter\def\csname plaid@Tensile2d@32@mean\endcsname{0.514}
\expandafter\def\csname plaid@Tensile2d@32@best\endcsname{6.84e-05}
\expandafter\def\csname plaid@Tensile2d@32@bestname\endcsname{probe\_kumo[large,e1,v0.9999]}
\expandafter\def\csname plaid@Tensile2d@32@floor\endcsname{6.8e-05}
\expandafter\def\csname plaid@Tensile2d@32@ratio\endcsname{12.8}
\expandafter\def\csname plaid@Tensile2d@32@onfront\endcsname{no}
\expandafter\def\csname plaid@Tensile2d@32@onfronttotal\endcsname{yes}
\expandafter\def\csname plaid@Tensile2d@32@onfrontqten\endcsname{yes}
\expandafter\def\csname plaid@Tensile2d@32@onfrontqhundred\endcsname{no}
\expandafter\def\csname plaid@Tensile2d@32@beatsnn\endcsname{yes}
\expandafter\def\csname plaid@Tensile2d@32@mostaccurate\endcsname{no}
\expandafter\def\csname plaid@Tensile2d@32@fit\endcsname{16.2}
\expandafter\def\csname plaid@Tensile2d@32@query\endcsname{0.668}
\expandafter\def\csname plaid@Tensile2d@32@bestfit\endcsname{7.01}
\expandafter\def\csname plaid@Tensile2d@32@bestquery\endcsname{1.73}
\expandafter\def\csname plaid@Tensile2d@32@coverage\endcsname{0.944}
\expandafter\def\csname plaid@Tensile2d@32@scalars\endcsname{3}
\expandafter\def\csname plaid@Tensile2d@32@covfmin\endcsname{0.919}
\expandafter\def\csname plaid@Tensile2d@32@covfmax\endcsname{0.963}
\expandafter\def\csname plaid@Tensile2d@32@r2\endcsname{0.951}
\expandafter\def\csname plaid@Tensile2d@32@bestr2\endcsname{0.996}
\expandafter\def\csname plaid@Tensile2d@32@r2ahead\endcsname{0}
\expandafter\def\csname plaid@Tensile2d@32@knodes\endcsname{10}
\expandafter\def\csname plaid@Tensile2d@32@peak\endcsname{2.8}
\expandafter\def\csname plaid@VKI-LS59@32@physicl\endcsname{0.067}
\expandafter\def\csname plaid@VKI-LS59@32@nn\endcsname{0.106}
\expandafter\def\csname plaid@VKI-LS59@32@spread\endcsname{0.0151}
\expandafter\def\csname plaid@VKI-LS59@32@spreadpct\endcsname{23}
\expandafter\def\csname plaid@VKI-LS59@32@draws\endcsname{3}
\expandafter\def\csname plaid@VKI-LS59@32@mean\endcsname{0.527}
\expandafter\def\csname plaid@VKI-LS59@32@best\endcsname{0.0587}
\expandafter\def\csname plaid@VKI-LS59@32@bestname\endcsname{pca\_kumo[large,e1,v0.999]}
\expandafter\def\csname plaid@VKI-LS59@32@floor\endcsname{0.0041}
\expandafter\def\csname plaid@VKI-LS59@32@ratio\endcsname{1.14}
\expandafter\def\csname plaid@VKI-LS59@32@onfront\endcsname{no}
\expandafter\def\csname plaid@VKI-LS59@32@onfronttotal\endcsname{no}
\expandafter\def\csname plaid@VKI-LS59@32@onfrontqten\endcsname{no}
\expandafter\def\csname plaid@VKI-LS59@32@onfrontqhundred\endcsname{no}
\expandafter\def\csname plaid@VKI-LS59@32@beatsnn\endcsname{yes}
\expandafter\def\csname plaid@VKI-LS59@32@mostaccurate\endcsname{no}
\expandafter\def\csname plaid@VKI-LS59@32@fit\endcsname{16.5}
\expandafter\def\csname plaid@VKI-LS59@32@query\endcsname{1.2}
\expandafter\def\csname plaid@VKI-LS59@32@bestfit\endcsname{4.8}
\expandafter\def\csname plaid@VKI-LS59@32@bestquery\endcsname{1.32}
\expandafter\def\csname plaid@VKI-LS59@32@coverage\endcsname{0.93}
\expandafter\def\csname plaid@VKI-LS59@32@scalars\endcsname{6}
\expandafter\def\csname plaid@VKI-LS59@32@covfmin\endcsname{0.911}
\expandafter\def\csname plaid@VKI-LS59@32@covfmax\endcsname{0.949}
\expandafter\def\csname plaid@VKI-LS59@32@r2\endcsname{0.563}
\expandafter\def\csname plaid@VKI-LS59@32@bestr2\endcsname{0.901}
\expandafter\def\csname plaid@VKI-LS59@32@r2ahead\endcsname{0}
\expandafter\def\csname plaid@VKI-LS59@32@knodes\endcsname{0}
\expandafter\def\csname plaid@VKI-LS59@32@peak\endcsname{2.86}
\expandafter\def\csname plaid@Rotor37@32@physicl\endcsname{0.0261}
\expandafter\def\csname plaid@Rotor37@32@nn\endcsname{0.0301}
\expandafter\def\csname plaid@Rotor37@32@spread\endcsname{0.00103}
\expandafter\def\csname plaid@Rotor37@32@spreadpct\endcsname{4}
\expandafter\def\csname plaid@Rotor37@32@draws\endcsname{3}
\expandafter\def\csname plaid@Rotor37@32@mean\endcsname{0.999}
\expandafter\def\csname plaid@Rotor37@32@best\endcsname{0.00601}
\expandafter\def\csname plaid@Rotor37@32@bestname\endcsname{pca\_kumo[large,e4,v0.999]}
\expandafter\def\csname plaid@Rotor37@32@floor\endcsname{0.0014}
\expandafter\def\csname plaid@Rotor37@32@ratio\endcsname{4.34}
\expandafter\def\csname plaid@Rotor37@32@onfront\endcsname{no}
\expandafter\def\csname plaid@Rotor37@32@onfronttotal\endcsname{no}
\expandafter\def\csname plaid@Rotor37@32@onfrontqten\endcsname{no}
\expandafter\def\csname plaid@Rotor37@32@onfrontqhundred\endcsname{no}
\expandafter\def\csname plaid@Rotor37@32@beatsnn\endcsname{yes}
\expandafter\def\csname plaid@Rotor37@32@mostaccurate\endcsname{no}
\expandafter\def\csname plaid@Rotor37@32@fit\endcsname{18}
\expandafter\def\csname plaid@Rotor37@32@query\endcsname{0.898}
\expandafter\def\csname plaid@Rotor37@32@bestfit\endcsname{9.37}
\expandafter\def\csname plaid@Rotor37@32@bestquery\endcsname{2.71}
\expandafter\def\csname plaid@Rotor37@32@coverage\endcsname{0.911}
\expandafter\def\csname plaid@Rotor37@32@scalars\endcsname{3}
\expandafter\def\csname plaid@Rotor37@32@covfmin\endcsname{0.91}
\expandafter\def\csname plaid@Rotor37@32@covfmax\endcsname{0.913}
\expandafter\def\csname plaid@Rotor37@32@r2\endcsname{0.785}
\expandafter\def\csname plaid@Rotor37@32@bestr2\endcsname{0.92}
\expandafter\def\csname plaid@Rotor37@32@r2ahead\endcsname{0}
\expandafter\def\csname plaid@Rotor37@32@knodes\endcsname{30}
\expandafter\def\csname plaid@Rotor37@32@peak\endcsname{2.93}
\expandafter\def\csname plaid@all@32@onfront\endcsname{2}
\expandafter\def\csname plaid@all@32@onfronttotal\endcsname{3}
\expandafter\def\csname plaid@all@32@onfrontqten\endcsname{3}
\expandafter\def\csname plaid@all@32@onfrontqhundred\endcsname{2}
\expandafter\def\csname plaid@all@32@mostaccurate\endcsname{2}
\expandafter\def\csname plaid@all@32@beatsnn\endcsname{5}
\expandafter\def\csname plaid@all@32@datasets\endcsname{5}
\expandafter\def\csname plaid@all@32@scalars\endcsname{13}
\expandafter\def\csname plaid@all@32@r2ahead\endcsname{0}
\expandafter\def\csname plaid@all@32@covmin\endcsname{0.895}
\expandafter\def\csname plaid@all@32@covmax\endcsname{0.944}
\expandafter\def\csname plaid@all@32@covfieldmin\endcsname{0.829}
\expandafter\def\csname plaid@all@32@covfieldmax\endcsname{0.963}
\expandafter\def\csname plaid@2D_profile@64@physicl\endcsname{0.064}
\expandafter\def\csname plaid@2D_profile@64@nn\endcsname{0.144}
\expandafter\def\csname plaid@2D_profile@64@spread\endcsname{0.0065}
\expandafter\def\csname plaid@2D_profile@64@spreadpct\endcsname{10}
\expandafter\def\csname plaid@2D_profile@64@draws\endcsname{3}
\expandafter\def\csname plaid@2D_profile@64@mean\endcsname{0.941}
\expandafter\def\csname plaid@2D_profile@64@best\endcsname{0.0693}
\expandafter\def\csname plaid@2D_profile@64@bestname\endcsname{probe\_kumo[medium,e8,v0.99]}
\expandafter\def\csname plaid@2D_profile@64@floor\endcsname{0.0416}
\expandafter\def\csname plaid@2D_profile@64@ratio\endcsname{0.923}
\expandafter\def\csname plaid@2D_profile@64@onfront\endcsname{yes}
\expandafter\def\csname plaid@2D_profile@64@onfronttotal\endcsname{yes}
\expandafter\def\csname plaid@2D_profile@64@onfrontqten\endcsname{yes}
\expandafter\def\csname plaid@2D_profile@64@onfrontqhundred\endcsname{yes}
\expandafter\def\csname plaid@2D_profile@64@beatsnn\endcsname{yes}
\expandafter\def\csname plaid@2D_profile@64@mostaccurate\endcsname{yes}
\expandafter\def\csname plaid@2D_profile@64@fit\endcsname{17.5}
\expandafter\def\csname plaid@2D_profile@64@query\endcsname{0.895}
\expandafter\def\csname plaid@2D_profile@64@bestfit\endcsname{22.3}
\expandafter\def\csname plaid@2D_profile@64@bestquery\endcsname{8.87}
\expandafter\def\csname plaid@2D_profile@64@coverage\endcsname{0.908}
\expandafter\def\csname plaid@2D_profile@64@scalars\endcsname{0}
\expandafter\def\csname plaid@2D_profile@64@covfmin\endcsname{0.868}
\expandafter\def\csname plaid@2D_profile@64@covfmax\endcsname{0.934}
\expandafter\def\csname plaid@2D_profile@64@r2\endcsname{--}
\expandafter\def\csname plaid@2D_profile@64@bestr2\endcsname{--}
\expandafter\def\csname plaid@2D_profile@64@r2ahead\endcsname{0}
\expandafter\def\csname plaid@2D_profile@64@knodes\endcsname{37}
\expandafter\def\csname plaid@2D_profile@64@peak\endcsname{4.51}
\expandafter\def\csname plaid@2D_MSHE@64@physicl\endcsname{0.44}
\expandafter\def\csname plaid@2D_MSHE@64@nn\endcsname{1.16}
\expandafter\def\csname plaid@2D_MSHE@64@spread\endcsname{0.0182}
\expandafter\def\csname plaid@2D_MSHE@64@spreadpct\endcsname{4}
\expandafter\def\csname plaid@2D_MSHE@64@draws\endcsname{3}
\expandafter\def\csname plaid@2D_MSHE@64@mean\endcsname{0.913}
\expandafter\def\csname plaid@2D_MSHE@64@best\endcsname{0.48}
\expandafter\def\csname plaid@2D_MSHE@64@bestname\endcsname{probe\_kumo[large,e8,v0.99]}
\expandafter\def\csname plaid@2D_MSHE@64@floor\endcsname{0.319}
\expandafter\def\csname plaid@2D_MSHE@64@ratio\endcsname{0.916}
\expandafter\def\csname plaid@2D_MSHE@64@onfront\endcsname{yes}
\expandafter\def\csname plaid@2D_MSHE@64@onfronttotal\endcsname{yes}
\expandafter\def\csname plaid@2D_MSHE@64@onfrontqten\endcsname{yes}
\expandafter\def\csname plaid@2D_MSHE@64@onfrontqhundred\endcsname{yes}
\expandafter\def\csname plaid@2D_MSHE@64@beatsnn\endcsname{yes}
\expandafter\def\csname plaid@2D_MSHE@64@mostaccurate\endcsname{yes}
\expandafter\def\csname plaid@2D_MSHE@64@fit\endcsname{16.6}
\expandafter\def\csname plaid@2D_MSHE@64@query\endcsname{0.803}
\expandafter\def\csname plaid@2D_MSHE@64@bestfit\endcsname{63.4}
\expandafter\def\csname plaid@2D_MSHE@64@bestquery\endcsname{15.1}
\expandafter\def\csname plaid@2D_MSHE@64@coverage\endcsname{0.888}
\expandafter\def\csname plaid@2D_MSHE@64@scalars\endcsname{1}
\expandafter\def\csname plaid@2D_MSHE@64@covfmin\endcsname{0.796}
\expandafter\def\csname plaid@2D_MSHE@64@covfmax\endcsname{0.936}
\expandafter\def\csname plaid@2D_MSHE@64@r2\endcsname{0.433}
\expandafter\def\csname plaid@2D_MSHE@64@bestr2\endcsname{0.922}
\expandafter\def\csname plaid@2D_MSHE@64@r2ahead\endcsname{0}
\expandafter\def\csname plaid@2D_MSHE@64@knodes\endcsname{6}
\expandafter\def\csname plaid@2D_MSHE@64@peak\endcsname{4.07}
\expandafter\def\csname plaid@Tensile2d@64@physicl\endcsname{0.000504}
\expandafter\def\csname plaid@Tensile2d@64@nn\endcsname{0.0193}
\expandafter\def\csname plaid@Tensile2d@64@spread\endcsname{9.04e-05}
\expandafter\def\csname plaid@Tensile2d@64@spreadpct\endcsname{18}
\expandafter\def\csname plaid@Tensile2d@64@draws\endcsname{3}
\expandafter\def\csname plaid@Tensile2d@64@mean\endcsname{0.583}
\expandafter\def\csname plaid@Tensile2d@64@best\endcsname{3.94e-05}
\expandafter\def\csname plaid@Tensile2d@64@bestname\endcsname{probe\_kumo[large,e1,v0.9999]}
\expandafter\def\csname plaid@Tensile2d@64@floor\endcsname{4.15e-05}
\expandafter\def\csname plaid@Tensile2d@64@ratio\endcsname{12.8}
\expandafter\def\csname plaid@Tensile2d@64@onfront\endcsname{no}
\expandafter\def\csname plaid@Tensile2d@64@onfronttotal\endcsname{yes}
\expandafter\def\csname plaid@Tensile2d@64@onfrontqten\endcsname{yes}
\expandafter\def\csname plaid@Tensile2d@64@onfrontqhundred\endcsname{no}
\expandafter\def\csname plaid@Tensile2d@64@beatsnn\endcsname{yes}
\expandafter\def\csname plaid@Tensile2d@64@mostaccurate\endcsname{no}
\expandafter\def\csname plaid@Tensile2d@64@fit\endcsname{17.9}
\expandafter\def\csname plaid@Tensile2d@64@query\endcsname{0.648}
\expandafter\def\csname plaid@Tensile2d@64@bestfit\endcsname{12.3}
\expandafter\def\csname plaid@Tensile2d@64@bestquery\endcsname{2.73}
\expandafter\def\csname plaid@Tensile2d@64@coverage\endcsname{0.957}
\expandafter\def\csname plaid@Tensile2d@64@scalars\endcsname{3}
\expandafter\def\csname plaid@Tensile2d@64@covfmin\endcsname{0.938}
\expandafter\def\csname plaid@Tensile2d@64@covfmax\endcsname{0.975}
\expandafter\def\csname plaid@Tensile2d@64@r2\endcsname{0.978}
\expandafter\def\csname plaid@Tensile2d@64@bestr2\endcsname{0.997}
\expandafter\def\csname plaid@Tensile2d@64@r2ahead\endcsname{0}
\expandafter\def\csname plaid@Tensile2d@64@knodes\endcsname{10}
\expandafter\def\csname plaid@Tensile2d@64@peak\endcsname{4.27}
\expandafter\def\csname plaid@VKI-LS59@64@physicl\endcsname{0.0536}
\expandafter\def\csname plaid@VKI-LS59@64@nn\endcsname{0.103}
\expandafter\def\csname plaid@VKI-LS59@64@spread\endcsname{0.0133}
\expandafter\def\csname plaid@VKI-LS59@64@spreadpct\endcsname{25}
\expandafter\def\csname plaid@VKI-LS59@64@draws\endcsname{3}
\expandafter\def\csname plaid@VKI-LS59@64@mean\endcsname{0.56}
\expandafter\def\csname plaid@VKI-LS59@64@best\endcsname{0.0337}
\expandafter\def\csname plaid@VKI-LS59@64@bestname\endcsname{pca\_kumo[large,e8,v0.999]}
\expandafter\def\csname plaid@VKI-LS59@64@floor\endcsname{0.00149}
\expandafter\def\csname plaid@VKI-LS59@64@ratio\endcsname{1.59}
\expandafter\def\csname plaid@VKI-LS59@64@onfront\endcsname{yes}
\expandafter\def\csname plaid@VKI-LS59@64@onfronttotal\endcsname{yes}
\expandafter\def\csname plaid@VKI-LS59@64@onfrontqten\endcsname{yes}
\expandafter\def\csname plaid@VKI-LS59@64@onfrontqhundred\endcsname{yes}
\expandafter\def\csname plaid@VKI-LS59@64@beatsnn\endcsname{yes}
\expandafter\def\csname plaid@VKI-LS59@64@mostaccurate\endcsname{no}
\expandafter\def\csname plaid@VKI-LS59@64@fit\endcsname{17.6}
\expandafter\def\csname plaid@VKI-LS59@64@query\endcsname{0.983}
\expandafter\def\csname plaid@VKI-LS59@64@bestfit\endcsname{14.5}
\expandafter\def\csname plaid@VKI-LS59@64@bestquery\endcsname{5.91}
\expandafter\def\csname plaid@VKI-LS59@64@coverage\endcsname{0.954}
\expandafter\def\csname plaid@VKI-LS59@64@scalars\endcsname{6}
\expandafter\def\csname plaid@VKI-LS59@64@covfmin\endcsname{0.943}
\expandafter\def\csname plaid@VKI-LS59@64@covfmax\endcsname{0.966}
\expandafter\def\csname plaid@VKI-LS59@64@r2\endcsname{0.866}
\expandafter\def\csname plaid@VKI-LS59@64@bestr2\endcsname{0.973}
\expandafter\def\csname plaid@VKI-LS59@64@r2ahead\endcsname{0}
\expandafter\def\csname plaid@VKI-LS59@64@knodes\endcsname{0}
\expandafter\def\csname plaid@VKI-LS59@64@peak\endcsname{4}
\expandafter\def\csname plaid@Rotor37@64@physicl\endcsname{0.0228}
\expandafter\def\csname plaid@Rotor37@64@nn\endcsname{0.0292}
\expandafter\def\csname plaid@Rotor37@64@spread\endcsname{0.000822}
\expandafter\def\csname plaid@Rotor37@64@spreadpct\endcsname{4}
\expandafter\def\csname plaid@Rotor37@64@draws\endcsname{3}
\expandafter\def\csname plaid@Rotor37@64@mean\endcsname{0.987}
\expandafter\def\csname plaid@Rotor37@64@best\endcsname{0.00331}
\expandafter\def\csname plaid@Rotor37@64@bestname\endcsname{pca\_kumo[medium,e8,v0.999]}
\expandafter\def\csname plaid@Rotor37@64@floor\endcsname{0.000497}
\expandafter\def\csname plaid@Rotor37@64@ratio\endcsname{6.9}
\expandafter\def\csname plaid@Rotor37@64@onfront\endcsname{no}
\expandafter\def\csname plaid@Rotor37@64@onfronttotal\endcsname{yes}
\expandafter\def\csname plaid@Rotor37@64@onfrontqten\endcsname{yes}
\expandafter\def\csname plaid@Rotor37@64@onfrontqhundred\endcsname{no}
\expandafter\def\csname plaid@Rotor37@64@beatsnn\endcsname{yes}
\expandafter\def\csname plaid@Rotor37@64@mostaccurate\endcsname{no}
\expandafter\def\csname plaid@Rotor37@64@fit\endcsname{18.2}
\expandafter\def\csname plaid@Rotor37@64@query\endcsname{1.28}
\expandafter\def\csname plaid@Rotor37@64@bestfit\endcsname{23.4}
\expandafter\def\csname plaid@Rotor37@64@bestquery\endcsname{6.46}
\expandafter\def\csname plaid@Rotor37@64@coverage\endcsname{0.926}
\expandafter\def\csname plaid@Rotor37@64@scalars\endcsname{3}
\expandafter\def\csname plaid@Rotor37@64@covfmin\endcsname{0.912}
\expandafter\def\csname plaid@Rotor37@64@covfmax\endcsname{0.933}
\expandafter\def\csname plaid@Rotor37@64@r2\endcsname{0.86}
\expandafter\def\csname plaid@Rotor37@64@bestr2\endcsname{0.936}
\expandafter\def\csname plaid@Rotor37@64@r2ahead\endcsname{0}
\expandafter\def\csname plaid@Rotor37@64@knodes\endcsname{30}
\expandafter\def\csname plaid@Rotor37@64@peak\endcsname{3.98}
\expandafter\def\csname plaid@all@64@onfront\endcsname{3}
\expandafter\def\csname plaid@all@64@onfronttotal\endcsname{5}
\expandafter\def\csname plaid@all@64@onfrontqten\endcsname{5}
\expandafter\def\csname plaid@all@64@onfrontqhundred\endcsname{3}
\expandafter\def\csname plaid@all@64@mostaccurate\endcsname{2}
\expandafter\def\csname plaid@all@64@beatsnn\endcsname{5}
\expandafter\def\csname plaid@all@64@datasets\endcsname{5}
\expandafter\def\csname plaid@all@64@scalars\endcsname{13}
\expandafter\def\csname plaid@all@64@r2ahead\endcsname{0}
\expandafter\def\csname plaid@all@64@covmin\endcsname{0.888}
\expandafter\def\csname plaid@all@64@covmax\endcsname{0.957}
\expandafter\def\csname plaid@all@64@covfieldmin\endcsname{0.796}
\expandafter\def\csname plaid@all@64@covfieldmax\endcsname{0.975}
\expandafter\def\csname plaid@2D_profile@125@physicl\endcsname{0.0512}
\expandafter\def\csname plaid@2D_profile@125@nn\endcsname{0.109}
\expandafter\def\csname plaid@2D_profile@125@spread\endcsname{0.00375}
\expandafter\def\csname plaid@2D_profile@125@spreadpct\endcsname{7}
\expandafter\def\csname plaid@2D_profile@125@draws\endcsname{3}
\expandafter\def\csname plaid@2D_profile@125@mean\endcsname{0.897}
\expandafter\def\csname plaid@2D_profile@125@best\endcsname{0.0486}
\expandafter\def\csname plaid@2D_profile@125@bestname\endcsname{probe\_kumo[medium,e8,v0.99]}
\expandafter\def\csname plaid@2D_profile@125@floor\endcsname{0.0298}
\expandafter\def\csname plaid@2D_profile@125@ratio\endcsname{1.05}
\expandafter\def\csname plaid@2D_profile@125@onfront\endcsname{yes}
\expandafter\def\csname plaid@2D_profile@125@onfronttotal\endcsname{yes}
\expandafter\def\csname plaid@2D_profile@125@onfrontqten\endcsname{yes}
\expandafter\def\csname plaid@2D_profile@125@onfrontqhundred\endcsname{yes}
\expandafter\def\csname plaid@2D_profile@125@beatsnn\endcsname{yes}
\expandafter\def\csname plaid@2D_profile@125@mostaccurate\endcsname{yes}
\expandafter\def\csname plaid@2D_profile@125@fit\endcsname{19.2}
\expandafter\def\csname plaid@2D_profile@125@query\endcsname{0.87}
\expandafter\def\csname plaid@2D_profile@125@bestfit\endcsname{38.6}
\expandafter\def\csname plaid@2D_profile@125@bestquery\endcsname{16.6}
\expandafter\def\csname plaid@2D_profile@125@coverage\endcsname{0.921}
\expandafter\def\csname plaid@2D_profile@125@scalars\endcsname{0}
\expandafter\def\csname plaid@2D_profile@125@covfmin\endcsname{0.907}
\expandafter\def\csname plaid@2D_profile@125@covfmax\endcsname{0.934}
\expandafter\def\csname plaid@2D_profile@125@r2\endcsname{--}
\expandafter\def\csname plaid@2D_profile@125@bestr2\endcsname{--}
\expandafter\def\csname plaid@2D_profile@125@r2ahead\endcsname{0}
\expandafter\def\csname plaid@2D_profile@125@knodes\endcsname{37}
\expandafter\def\csname plaid@2D_profile@125@peak\endcsname{7.61}
\expandafter\def\csname plaid@2D_MSHE@125@physicl\endcsname{0.418}
\expandafter\def\csname plaid@2D_MSHE@125@nn\endcsname{1.17}
\expandafter\def\csname plaid@2D_MSHE@125@spread\endcsname{0.00808}
\expandafter\def\csname plaid@2D_MSHE@125@spreadpct\endcsname{2}
\expandafter\def\csname plaid@2D_MSHE@125@draws\endcsname{3}
\expandafter\def\csname plaid@2D_MSHE@125@mean\endcsname{0.922}
\expandafter\def\csname plaid@2D_MSHE@125@best\endcsname{0.433}
\expandafter\def\csname plaid@2D_MSHE@125@bestname\endcsname{probe\_kumo[large,e8,v0.99]}
\expandafter\def\csname plaid@2D_MSHE@125@floor\endcsname{0.232}
\expandafter\def\csname plaid@2D_MSHE@125@ratio\endcsname{0.965}
\expandafter\def\csname plaid@2D_MSHE@125@onfront\endcsname{yes}
\expandafter\def\csname plaid@2D_MSHE@125@onfronttotal\endcsname{yes}
\expandafter\def\csname plaid@2D_MSHE@125@onfrontqten\endcsname{yes}
\expandafter\def\csname plaid@2D_MSHE@125@onfrontqhundred\endcsname{yes}
\expandafter\def\csname plaid@2D_MSHE@125@beatsnn\endcsname{yes}
\expandafter\def\csname plaid@2D_MSHE@125@mostaccurate\endcsname{yes}
\expandafter\def\csname plaid@2D_MSHE@125@fit\endcsname{18.2}
\expandafter\def\csname plaid@2D_MSHE@125@query\endcsname{0.834}
\expandafter\def\csname plaid@2D_MSHE@125@bestfit\endcsname{152}
\expandafter\def\csname plaid@2D_MSHE@125@bestquery\endcsname{41.9}
\expandafter\def\csname plaid@2D_MSHE@125@coverage\endcsname{0.891}
\expandafter\def\csname plaid@2D_MSHE@125@scalars\endcsname{1}
\expandafter\def\csname plaid@2D_MSHE@125@covfmin\endcsname{0.805}
\expandafter\def\csname plaid@2D_MSHE@125@covfmax\endcsname{0.924}
\expandafter\def\csname plaid@2D_MSHE@125@r2\endcsname{0.509}
\expandafter\def\csname plaid@2D_MSHE@125@bestr2\endcsname{0.946}
\expandafter\def\csname plaid@2D_MSHE@125@r2ahead\endcsname{0}
\expandafter\def\csname plaid@2D_MSHE@125@knodes\endcsname{6}
\expandafter\def\csname plaid@2D_MSHE@125@peak\endcsname{6.73}
\expandafter\def\csname plaid@Tensile2d@125@physicl\endcsname{0.000436}
\expandafter\def\csname plaid@Tensile2d@125@nn\endcsname{0.0212}
\expandafter\def\csname plaid@Tensile2d@125@spread\endcsname{5.68e-05}
\expandafter\def\csname plaid@Tensile2d@125@spreadpct\endcsname{13}
\expandafter\def\csname plaid@Tensile2d@125@draws\endcsname{3}
\expandafter\def\csname plaid@Tensile2d@125@mean\endcsname{0.561}
\expandafter\def\csname plaid@Tensile2d@125@best\endcsname{3.19e-05}
\expandafter\def\csname plaid@Tensile2d@125@bestname\endcsname{probe\_kumo[small,e8,v0.9999]}
\expandafter\def\csname plaid@Tensile2d@125@floor\endcsname{3.19e-05}
\expandafter\def\csname plaid@Tensile2d@125@ratio\endcsname{13.7}
\expandafter\def\csname plaid@Tensile2d@125@onfront\endcsname{no}
\expandafter\def\csname plaid@Tensile2d@125@onfronttotal\endcsname{yes}
\expandafter\def\csname plaid@Tensile2d@125@onfrontqten\endcsname{yes}
\expandafter\def\csname plaid@Tensile2d@125@onfrontqhundred\endcsname{no}
\expandafter\def\csname plaid@Tensile2d@125@beatsnn\endcsname{yes}
\expandafter\def\csname plaid@Tensile2d@125@mostaccurate\endcsname{no}
\expandafter\def\csname plaid@Tensile2d@125@fit\endcsname{18.2}
\expandafter\def\csname plaid@Tensile2d@125@query\endcsname{0.729}
\expandafter\def\csname plaid@Tensile2d@125@bestfit\endcsname{23.6}
\expandafter\def\csname plaid@Tensile2d@125@bestquery\endcsname{6.57}
\expandafter\def\csname plaid@Tensile2d@125@coverage\endcsname{0.949}
\expandafter\def\csname plaid@Tensile2d@125@scalars\endcsname{3}
\expandafter\def\csname plaid@Tensile2d@125@covfmin\endcsname{0.916}
\expandafter\def\csname plaid@Tensile2d@125@covfmax\endcsname{0.975}
\expandafter\def\csname plaid@Tensile2d@125@r2\endcsname{0.991}
\expandafter\def\csname plaid@Tensile2d@125@bestr2\endcsname{0.999}
\expandafter\def\csname plaid@Tensile2d@125@r2ahead\endcsname{1}
\expandafter\def\csname plaid@Tensile2d@125@knodes\endcsname{10}
\expandafter\def\csname plaid@Tensile2d@125@peak\endcsname{7.15}
\expandafter\def\csname plaid@VKI-LS59@125@physicl\endcsname{0.0503}
\expandafter\def\csname plaid@VKI-LS59@125@nn\endcsname{0.102}
\expandafter\def\csname plaid@VKI-LS59@125@spread\endcsname{0.00784}
\expandafter\def\csname plaid@VKI-LS59@125@spreadpct\endcsname{16}
\expandafter\def\csname plaid@VKI-LS59@125@draws\endcsname{3}
\expandafter\def\csname plaid@VKI-LS59@125@mean\endcsname{0.581}
\expandafter\def\csname plaid@VKI-LS59@125@best\endcsname{0.0115}
\expandafter\def\csname plaid@VKI-LS59@125@bestname\endcsname{pca\_kumo[large,e8,v0.999]}
\expandafter\def\csname plaid@VKI-LS59@125@floor\endcsname{0.00126}
\expandafter\def\csname plaid@VKI-LS59@125@ratio\endcsname{4.39}
\expandafter\def\csname plaid@VKI-LS59@125@onfront\endcsname{yes}
\expandafter\def\csname plaid@VKI-LS59@125@onfronttotal\endcsname{yes}
\expandafter\def\csname plaid@VKI-LS59@125@onfrontqten\endcsname{yes}
\expandafter\def\csname plaid@VKI-LS59@125@onfrontqhundred\endcsname{yes}
\expandafter\def\csname plaid@VKI-LS59@125@beatsnn\endcsname{yes}
\expandafter\def\csname plaid@VKI-LS59@125@mostaccurate\endcsname{no}
\expandafter\def\csname plaid@VKI-LS59@125@fit\endcsname{19.6}
\expandafter\def\csname plaid@VKI-LS59@125@query\endcsname{0.98}
\expandafter\def\csname plaid@VKI-LS59@125@bestfit\endcsname{24.3}
\expandafter\def\csname plaid@VKI-LS59@125@bestquery\endcsname{5.76}
\expandafter\def\csname plaid@VKI-LS59@125@coverage\endcsname{0.933}
\expandafter\def\csname plaid@VKI-LS59@125@scalars\endcsname{6}
\expandafter\def\csname plaid@VKI-LS59@125@covfmin\endcsname{0.908}
\expandafter\def\csname plaid@VKI-LS59@125@covfmax\endcsname{0.958}
\expandafter\def\csname plaid@VKI-LS59@125@r2\endcsname{0.937}
\expandafter\def\csname plaid@VKI-LS59@125@bestr2\endcsname{0.987}
\expandafter\def\csname plaid@VKI-LS59@125@r2ahead\endcsname{0}
\expandafter\def\csname plaid@VKI-LS59@125@knodes\endcsname{0}
\expandafter\def\csname plaid@VKI-LS59@125@peak\endcsname{6.6}
\expandafter\def\csname plaid@Rotor37@125@physicl\endcsname{0.0206}
\expandafter\def\csname plaid@Rotor37@125@nn\endcsname{0.0275}
\expandafter\def\csname plaid@Rotor37@125@spread\endcsname{0.000521}
\expandafter\def\csname plaid@Rotor37@125@spreadpct\endcsname{3}
\expandafter\def\csname plaid@Rotor37@125@draws\endcsname{3}
\expandafter\def\csname plaid@Rotor37@125@mean\endcsname{0.985}
\expandafter\def\csname plaid@Rotor37@125@best\endcsname{0.00222}
\expandafter\def\csname plaid@Rotor37@125@bestname\endcsname{pca\_kumo[large,e1,v0.999]}
\expandafter\def\csname plaid@Rotor37@125@floor\endcsname{0.000308}
\expandafter\def\csname plaid@Rotor37@125@ratio\endcsname{9.25}
\expandafter\def\csname plaid@Rotor37@125@onfront\endcsname{no}
\expandafter\def\csname plaid@Rotor37@125@onfronttotal\endcsname{yes}
\expandafter\def\csname plaid@Rotor37@125@onfrontqten\endcsname{yes}
\expandafter\def\csname plaid@Rotor37@125@onfrontqhundred\endcsname{no}
\expandafter\def\csname plaid@Rotor37@125@beatsnn\endcsname{yes}
\expandafter\def\csname plaid@Rotor37@125@mostaccurate\endcsname{no}
\expandafter\def\csname plaid@Rotor37@125@fit\endcsname{21}
\expandafter\def\csname plaid@Rotor37@125@query\endcsname{0.855}
\expandafter\def\csname plaid@Rotor37@125@bestfit\endcsname{23.8}
\expandafter\def\csname plaid@Rotor37@125@bestquery\endcsname{4.73}
\expandafter\def\csname plaid@Rotor37@125@coverage\endcsname{0.921}
\expandafter\def\csname plaid@Rotor37@125@scalars\endcsname{3}
\expandafter\def\csname plaid@Rotor37@125@covfmin\endcsname{0.91}
\expandafter\def\csname plaid@Rotor37@125@covfmax\endcsname{0.928}
\expandafter\def\csname plaid@Rotor37@125@r2\endcsname{0.913}
\expandafter\def\csname plaid@Rotor37@125@bestr2\endcsname{0.967}
\expandafter\def\csname plaid@Rotor37@125@r2ahead\endcsname{0}
\expandafter\def\csname plaid@Rotor37@125@knodes\endcsname{30}
\expandafter\def\csname plaid@Rotor37@125@peak\endcsname{6.56}
\expandafter\def\csname plaid@all@125@onfront\endcsname{3}
\expandafter\def\csname plaid@all@125@onfronttotal\endcsname{5}
\expandafter\def\csname plaid@all@125@onfrontqten\endcsname{5}
\expandafter\def\csname plaid@all@125@onfrontqhundred\endcsname{3}
\expandafter\def\csname plaid@all@125@mostaccurate\endcsname{2}
\expandafter\def\csname plaid@all@125@beatsnn\endcsname{5}
\expandafter\def\csname plaid@all@125@datasets\endcsname{5}
\expandafter\def\csname plaid@all@125@scalars\endcsname{13}
\expandafter\def\csname plaid@all@125@r2ahead\endcsname{1}
\expandafter\def\csname plaid@all@125@covmin\endcsname{0.891}
\expandafter\def\csname plaid@all@125@covmax\endcsname{0.949}
\expandafter\def\csname plaid@all@125@covfieldmin\endcsname{0.805}
\expandafter\def\csname plaid@all@125@covfieldmax\endcsname{0.975}
\expandafter\def\csname plaid@2D_MSHE@250@physicl\endcsname{0.399}
\expandafter\def\csname plaid@2D_MSHE@250@nn\endcsname{1.12}
\expandafter\def\csname plaid@2D_MSHE@250@spread\endcsname{0.00531}
\expandafter\def\csname plaid@2D_MSHE@250@spreadpct\endcsname{1}
\expandafter\def\csname plaid@2D_MSHE@250@draws\endcsname{3}
\expandafter\def\csname plaid@2D_MSHE@250@mean\endcsname{0.934}
\expandafter\def\csname plaid@2D_MSHE@250@best\endcsname{0.38}
\expandafter\def\csname plaid@2D_MSHE@250@bestname\endcsname{probe\_kumo[large,e8,v0.99]}
\expandafter\def\csname plaid@2D_MSHE@250@floor\endcsname{0.151}
\expandafter\def\csname plaid@2D_MSHE@250@ratio\endcsname{1.05}
\expandafter\def\csname plaid@2D_MSHE@250@onfront\endcsname{yes}
\expandafter\def\csname plaid@2D_MSHE@250@onfronttotal\endcsname{yes}
\expandafter\def\csname plaid@2D_MSHE@250@onfrontqten\endcsname{yes}
\expandafter\def\csname plaid@2D_MSHE@250@onfrontqhundred\endcsname{yes}
\expandafter\def\csname plaid@2D_MSHE@250@beatsnn\endcsname{yes}
\expandafter\def\csname plaid@2D_MSHE@250@mostaccurate\endcsname{no}
\expandafter\def\csname plaid@2D_MSHE@250@fit\endcsname{21.5}
\expandafter\def\csname plaid@2D_MSHE@250@query\endcsname{1.06}
\expandafter\def\csname plaid@2D_MSHE@250@bestfit\endcsname{379}
\expandafter\def\csname plaid@2D_MSHE@250@bestquery\endcsname{69.2}
\expandafter\def\csname plaid@2D_MSHE@250@coverage\endcsname{0.901}
\expandafter\def\csname plaid@2D_MSHE@250@scalars\endcsname{1}
\expandafter\def\csname plaid@2D_MSHE@250@covfmin\endcsname{0.83}
\expandafter\def\csname plaid@2D_MSHE@250@covfmax\endcsname{0.936}
\expandafter\def\csname plaid@2D_MSHE@250@r2\endcsname{0.645}
\expandafter\def\csname plaid@2D_MSHE@250@bestr2\endcsname{0.969}
\expandafter\def\csname plaid@2D_MSHE@250@r2ahead\endcsname{0}
\expandafter\def\csname plaid@2D_MSHE@250@knodes\endcsname{6}
\expandafter\def\csname plaid@2D_MSHE@250@peak\endcsname{12.7}
\expandafter\def\csname plaid@Tensile2d@250@physicl\endcsname{0.00037}
\expandafter\def\csname plaid@Tensile2d@250@nn\endcsname{0.0164}
\expandafter\def\csname plaid@Tensile2d@250@spread\endcsname{2.63e-05}
\expandafter\def\csname plaid@Tensile2d@250@spreadpct\endcsname{7}
\expandafter\def\csname plaid@Tensile2d@250@draws\endcsname{3}
\expandafter\def\csname plaid@Tensile2d@250@mean\endcsname{0.553}
\expandafter\def\csname plaid@Tensile2d@250@best\endcsname{2.42e-05}
\expandafter\def\csname plaid@Tensile2d@250@bestname\endcsname{probe\_kumo[small,e8,v0.9999]}
\expandafter\def\csname plaid@Tensile2d@250@floor\endcsname{2.55e-05}
\expandafter\def\csname plaid@Tensile2d@250@ratio\endcsname{15.3}
\expandafter\def\csname plaid@Tensile2d@250@onfront\endcsname{no}
\expandafter\def\csname plaid@Tensile2d@250@onfronttotal\endcsname{yes}
\expandafter\def\csname plaid@Tensile2d@250@onfrontqten\endcsname{yes}
\expandafter\def\csname plaid@Tensile2d@250@onfrontqhundred\endcsname{no}
\expandafter\def\csname plaid@Tensile2d@250@beatsnn\endcsname{yes}
\expandafter\def\csname plaid@Tensile2d@250@mostaccurate\endcsname{no}
\expandafter\def\csname plaid@Tensile2d@250@fit\endcsname{21.6}
\expandafter\def\csname plaid@Tensile2d@250@query\endcsname{0.88}
\expandafter\def\csname plaid@Tensile2d@250@bestfit\endcsname{33.2}
\expandafter\def\csname plaid@Tensile2d@250@bestquery\endcsname{7.29}
\expandafter\def\csname plaid@Tensile2d@250@coverage\endcsname{0.949}
\expandafter\def\csname plaid@Tensile2d@250@scalars\endcsname{3}
\expandafter\def\csname plaid@Tensile2d@250@covfmin\endcsname{0.915}
\expandafter\def\csname plaid@Tensile2d@250@covfmax\endcsname{0.98}
\expandafter\def\csname plaid@Tensile2d@250@r2\endcsname{0.994}
\expandafter\def\csname plaid@Tensile2d@250@bestr2\endcsname{0.999}
\expandafter\def\csname plaid@Tensile2d@250@r2ahead\endcsname{2}
\expandafter\def\csname plaid@Tensile2d@250@knodes\endcsname{10}
\expandafter\def\csname plaid@Tensile2d@250@peak\endcsname{13}
\expandafter\def\csname plaid@VKI-LS59@250@physicl\endcsname{0.0452}
\expandafter\def\csname plaid@VKI-LS59@250@nn\endcsname{0.0649}
\expandafter\def\csname plaid@VKI-LS59@250@spread\endcsname{0.00382}
\expandafter\def\csname plaid@VKI-LS59@250@spreadpct\endcsname{8}
\expandafter\def\csname plaid@VKI-LS59@250@draws\endcsname{3}
\expandafter\def\csname plaid@VKI-LS59@250@mean\endcsname{0.577}
\expandafter\def\csname plaid@VKI-LS59@250@best\endcsname{0.00583}
\expandafter\def\csname plaid@VKI-LS59@250@bestname\endcsname{pca\_kumo[large,e8,v0.999]}
\expandafter\def\csname plaid@VKI-LS59@250@floor\endcsname{0.000736}
\expandafter\def\csname plaid@VKI-LS59@250@ratio\endcsname{7.74}
\expandafter\def\csname plaid@VKI-LS59@250@onfront\endcsname{yes}
\expandafter\def\csname plaid@VKI-LS59@250@onfronttotal\endcsname{yes}
\expandafter\def\csname plaid@VKI-LS59@250@onfrontqten\endcsname{yes}
\expandafter\def\csname plaid@VKI-LS59@250@onfrontqhundred\endcsname{yes}
\expandafter\def\csname plaid@VKI-LS59@250@beatsnn\endcsname{yes}
\expandafter\def\csname plaid@VKI-LS59@250@mostaccurate\endcsname{no}
\expandafter\def\csname plaid@VKI-LS59@250@fit\endcsname{23.2}
\expandafter\def\csname plaid@VKI-LS59@250@query\endcsname{1.23}
\expandafter\def\csname plaid@VKI-LS59@250@bestfit\endcsname{41.1}
\expandafter\def\csname plaid@VKI-LS59@250@bestquery\endcsname{11.9}
\expandafter\def\csname plaid@VKI-LS59@250@coverage\endcsname{0.931}
\expandafter\def\csname plaid@VKI-LS59@250@scalars\endcsname{6}
\expandafter\def\csname plaid@VKI-LS59@250@covfmin\endcsname{0.904}
\expandafter\def\csname plaid@VKI-LS59@250@covfmax\endcsname{0.959}
\expandafter\def\csname plaid@VKI-LS59@250@r2\endcsname{0.958}
\expandafter\def\csname plaid@VKI-LS59@250@bestr2\endcsname{0.993}
\expandafter\def\csname plaid@VKI-LS59@250@r2ahead\endcsname{0}
\expandafter\def\csname plaid@VKI-LS59@250@knodes\endcsname{0}
\expandafter\def\csname plaid@VKI-LS59@250@peak\endcsname{11.9}
\expandafter\def\csname plaid@Rotor37@250@physicl\endcsname{0.0193}
\expandafter\def\csname plaid@Rotor37@250@nn\endcsname{0.02}
\expandafter\def\csname plaid@Rotor37@250@spread\endcsname{0.000461}
\expandafter\def\csname plaid@Rotor37@250@spreadpct\endcsname{2}
\expandafter\def\csname plaid@Rotor37@250@draws\endcsname{3}
\expandafter\def\csname plaid@Rotor37@250@mean\endcsname{0.986}
\expandafter\def\csname plaid@Rotor37@250@best\endcsname{0.00141}
\expandafter\def\csname plaid@Rotor37@250@bestname\endcsname{pca\_kumo[large,e1,v0.999]}
\expandafter\def\csname plaid@Rotor37@250@floor\endcsname{0.000161}
\expandafter\def\csname plaid@Rotor37@250@ratio\endcsname{13.7}
\expandafter\def\csname plaid@Rotor37@250@onfront\endcsname{no}
\expandafter\def\csname plaid@Rotor37@250@onfronttotal\endcsname{yes}
\expandafter\def\csname plaid@Rotor37@250@onfrontqten\endcsname{yes}
\expandafter\def\csname plaid@Rotor37@250@onfrontqhundred\endcsname{no}
\expandafter\def\csname plaid@Rotor37@250@beatsnn\endcsname{yes}
\expandafter\def\csname plaid@Rotor37@250@mostaccurate\endcsname{no}
\expandafter\def\csname plaid@Rotor37@250@fit\endcsname{23.5}
\expandafter\def\csname plaid@Rotor37@250@query\endcsname{1.52}
\expandafter\def\csname plaid@Rotor37@250@bestfit\endcsname{38.7}
\expandafter\def\csname plaid@Rotor37@250@bestquery\endcsname{5.97}
\expandafter\def\csname plaid@Rotor37@250@coverage\endcsname{0.918}
\expandafter\def\csname plaid@Rotor37@250@scalars\endcsname{3}
\expandafter\def\csname plaid@Rotor37@250@covfmin\endcsname{0.91}
\expandafter\def\csname plaid@Rotor37@250@covfmax\endcsname{0.924}
\expandafter\def\csname plaid@Rotor37@250@r2\endcsname{0.941}
\expandafter\def\csname plaid@Rotor37@250@bestr2\endcsname{0.979}
\expandafter\def\csname plaid@Rotor37@250@r2ahead\endcsname{0}
\expandafter\def\csname plaid@Rotor37@250@knodes\endcsname{30}
\expandafter\def\csname plaid@Rotor37@250@peak\endcsname{11.9}
\expandafter\def\csname plaid@all@250@onfront\endcsname{2}
\expandafter\def\csname plaid@all@250@onfronttotal\endcsname{4}
\expandafter\def\csname plaid@all@250@onfrontqten\endcsname{4}
\expandafter\def\csname plaid@all@250@onfrontqhundred\endcsname{2}
\expandafter\def\csname plaid@all@250@mostaccurate\endcsname{0}
\expandafter\def\csname plaid@all@250@beatsnn\endcsname{4}
\expandafter\def\csname plaid@all@250@datasets\endcsname{4}
\expandafter\def\csname plaid@all@250@scalars\endcsname{13}
\expandafter\def\csname plaid@all@250@r2ahead\endcsname{2}
\expandafter\def\csname plaid@all@250@covmin\endcsname{0.901}
\expandafter\def\csname plaid@all@250@covmax\endcsname{0.949}
\expandafter\def\csname plaid@all@250@covfieldmin\endcsname{0.83}
\expandafter\def\csname plaid@all@250@covfieldmax\endcsname{0.98}
\expandafter\def\csname plaid@all@64@decreasing\endcsname{5}
}{\providecommand{\plaid}[3]{\todo{#1 #2 #3}}}
\IfFileExists{results/competitions.tex}{\providecommand{\competition}[2]{\ifcsname competition@#1@#2\endcsname\csname competition@#1@#2\endcsname\else\todo{#1 #2}\fi}
\expandafter\def\csname competition@2D_MSHE@min\endcsname{1.6}
\expandafter\def\csname competition@2D_MSHE@max\endcsname{5.5}
\expandafter\def\csname competition@2D_profile@min\endcsname{4.9}
\expandafter\def\csname competition@2D_profile@max\endcsname{9.9}
\expandafter\def\csname competition@Rotor37@min\endcsname{2.8}
\expandafter\def\csname competition@Rotor37@max\endcsname{5.6}
\expandafter\def\csname competition@Tensile2d@min\endcsname{4.4}
\expandafter\def\csname competition@Tensile2d@max\endcsname{7.8}
\expandafter\def\csname competition@VKI-LS59@min\endcsname{3.1}
\expandafter\def\csname competition@VKI-LS59@max\endcsname{3.3}
\expandafter\def\csname competition@all@min\endcsname{2.8}
\expandafter\def\csname competition@all@max\endcsname{9.9}
}{\providecommand{\competition}[2]{\todo{#1 #2}}}
\IfFileExists{results/heldout.tex}{\providecommand{\heldout}[3]{\ifcsname heldout@#1@#2@#3\endcsname\csname heldout@#1@#2@#3\endcsname\else\todo{#1 #2 #3}\fi}
\expandafter\def\csname heldout@all@8@tasks\endcsname{28}
\expandafter\def\csname heldout@all@8@onfront\endcsname{23}
\expandafter\def\csname heldout@all@8@onfronttotal\endcsname{23}
\expandafter\def\csname heldout@all@8@mostaccurate\endcsname{21}
\expandafter\def\csname heldout@all@8@ahead\endcsname{21}
\expandafter\def\csname heldout@all@8@ratio\endcsname{0.5}
\expandafter\def\csname heldout@all@8@ratiomin\endcsname{0.082}
\expandafter\def\csname heldout@all@8@ratiomax\endcsname{4.1}
\expandafter\def\csname heldout@radial_plate@8@tasks\endcsname{4}
\expandafter\def\csname heldout@radial_plate@8@onfront\endcsname{3}
\expandafter\def\csname heldout@radial_plate@8@onfronttotal\endcsname{3}
\expandafter\def\csname heldout@radial_plate@8@mostaccurate\endcsname{2}
\expandafter\def\csname heldout@radial_plate@8@ahead\endcsname{2}
\expandafter\def\csname heldout@radial_plate@8@ratio\endcsname{0.99}
\expandafter\def\csname heldout@radial_plate@8@ratiomin\endcsname{0.51}
\expandafter\def\csname heldout@radial_plate@8@ratiomax\endcsname{1.2}
\expandafter\def\csname heldout@radial_heat@8@tasks\endcsname{4}
\expandafter\def\csname heldout@radial_heat@8@onfront\endcsname{4}
\expandafter\def\csname heldout@radial_heat@8@onfronttotal\endcsname{4}
\expandafter\def\csname heldout@radial_heat@8@mostaccurate\endcsname{4}
\expandafter\def\csname heldout@radial_heat@8@ahead\endcsname{4}
\expandafter\def\csname heldout@radial_heat@8@ratio\endcsname{0.25}
\expandafter\def\csname heldout@radial_heat@8@ratiomin\endcsname{0.2}
\expandafter\def\csname heldout@radial_heat@8@ratiomax\endcsname{0.48}
\expandafter\def\csname heldout@radial_flow_darcy@8@tasks\endcsname{4}
\expandafter\def\csname heldout@radial_flow_darcy@8@onfront\endcsname{3}
\expandafter\def\csname heldout@radial_flow_darcy@8@onfronttotal\endcsname{3}
\expandafter\def\csname heldout@radial_flow_darcy@8@mostaccurate\endcsname{3}
\expandafter\def\csname heldout@radial_flow_darcy@8@ahead\endcsname{3}
\expandafter\def\csname heldout@radial_flow_darcy@8@ratio\endcsname{0.44}
\expandafter\def\csname heldout@radial_flow_darcy@8@ratiomin\endcsname{0.35}
\expandafter\def\csname heldout@radial_flow_darcy@8@ratiomax\endcsname{1.3}
\expandafter\def\csname heldout@shapes_plate@8@tasks\endcsname{4}
\expandafter\def\csname heldout@shapes_plate@8@onfront\endcsname{4}
\expandafter\def\csname heldout@shapes_plate@8@onfronttotal\endcsname{4}
\expandafter\def\csname heldout@shapes_plate@8@mostaccurate\endcsname{4}
\expandafter\def\csname heldout@shapes_plate@8@ahead\endcsname{4}
\expandafter\def\csname heldout@shapes_plate@8@ratio\endcsname{0.46}
\expandafter\def\csname heldout@shapes_plate@8@ratiomin\endcsname{0.3}
\expandafter\def\csname heldout@shapes_plate@8@ratiomax\endcsname{0.91}
\expandafter\def\csname heldout@shapes_heat@8@tasks\endcsname{4}
\expandafter\def\csname heldout@shapes_heat@8@onfront\endcsname{1}
\expandafter\def\csname heldout@shapes_heat@8@onfronttotal\endcsname{1}
\expandafter\def\csname heldout@shapes_heat@8@mostaccurate\endcsname{0}
\expandafter\def\csname heldout@shapes_heat@8@ahead\endcsname{0}
\expandafter\def\csname heldout@shapes_heat@8@ratio\endcsname{2.2}
\expandafter\def\csname heldout@shapes_heat@8@ratiomin\endcsname{1.2}
\expandafter\def\csname heldout@shapes_heat@8@ratiomax\endcsname{4.1}
\expandafter\def\csname heldout@shapes_flow_darcy@8@tasks\endcsname{4}
\expandafter\def\csname heldout@shapes_flow_darcy@8@onfront\endcsname{4}
\expandafter\def\csname heldout@shapes_flow_darcy@8@onfronttotal\endcsname{4}
\expandafter\def\csname heldout@shapes_flow_darcy@8@mostaccurate\endcsname{4}
\expandafter\def\csname heldout@shapes_flow_darcy@8@ahead\endcsname{4}
\expandafter\def\csname heldout@shapes_flow_darcy@8@ratio\endcsname{0.74}
\expandafter\def\csname heldout@shapes_flow_darcy@8@ratiomin\endcsname{0.35}
\expandafter\def\csname heldout@shapes_flow_darcy@8@ratiomax\endcsname{0.93}
\expandafter\def\csname heldout@shapes_surface_strip_strip@8@tasks\endcsname{4}
\expandafter\def\csname heldout@shapes_surface_strip_strip@8@onfront\endcsname{4}
\expandafter\def\csname heldout@shapes_surface_strip_strip@8@onfronttotal\endcsname{4}
\expandafter\def\csname heldout@shapes_surface_strip_strip@8@mostaccurate\endcsname{4}
\expandafter\def\csname heldout@shapes_surface_strip_strip@8@ahead\endcsname{4}
\expandafter\def\csname heldout@shapes_surface_strip_strip@8@ratio\endcsname{0.28}
\expandafter\def\csname heldout@shapes_surface_strip_strip@8@ratiomin\endcsname{0.082}
\expandafter\def\csname heldout@shapes_surface_strip_strip@8@ratiomax\endcsname{0.46}
\expandafter\def\csname heldout@all@16@tasks\endcsname{28}
\expandafter\def\csname heldout@all@16@onfront\endcsname{24}
\expandafter\def\csname heldout@all@16@onfronttotal\endcsname{21}
\expandafter\def\csname heldout@all@16@mostaccurate\endcsname{19}
\expandafter\def\csname heldout@all@16@ahead\endcsname{19}
\expandafter\def\csname heldout@all@16@ratio\endcsname{0.54}
\expandafter\def\csname heldout@all@16@ratiomin\endcsname{0.075}
\expandafter\def\csname heldout@all@16@ratiomax\endcsname{4}
\expandafter\def\csname heldout@radial_plate@16@tasks\endcsname{4}
\expandafter\def\csname heldout@radial_plate@16@onfront\endcsname{3}
\expandafter\def\csname heldout@radial_plate@16@onfronttotal\endcsname{3}
\expandafter\def\csname heldout@radial_plate@16@mostaccurate\endcsname{2}
\expandafter\def\csname heldout@radial_plate@16@ahead\endcsname{2}
\expandafter\def\csname heldout@radial_plate@16@ratio\endcsname{0.92}
\expandafter\def\csname heldout@radial_plate@16@ratiomin\endcsname{0.33}
\expandafter\def\csname heldout@radial_plate@16@ratiomax\endcsname{1.8}
\expandafter\def\csname heldout@radial_heat@16@tasks\endcsname{4}
\expandafter\def\csname heldout@radial_heat@16@onfront\endcsname{4}
\expandafter\def\csname heldout@radial_heat@16@onfronttotal\endcsname{4}
\expandafter\def\csname heldout@radial_heat@16@mostaccurate\endcsname{4}
\expandafter\def\csname heldout@radial_heat@16@ahead\endcsname{4}
\expandafter\def\csname heldout@radial_heat@16@ratio\endcsname{0.48}
\expandafter\def\csname heldout@radial_heat@16@ratiomin\endcsname{0.26}
\expandafter\def\csname heldout@radial_heat@16@ratiomax\endcsname{0.54}
\expandafter\def\csname heldout@radial_flow_darcy@16@tasks\endcsname{4}
\expandafter\def\csname heldout@radial_flow_darcy@16@onfront\endcsname{2}
\expandafter\def\csname heldout@radial_flow_darcy@16@onfronttotal\endcsname{2}
\expandafter\def\csname heldout@radial_flow_darcy@16@mostaccurate\endcsname{2}
\expandafter\def\csname heldout@radial_flow_darcy@16@ahead\endcsname{2}
\expandafter\def\csname heldout@radial_flow_darcy@16@ratio\endcsname{0.92}
\expandafter\def\csname heldout@radial_flow_darcy@16@ratiomin\endcsname{0.27}
\expandafter\def\csname heldout@radial_flow_darcy@16@ratiomax\endcsname{1.4}
\expandafter\def\csname heldout@shapes_plate@16@tasks\endcsname{4}
\expandafter\def\csname heldout@shapes_plate@16@onfront\endcsname{4}
\expandafter\def\csname heldout@shapes_plate@16@onfronttotal\endcsname{4}
\expandafter\def\csname heldout@shapes_plate@16@mostaccurate\endcsname{3}
\expandafter\def\csname heldout@shapes_plate@16@ahead\endcsname{3}
\expandafter\def\csname heldout@shapes_plate@16@ratio\endcsname{0.52}
\expandafter\def\csname heldout@shapes_plate@16@ratiomin\endcsname{0.26}
\expandafter\def\csname heldout@shapes_plate@16@ratiomax\endcsname{1.2}
\expandafter\def\csname heldout@shapes_heat@16@tasks\endcsname{4}
\expandafter\def\csname heldout@shapes_heat@16@onfront\endcsname{3}
\expandafter\def\csname heldout@shapes_heat@16@onfronttotal\endcsname{1}
\expandafter\def\csname heldout@shapes_heat@16@mostaccurate\endcsname{1}
\expandafter\def\csname heldout@shapes_heat@16@ahead\endcsname{1}
\expandafter\def\csname heldout@shapes_heat@16@ratio\endcsname{3.1}
\expandafter\def\csname heldout@shapes_heat@16@ratiomin\endcsname{0.26}
\expandafter\def\csname heldout@shapes_heat@16@ratiomax\endcsname{4}
\expandafter\def\csname heldout@shapes_flow_darcy@16@tasks\endcsname{4}
\expandafter\def\csname heldout@shapes_flow_darcy@16@onfront\endcsname{4}
\expandafter\def\csname heldout@shapes_flow_darcy@16@onfronttotal\endcsname{3}
\expandafter\def\csname heldout@shapes_flow_darcy@16@mostaccurate\endcsname{3}
\expandafter\def\csname heldout@shapes_flow_darcy@16@ahead\endcsname{3}
\expandafter\def\csname heldout@shapes_flow_darcy@16@ratio\endcsname{0.77}
\expandafter\def\csname heldout@shapes_flow_darcy@16@ratiomin\endcsname{0.34}
\expandafter\def\csname heldout@shapes_flow_darcy@16@ratiomax\endcsname{1.7}
\expandafter\def\csname heldout@shapes_surface_strip_strip@16@tasks\endcsname{4}
\expandafter\def\csname heldout@shapes_surface_strip_strip@16@onfront\endcsname{4}
\expandafter\def\csname heldout@shapes_surface_strip_strip@16@onfronttotal\endcsname{4}
\expandafter\def\csname heldout@shapes_surface_strip_strip@16@mostaccurate\endcsname{4}
\expandafter\def\csname heldout@shapes_surface_strip_strip@16@ahead\endcsname{4}
\expandafter\def\csname heldout@shapes_surface_strip_strip@16@ratio\endcsname{0.23}
\expandafter\def\csname heldout@shapes_surface_strip_strip@16@ratiomin\endcsname{0.075}
\expandafter\def\csname heldout@shapes_surface_strip_strip@16@ratiomax\endcsname{0.37}
\expandafter\def\csname heldout@all@32@tasks\endcsname{28}
\expandafter\def\csname heldout@all@32@onfront\endcsname{24}
\expandafter\def\csname heldout@all@32@onfronttotal\endcsname{21}
\expandafter\def\csname heldout@all@32@mostaccurate\endcsname{18}
\expandafter\def\csname heldout@all@32@ahead\endcsname{18}
\expandafter\def\csname heldout@all@32@ratio\endcsname{0.74}
\expandafter\def\csname heldout@all@32@ratiomin\endcsname{0.074}
\expandafter\def\csname heldout@all@32@ratiomax\endcsname{3}
\expandafter\def\csname heldout@radial_plate@32@tasks\endcsname{4}
\expandafter\def\csname heldout@radial_plate@32@onfront\endcsname{3}
\expandafter\def\csname heldout@radial_plate@32@onfronttotal\endcsname{3}
\expandafter\def\csname heldout@radial_plate@32@mostaccurate\endcsname{3}
\expandafter\def\csname heldout@radial_plate@32@ahead\endcsname{3}
\expandafter\def\csname heldout@radial_plate@32@ratio\endcsname{0.89}
\expandafter\def\csname heldout@radial_plate@32@ratiomin\endcsname{0.35}
\expandafter\def\csname heldout@radial_plate@32@ratiomax\endcsname{1.2}
\expandafter\def\csname heldout@radial_heat@32@tasks\endcsname{4}
\expandafter\def\csname heldout@radial_heat@32@onfront\endcsname{4}
\expandafter\def\csname heldout@radial_heat@32@onfronttotal\endcsname{4}
\expandafter\def\csname heldout@radial_heat@32@mostaccurate\endcsname{4}
\expandafter\def\csname heldout@radial_heat@32@ahead\endcsname{4}
\expandafter\def\csname heldout@radial_heat@32@ratio\endcsname{0.54}
\expandafter\def\csname heldout@radial_heat@32@ratiomin\endcsname{0.36}
\expandafter\def\csname heldout@radial_heat@32@ratiomax\endcsname{0.93}
\expandafter\def\csname heldout@radial_flow_darcy@32@tasks\endcsname{4}
\expandafter\def\csname heldout@radial_flow_darcy@32@onfront\endcsname{2}
\expandafter\def\csname heldout@radial_flow_darcy@32@onfronttotal\endcsname{2}
\expandafter\def\csname heldout@radial_flow_darcy@32@mostaccurate\endcsname{2}
\expandafter\def\csname heldout@radial_flow_darcy@32@ahead\endcsname{2}
\expandafter\def\csname heldout@radial_flow_darcy@32@ratio\endcsname{1.1}
\expandafter\def\csname heldout@radial_flow_darcy@32@ratiomin\endcsname{0.48}
\expandafter\def\csname heldout@radial_flow_darcy@32@ratiomax\endcsname{3}
\expandafter\def\csname heldout@shapes_plate@32@tasks\endcsname{4}
\expandafter\def\csname heldout@shapes_plate@32@onfront\endcsname{4}
\expandafter\def\csname heldout@shapes_plate@32@onfronttotal\endcsname{4}
\expandafter\def\csname heldout@shapes_plate@32@mostaccurate\endcsname{3}
\expandafter\def\csname heldout@shapes_plate@32@ahead\endcsname{3}
\expandafter\def\csname heldout@shapes_plate@32@ratio\endcsname{0.72}
\expandafter\def\csname heldout@shapes_plate@32@ratiomin\endcsname{0.44}
\expandafter\def\csname heldout@shapes_plate@32@ratiomax\endcsname{1.6}
\expandafter\def\csname heldout@shapes_heat@32@tasks\endcsname{4}
\expandafter\def\csname heldout@shapes_heat@32@onfront\endcsname{3}
\expandafter\def\csname heldout@shapes_heat@32@onfronttotal\endcsname{2}
\expandafter\def\csname heldout@shapes_heat@32@mostaccurate\endcsname{0}
\expandafter\def\csname heldout@shapes_heat@32@ahead\endcsname{0}
\expandafter\def\csname heldout@shapes_heat@32@ratio\endcsname{2.5}
\expandafter\def\csname heldout@shapes_heat@32@ratiomin\endcsname{1.2}
\expandafter\def\csname heldout@shapes_heat@32@ratiomax\endcsname{2.9}
\expandafter\def\csname heldout@shapes_flow_darcy@32@tasks\endcsname{4}
\expandafter\def\csname heldout@shapes_flow_darcy@32@onfront\endcsname{4}
\expandafter\def\csname heldout@shapes_flow_darcy@32@onfronttotal\endcsname{2}
\expandafter\def\csname heldout@shapes_flow_darcy@32@mostaccurate\endcsname{2}
\expandafter\def\csname heldout@shapes_flow_darcy@32@ahead\endcsname{2}
\expandafter\def\csname heldout@shapes_flow_darcy@32@ratio\endcsname{0.96}
\expandafter\def\csname heldout@shapes_flow_darcy@32@ratiomin\endcsname{0.5}
\expandafter\def\csname heldout@shapes_flow_darcy@32@ratiomax\endcsname{1.7}
\expandafter\def\csname heldout@shapes_surface_strip_strip@32@tasks\endcsname{4}
\expandafter\def\csname heldout@shapes_surface_strip_strip@32@onfront\endcsname{4}
\expandafter\def\csname heldout@shapes_surface_strip_strip@32@onfronttotal\endcsname{4}
\expandafter\def\csname heldout@shapes_surface_strip_strip@32@mostaccurate\endcsname{4}
\expandafter\def\csname heldout@shapes_surface_strip_strip@32@ahead\endcsname{4}
\expandafter\def\csname heldout@shapes_surface_strip_strip@32@ratio\endcsname{0.22}
\expandafter\def\csname heldout@shapes_surface_strip_strip@32@ratiomin\endcsname{0.074}
\expandafter\def\csname heldout@shapes_surface_strip_strip@32@ratiomax\endcsname{0.33}
\expandafter\def\csname heldout@all@64@tasks\endcsname{28}
\expandafter\def\csname heldout@all@64@onfront\endcsname{21}
\expandafter\def\csname heldout@all@64@onfronttotal\endcsname{16}
\expandafter\def\csname heldout@all@64@mostaccurate\endcsname{14}
\expandafter\def\csname heldout@all@64@ahead\endcsname{15}
\expandafter\def\csname heldout@all@64@ratio\endcsname{0.98}
\expandafter\def\csname heldout@all@64@ratiomin\endcsname{0.066}
\expandafter\def\csname heldout@all@64@ratiomax\endcsname{8.7}
\expandafter\def\csname heldout@radial_plate@64@tasks\endcsname{4}
\expandafter\def\csname heldout@radial_plate@64@onfront\endcsname{2}
\expandafter\def\csname heldout@radial_plate@64@onfronttotal\endcsname{2}
\expandafter\def\csname heldout@radial_plate@64@mostaccurate\endcsname{2}
\expandafter\def\csname heldout@radial_plate@64@ahead\endcsname{3}
\expandafter\def\csname heldout@radial_plate@64@ratio\endcsname{0.93}
\expandafter\def\csname heldout@radial_plate@64@ratiomin\endcsname{0.42}
\expandafter\def\csname heldout@radial_plate@64@ratiomax\endcsname{3.6}
\expandafter\def\csname heldout@radial_heat@64@tasks\endcsname{4}
\expandafter\def\csname heldout@radial_heat@64@onfront\endcsname{3}
\expandafter\def\csname heldout@radial_heat@64@onfronttotal\endcsname{3}
\expandafter\def\csname heldout@radial_heat@64@mostaccurate\endcsname{3}
\expandafter\def\csname heldout@radial_heat@64@ahead\endcsname{3}
\expandafter\def\csname heldout@radial_heat@64@ratio\endcsname{0.92}
\expandafter\def\csname heldout@radial_heat@64@ratiomin\endcsname{0.39}
\expandafter\def\csname heldout@radial_heat@64@ratiomax\endcsname{1.5}
\expandafter\def\csname heldout@radial_flow_darcy@64@tasks\endcsname{4}
\expandafter\def\csname heldout@radial_flow_darcy@64@onfront\endcsname{3}
\expandafter\def\csname heldout@radial_flow_darcy@64@onfronttotal\endcsname{2}
\expandafter\def\csname heldout@radial_flow_darcy@64@mostaccurate\endcsname{2}
\expandafter\def\csname heldout@radial_flow_darcy@64@ahead\endcsname{2}
\expandafter\def\csname heldout@radial_flow_darcy@64@ratio\endcsname{1.2}
\expandafter\def\csname heldout@radial_flow_darcy@64@ratiomin\endcsname{0.64}
\expandafter\def\csname heldout@radial_flow_darcy@64@ratiomax\endcsname{8.7}
\expandafter\def\csname heldout@shapes_plate@64@tasks\endcsname{4}
\expandafter\def\csname heldout@shapes_plate@64@onfront\endcsname{3}
\expandafter\def\csname heldout@shapes_plate@64@onfronttotal\endcsname{3}
\expandafter\def\csname heldout@shapes_plate@64@mostaccurate\endcsname{2}
\expandafter\def\csname heldout@shapes_plate@64@ahead\endcsname{2}
\expandafter\def\csname heldout@shapes_plate@64@ratio\endcsname{1.2}
\expandafter\def\csname heldout@shapes_plate@64@ratiomin\endcsname{0.65}
\expandafter\def\csname heldout@shapes_plate@64@ratiomax\endcsname{2}
\expandafter\def\csname heldout@shapes_heat@64@tasks\endcsname{4}
\expandafter\def\csname heldout@shapes_heat@64@onfront\endcsname{2}
\expandafter\def\csname heldout@shapes_heat@64@onfronttotal\endcsname{1}
\expandafter\def\csname heldout@shapes_heat@64@mostaccurate\endcsname{0}
\expandafter\def\csname heldout@shapes_heat@64@ahead\endcsname{0}
\expandafter\def\csname heldout@shapes_heat@64@ratio\endcsname{3.1}
\expandafter\def\csname heldout@shapes_heat@64@ratiomin\endcsname{2.8}
\expandafter\def\csname heldout@shapes_heat@64@ratiomax\endcsname{7.1}
\expandafter\def\csname heldout@shapes_flow_darcy@64@tasks\endcsname{4}
\expandafter\def\csname heldout@shapes_flow_darcy@64@onfront\endcsname{4}
\expandafter\def\csname heldout@shapes_flow_darcy@64@onfronttotal\endcsname{1}
\expandafter\def\csname heldout@shapes_flow_darcy@64@mostaccurate\endcsname{1}
\expandafter\def\csname heldout@shapes_flow_darcy@64@ahead\endcsname{1}
\expandafter\def\csname heldout@shapes_flow_darcy@64@ratio\endcsname{1.3}
\expandafter\def\csname heldout@shapes_flow_darcy@64@ratiomin\endcsname{1}
\expandafter\def\csname heldout@shapes_flow_darcy@64@ratiomax\endcsname{1.5}
\expandafter\def\csname heldout@shapes_surface_strip_strip@64@tasks\endcsname{4}
\expandafter\def\csname heldout@shapes_surface_strip_strip@64@onfront\endcsname{4}
\expandafter\def\csname heldout@shapes_surface_strip_strip@64@onfronttotal\endcsname{4}
\expandafter\def\csname heldout@shapes_surface_strip_strip@64@mostaccurate\endcsname{4}
\expandafter\def\csname heldout@shapes_surface_strip_strip@64@ahead\endcsname{4}
\expandafter\def\csname heldout@shapes_surface_strip_strip@64@ratio\endcsname{0.21}
\expandafter\def\csname heldout@shapes_surface_strip_strip@64@ratiomin\endcsname{0.066}
\expandafter\def\csname heldout@shapes_surface_strip_strip@64@ratiomax\endcsname{0.32}
}{\providecommand{\heldout}[3]{\todo{#1 #2 #3}}}
\IfFileExists{results/finetune.tex}{\providecommand{\ft}[3]{\ifcsname ft@#1@#2@#3\endcsname\csname ft@#1@#2@#3\endcsname\else\todo{#1 #2 #3}\fi}
\expandafter\def\csname ft@pretrained@16@fields\endcsname{0.515}
\expandafter\def\csname ft@pretrained@16@fieldsnear\endcsname{0.857}
\expandafter\def\csname ft@pretrained@16@fieldscov\endcsname{0.896}
\expandafter\def\csname ft@proxy@16@fields\endcsname{0.236}
\expandafter\def\csname ft@proxy@16@fieldsnear\endcsname{0.317}
\expandafter\def\csname ft@proxy@16@fieldscov\endcsname{0.151}
\expandafter\def\csname ft@augmentedproxy@16@fields\endcsname{0.203}
\expandafter\def\csname ft@augmentedproxy@16@fieldsnear\endcsname{0.254}
\expandafter\def\csname ft@augmentedproxy@16@fieldscov\endcsname{0.711}
\expandafter\def\csname ft@pretrained@16@fieldsgain\endcsname{1}
\expandafter\def\csname ft@pretrained@16@fieldsneargain\endcsname{1}
\expandafter\def\csname ft@proxy@16@fieldsgain\endcsname{2.2}
\expandafter\def\csname ft@proxy@16@fieldsneargain\endcsname{2.7}
\expandafter\def\csname ft@augmentedproxy@16@fieldsgain\endcsname{2.5}
\expandafter\def\csname ft@augmentedproxy@16@fieldsneargain\endcsname{3.4}
\expandafter\def\csname ft@best@16@fields\endcsname{0.621}
\expandafter\def\csname ft@floor@16@fields\endcsname{0.505}
\expandafter\def\csname ft@pretrained@16@stress\endcsname{0.805}
\expandafter\def\csname ft@pretrained@16@stressnear\endcsname{1.14}
\expandafter\def\csname ft@pretrained@16@stresscov\endcsname{0.887}
\expandafter\def\csname ft@proxy@16@stress\endcsname{0.369}
\expandafter\def\csname ft@proxy@16@stressnear\endcsname{0.408}
\expandafter\def\csname ft@proxy@16@stresscov\endcsname{0.161}
\expandafter\def\csname ft@augmentedproxy@16@stress\endcsname{0.315}
\expandafter\def\csname ft@augmentedproxy@16@stressnear\endcsname{0.304}
\expandafter\def\csname ft@augmentedproxy@16@stresscov\endcsname{0.902}
\expandafter\def\csname ft@pretrained@16@stressgain\endcsname{1}
\expandafter\def\csname ft@pretrained@16@stressneargain\endcsname{1}
\expandafter\def\csname ft@proxy@16@stressgain\endcsname{2.2}
\expandafter\def\csname ft@proxy@16@stressneargain\endcsname{2.8}
\expandafter\def\csname ft@augmentedproxy@16@stressgain\endcsname{2.6}
\expandafter\def\csname ft@augmentedproxy@16@stressneargain\endcsname{3.8}
\expandafter\def\csname ft@best@16@stress\endcsname{0.818}
\expandafter\def\csname ft@floor@16@stress\endcsname{0.661}
\expandafter\def\csname ft@pretrained@16@psi\endcsname{0.257}
\expandafter\def\csname ft@pretrained@16@psinear\endcsname{0.907}
\expandafter\def\csname ft@pretrained@16@psicov\endcsname{0.831}
\expandafter\def\csname ft@proxy@16@psi\endcsname{0.125}
\expandafter\def\csname ft@proxy@16@psinear\endcsname{0.449}
\expandafter\def\csname ft@proxy@16@psicov\endcsname{0.151}
\expandafter\def\csname ft@augmentedproxy@16@psi\endcsname{0.118}
\expandafter\def\csname ft@augmentedproxy@16@psinear\endcsname{0.464}
\expandafter\def\csname ft@augmentedproxy@16@psicov\endcsname{0.581}
\expandafter\def\csname ft@pretrained@16@psigain\endcsname{1}
\expandafter\def\csname ft@pretrained@16@psineargain\endcsname{1}
\expandafter\def\csname ft@proxy@16@psigain\endcsname{2}
\expandafter\def\csname ft@proxy@16@psineargain\endcsname{2}
\expandafter\def\csname ft@augmentedproxy@16@psigain\endcsname{2.2}
\expandafter\def\csname ft@augmentedproxy@16@psineargain\endcsname{2}
\expandafter\def\csname ft@best@16@psi\endcsname{0.954}
\expandafter\def\csname ft@floor@16@psi\endcsname{0.837}
\expandafter\def\csname ft@pretrained@16@displacement\endcsname{0.0633}
\expandafter\def\csname ft@pretrained@16@displacementnear\endcsname{0.265}
\expandafter\def\csname ft@pretrained@16@displacementcov\endcsname{0.947}
\expandafter\def\csname ft@proxy@16@displacement\endcsname{0.0257}
\expandafter\def\csname ft@proxy@16@displacementnear\endcsname{0.0669}
\expandafter\def\csname ft@proxy@16@displacementcov\endcsname{0.131}
\expandafter\def\csname ft@augmentedproxy@16@displacement\endcsname{0.0199}
\expandafter\def\csname ft@augmentedproxy@16@displacementnear\endcsname{0.0474}
\expandafter\def\csname ft@augmentedproxy@16@displacementcov\endcsname{0.396}
\expandafter\def\csname ft@pretrained@16@displacementgain\endcsname{1}
\expandafter\def\csname ft@pretrained@16@displacementneargain\endcsname{1}
\expandafter\def\csname ft@proxy@16@displacementgain\endcsname{2.5}
\expandafter\def\csname ft@proxy@16@displacementneargain\endcsname{4}
\expandafter\def\csname ft@augmentedproxy@16@displacementgain\endcsname{3.2}
\expandafter\def\csname ft@augmentedproxy@16@displacementneargain\endcsname{5.6}
\expandafter\def\csname ft@best@16@displacement\endcsname{0.0598}
\expandafter\def\csname ft@floor@16@displacement\endcsname{0.0293}
\expandafter\def\csname ft@pretrained@64@fields\endcsname{0.456}
\expandafter\def\csname ft@pretrained@64@fieldsnear\endcsname{0.794}
\expandafter\def\csname ft@pretrained@64@fieldscov\endcsname{0.877}
\expandafter\def\csname ft@proxy@64@fields\endcsname{0.228}
\expandafter\def\csname ft@proxy@64@fieldsnear\endcsname{0.307}
\expandafter\def\csname ft@proxy@64@fieldscov\endcsname{0.145}
\expandafter\def\csname ft@augmentedproxy@64@fields\endcsname{0.174}
\expandafter\def\csname ft@augmentedproxy@64@fieldsnear\endcsname{0.24}
\expandafter\def\csname ft@augmentedproxy@64@fieldscov\endcsname{0.725}
\expandafter\def\csname ft@pretrained@64@fieldsgain\endcsname{1}
\expandafter\def\csname ft@pretrained@64@fieldsneargain\endcsname{1}
\expandafter\def\csname ft@proxy@64@fieldsgain\endcsname{2}
\expandafter\def\csname ft@proxy@64@fieldsneargain\endcsname{2.6}
\expandafter\def\csname ft@augmentedproxy@64@fieldsgain\endcsname{2.6}
\expandafter\def\csname ft@augmentedproxy@64@fieldsneargain\endcsname{3.3}
\expandafter\def\csname ft@best@64@fields\endcsname{0.479}
\expandafter\def\csname ft@floor@64@fields\endcsname{0.319}
\expandafter\def\csname ft@pretrained@64@stress\endcsname{0.717}
\expandafter\def\csname ft@pretrained@64@stressnear\endcsname{1.08}
\expandafter\def\csname ft@pretrained@64@stresscov\endcsname{0.877}
\expandafter\def\csname ft@proxy@64@stress\endcsname{0.356}
\expandafter\def\csname ft@proxy@64@stressnear\endcsname{0.399}
\expandafter\def\csname ft@proxy@64@stresscov\endcsname{0.156}
\expandafter\def\csname ft@augmentedproxy@64@stress\endcsname{0.27}
\expandafter\def\csname ft@augmentedproxy@64@stressnear\endcsname{0.297}
\expandafter\def\csname ft@augmentedproxy@64@stresscov\endcsname{0.925}
\expandafter\def\csname ft@pretrained@64@stressgain\endcsname{1}
\expandafter\def\csname ft@pretrained@64@stressneargain\endcsname{1}
\expandafter\def\csname ft@proxy@64@stressgain\endcsname{2}
\expandafter\def\csname ft@proxy@64@stressneargain\endcsname{2.7}
\expandafter\def\csname ft@augmentedproxy@64@stressgain\endcsname{2.7}
\expandafter\def\csname ft@augmentedproxy@64@stressneargain\endcsname{3.7}
\expandafter\def\csname ft@best@64@stress\endcsname{0.632}
\expandafter\def\csname ft@floor@64@stress\endcsname{0.443}
\expandafter\def\csname ft@pretrained@64@psi\endcsname{0.235}
\expandafter\def\csname ft@pretrained@64@psinear\endcsname{0.776}
\expandafter\def\csname ft@pretrained@64@psicov\endcsname{0.79}
\expandafter\def\csname ft@proxy@64@psi\endcsname{0.126}
\expandafter\def\csname ft@proxy@64@psinear\endcsname{0.434}
\expandafter\def\csname ft@proxy@64@psicov\endcsname{0.132}
\expandafter\def\csname ft@augmentedproxy@64@psi\endcsname{0.106}
\expandafter\def\csname ft@augmentedproxy@64@psinear\endcsname{0.419}
\expandafter\def\csname ft@augmentedproxy@64@psicov\endcsname{0.518}
\expandafter\def\csname ft@pretrained@64@psigain\endcsname{1}
\expandafter\def\csname ft@pretrained@64@psineargain\endcsname{1}
\expandafter\def\csname ft@proxy@64@psigain\endcsname{1.9}
\expandafter\def\csname ft@proxy@64@psineargain\endcsname{1.8}
\expandafter\def\csname ft@augmentedproxy@64@psigain\endcsname{2.2}
\expandafter\def\csname ft@augmentedproxy@64@psineargain\endcsname{1.9}
\expandafter\def\csname ft@best@64@psi\endcsname{0.779}
\expandafter\def\csname ft@floor@64@psi\endcsname{0.448}
\expandafter\def\csname ft@pretrained@64@displacement\endcsname{0.0454}
\expandafter\def\csname ft@pretrained@64@displacementnear\endcsname{0.222}
\expandafter\def\csname ft@pretrained@64@displacementcov\endcsname{0.922}
\expandafter\def\csname ft@proxy@64@displacement\endcsname{0.0229}
\expandafter\def\csname ft@proxy@64@displacementnear\endcsname{0.0588}
\expandafter\def\csname ft@proxy@64@displacementcov\endcsname{0.13}
\expandafter\def\csname ft@augmentedproxy@64@displacement\endcsname{0.0154}
\expandafter\def\csname ft@augmentedproxy@64@displacementnear\endcsname{0.037}
\expandafter\def\csname ft@augmentedproxy@64@displacementcov\endcsname{0.429}
\expandafter\def\csname ft@pretrained@64@displacementgain\endcsname{1}
\expandafter\def\csname ft@pretrained@64@displacementneargain\endcsname{1}
\expandafter\def\csname ft@proxy@64@displacementgain\endcsname{2}
\expandafter\def\csname ft@proxy@64@displacementneargain\endcsname{3.8}
\expandafter\def\csname ft@augmentedproxy@64@displacementgain\endcsname{2.9}
\expandafter\def\csname ft@augmentedproxy@64@displacementneargain\endcsname{6}
\expandafter\def\csname ft@best@64@displacement\endcsname{0.0222}
\expandafter\def\csname ft@floor@64@displacement\endcsname{0.00727}
\expandafter\def\csname ft@pretrained@250@fields\endcsname{0.402}
\expandafter\def\csname ft@pretrained@250@fieldsnear\endcsname{0.704}
\expandafter\def\csname ft@pretrained@250@fieldscov\endcsname{0.898}
\expandafter\def\csname ft@proxy@250@fields\endcsname{0.225}
\expandafter\def\csname ft@proxy@250@fieldsnear\endcsname{0.302}
\expandafter\def\csname ft@proxy@250@fieldscov\endcsname{0.143}
\expandafter\def\csname ft@augmentedproxy@250@fields\endcsname{0.147}
\expandafter\def\csname ft@augmentedproxy@250@fieldsnear\endcsname{0.23}
\expandafter\def\csname ft@augmentedproxy@250@fieldscov\endcsname{0.718}
\expandafter\def\csname ft@pretrained@250@fieldsgain\endcsname{1}
\expandafter\def\csname ft@pretrained@250@fieldsneargain\endcsname{1}
\expandafter\def\csname ft@proxy@250@fieldsgain\endcsname{1.8}
\expandafter\def\csname ft@proxy@250@fieldsneargain\endcsname{2.3}
\expandafter\def\csname ft@augmentedproxy@250@fieldsgain\endcsname{2.7}
\expandafter\def\csname ft@augmentedproxy@250@fieldsneargain\endcsname{3.1}
\expandafter\def\csname ft@best@250@fields\endcsname{0.38}
\expandafter\def\csname ft@floor@250@fields\endcsname{0.151}
\expandafter\def\csname ft@pretrained@250@stress\endcsname{0.635}
\expandafter\def\csname ft@pretrained@250@stressnear\endcsname{0.978}
\expandafter\def\csname ft@pretrained@250@stresscov\endcsname{0.9}
\expandafter\def\csname ft@proxy@250@stress\endcsname{0.35}
\expandafter\def\csname ft@proxy@250@stressnear\endcsname{0.394}
\expandafter\def\csname ft@proxy@250@stresscov\endcsname{0.155}
\expandafter\def\csname ft@augmentedproxy@250@stress\endcsname{0.225}
\expandafter\def\csname ft@augmentedproxy@250@stressnear\endcsname{0.284}
\expandafter\def\csname ft@augmentedproxy@250@stresscov\endcsname{0.921}
\expandafter\def\csname ft@pretrained@250@stressgain\endcsname{1}
\expandafter\def\csname ft@pretrained@250@stressneargain\endcsname{1}
\expandafter\def\csname ft@proxy@250@stressgain\endcsname{1.8}
\expandafter\def\csname ft@proxy@250@stressneargain\endcsname{2.5}
\expandafter\def\csname ft@augmentedproxy@250@stressgain\endcsname{2.8}
\expandafter\def\csname ft@augmentedproxy@250@stressneargain\endcsname{3.4}
\expandafter\def\csname ft@best@250@stress\endcsname{0.52}
\expandafter\def\csname ft@floor@250@stress\endcsname{0.225}
\expandafter\def\csname ft@pretrained@250@psi\endcsname{0.205}
\expandafter\def\csname ft@pretrained@250@psinear\endcsname{0.647}
\expandafter\def\csname ft@pretrained@250@psicov\endcsname{0.821}
\expandafter\def\csname ft@proxy@250@psi\endcsname{0.127}
\expandafter\def\csname ft@proxy@250@psinear\endcsname{0.424}
\expandafter\def\csname ft@proxy@250@psicov\endcsname{0.128}
\expandafter\def\csname ft@augmentedproxy@250@psi\endcsname{0.104}
\expandafter\def\csname ft@augmentedproxy@250@psinear\endcsname{0.406}
\expandafter\def\csname ft@augmentedproxy@250@psicov\endcsname{0.5}
\expandafter\def\csname ft@pretrained@250@psigain\endcsname{1}
\expandafter\def\csname ft@pretrained@250@psineargain\endcsname{1}
\expandafter\def\csname ft@proxy@250@psigain\endcsname{1.6}
\expandafter\def\csname ft@proxy@250@psineargain\endcsname{1.5}
\expandafter\def\csname ft@augmentedproxy@250@psigain\endcsname{2}
\expandafter\def\csname ft@augmentedproxy@250@psineargain\endcsname{1.6}
\expandafter\def\csname ft@best@250@psi\endcsname{0.561}
\expandafter\def\csname ft@floor@250@psi\endcsname{0.152}
\expandafter\def\csname ft@pretrained@250@displacement\endcsname{0.0354}
\expandafter\def\csname ft@pretrained@250@displacementnear\endcsname{0.185}
\expandafter\def\csname ft@pretrained@250@displacementcov\endcsname{0.934}
\expandafter\def\csname ft@proxy@250@displacement\endcsname{0.0223}
\expandafter\def\csname ft@proxy@250@displacementnear\endcsname{0.0571}
\expandafter\def\csname ft@proxy@250@displacementcov\endcsname{0.126}
\expandafter\def\csname ft@augmentedproxy@250@displacement\endcsname{0.0142}
\expandafter\def\csname ft@augmentedproxy@250@displacementnear\endcsname{0.0345}
\expandafter\def\csname ft@augmentedproxy@250@displacementcov\endcsname{0.423}
\expandafter\def\csname ft@pretrained@250@displacementgain\endcsname{1}
\expandafter\def\csname ft@pretrained@250@displacementneargain\endcsname{1}
\expandafter\def\csname ft@proxy@250@displacementgain\endcsname{1.6}
\expandafter\def\csname ft@proxy@250@displacementneargain\endcsname{3.2}
\expandafter\def\csname ft@augmentedproxy@250@displacementgain\endcsname{2.5}
\expandafter\def\csname ft@augmentedproxy@250@displacementneargain\endcsname{5.4}
\expandafter\def\csname ft@best@250@displacement\endcsname{0.00993}
\expandafter\def\csname ft@floor@250@displacement\endcsname{0.00195}
}{\providecommand{\ft}[3]{\todo{#1 #2 #3}}}
\IfFileExists{results/context_control.tex}{\providecommand{\ctrl}[3]{\ifcsname ctrl@#1@#2@#3\endcsname\csname ctrl@#1@#2@#3\endcsname\else\todo{#1 #2 #3}\fi}
\expandafter\def\csname ctrl@Rotor37@outputs@8\endcsname{2.1}
\expandafter\def\csname ctrl@Rotor37@outputs@16\endcsname{2.5}
\expandafter\def\csname ctrl@Rotor37@outputs@32\endcsname{2.8}
\expandafter\def\csname ctrl@Rotor37@outputs@64\endcsname{2.9}
\expandafter\def\csname ctrl@Rotor37@outputs@125\endcsname{2.9}
\expandafter\def\csname ctrl@Rotor37@outputs@250\endcsname{3.0}
\expandafter\def\csname ctrl@Rotor37@inputs@8\endcsname{2.9}
\expandafter\def\csname ctrl@Rotor37@inputs@16\endcsname{2.6}
\expandafter\def\csname ctrl@Rotor37@inputs@32\endcsname{2.7}
\expandafter\def\csname ctrl@Rotor37@inputs@64\endcsname{2.9}
\expandafter\def\csname ctrl@Rotor37@inputs@125\endcsname{3.0}
\expandafter\def\csname ctrl@Rotor37@inputs@250\endcsname{3.0}
\expandafter\def\csname ctrl@VKI-LS59@outputs@8\endcsname{2.0}
\expandafter\def\csname ctrl@VKI-LS59@outputs@16\endcsname{2.3}
\expandafter\def\csname ctrl@VKI-LS59@outputs@32\endcsname{2.9}
\expandafter\def\csname ctrl@VKI-LS59@outputs@64\endcsname{3.7}
\expandafter\def\csname ctrl@VKI-LS59@outputs@125\endcsname{3.8}
\expandafter\def\csname ctrl@VKI-LS59@outputs@250\endcsname{3.7}
\expandafter\def\csname ctrl@VKI-LS59@inputs@8\endcsname{1.8}
\expandafter\def\csname ctrl@VKI-LS59@inputs@16\endcsname{1.6}
\expandafter\def\csname ctrl@VKI-LS59@inputs@32\endcsname{1.6}
\expandafter\def\csname ctrl@VKI-LS59@inputs@64\endcsname{1.6}
\expandafter\def\csname ctrl@VKI-LS59@inputs@125\endcsname{1.7}
\expandafter\def\csname ctrl@VKI-LS59@inputs@250\endcsname{1.8}
}{\providecommand{\ctrl}[3]{\todo{#1 #2 #3}}}
\icmltitlerunning{PhysICL: In-Context Learning of Physical Fields from Procedural Priors}

\begin{document}

\twocolumn[
  \icmltitle{PhysICL: In-Context Learning of Physical Fields\\ on Unstructured Meshes from Procedural Priors}

  \begin{icmlauthorlist}
    \icmlauthor{Fabien Casenave}{safran}
  \end{icmlauthorlist}

  \icmlaffiliation{safran}{Safran Tech, Digital Sciences \& Technologies Department, Rue des Jeunes Bois, Ch\^ateaufort, 78114 Magny-Les-Hameaux, France}
  \icmlcorrespondingauthor{Fabien Casenave}{fabien.casenave@safrangroup.com}
  \icmlkeywords{in-context learning, foundation models, physics simulation, unstructured meshes, priors}

  \vskip 0.3in
]

\printAffiliationsAndNotice{}

\begin{abstract}
Engineering design studies produce a few tens to a few hundred simulations of one problem, each on its own mesh, and their
surrogates are retrained for every study. Tabular foundation models instead infer a new task from its examples in one
forward pass, after pretraining on synthetic tasks. We demonstrate the feasibility of in-context learning for
physical field prediction on unstructured meshes. A single model, \physicl{}, pretrained solely on procedural tasks,
predicts fields, with quantiles, directly on each query mesh from labeled simulations, without updating its weights.
The same checkpoint accepts variable numbers of context examples, scalar inputs and input fields, and any number of
output fields, while mesh size and connectivity can vary within a task. Its procedural prior draws generic laws and
boundary-element physical models on morphed 2D and 3D template meshes, evaluated only at the points the model
samples, fast enough that tasks generated on the fly by CPU workers keep the GPUs busy throughout pretraining. On five
PLAID datasets never seen in pretraining, its field error decreases as the context grows and,
from $K = 32$, is below that of the nearest context simulation on every dataset and context draw. It is on par with
per-dataset reduced-order models on two of them, behind them where the fields are low-rank. On AhmedML surfaces with a median of 1.1 million nodes,
twenty times the largest pretraining mesh, its fixed spatial budgets keep the peak GPU memory at $K = 64$ at
\plaid{AhmedML}{64}{peak}\,GiB, against
\plaid{2D_MSHE}{64}{peak}\,GiB on the \plaid{2D_MSHE}{64}{knodes}{,}000-node meshes of 2D\_MSHE. Finally, on 2D\_MSHE, far from the prior,
fine-tuning on procedural tasks that reproduce its geometry and loads, without any of its outputs, at a tenth of the
pretraining cost, divides the stress error by \ft{augmentedproxy}{64}{stressgain}.
\end{abstract}

\section{Introduction}
\label{sec:intro}

A design study in engineering explores a family of shapes and operating conditions of one physical problem. Each
simulation costs minutes to days, so a study holds a few tens to a few hundred of them, each on its own mesh. Learned
surrogates of such studies, from reduced-order models on a common or morphed mesh such as
MMGP~\citep{casenave2023mmgp} to graph and transformer neural operators~\citep{pfaff2021meshgraphnets, li2023gino,
wu2024transolver, alkin2024upt, wen2025gaot}, are trained per dataset: each new study needs a new fit.

Tabular foundation models changed this picture for tables. TabPFN~\citep{hollmann2022tabpfn, hollmann2025tabpfnv2} and
TabICL~\citep{qu2025tabicl, qu2026tabiclv2} are transformers pretrained on millions of synthetic tasks drawn from a
prior; at test time, the training set of a new task is their input and they predict its test rows in one forward pass,
an approximation of the posterior predictive of the prior~\citep{muller2021pfn}. A simulation, however, is not a row of a table: its input is a mesh with fields,
its output is a field on that mesh, and meshes differ in size and connectivity across the study.

We present \physicl{}, an in-context learner whose rows are simulations: a single checkpoint, pretrained only on
procedural tasks, predicts the fields of unseen engineering studies at the nodes of each query's own unstructured mesh,
from labeled simulations whose geometry and mesh vary within the task, with no common grid and no weight update. We
propose four contributions:
\begin{itemize}[leftmargin=*, itemsep=1pt, topsep=2pt]
  \item \textbf{An architecture for ICL on meshes} (Section~\ref{sec:model}) that follows the three stages of
    TabICL~\citep{qu2025tabicl}, with mesh patches as cells, and adds a field path: query patches read context
    patches, then each query node reads context anchors. Every scale is fitted on the context, which makes predictions
    invariant to translations and scalings of the geometry and equivariant to affine changes of units. Variables are
    identified by episode-local random codes, so their counts are unbounded; each output field is decoded alone.
  \item \textbf{A procedural prior co-designed with the sampling, for online pretraining} (Section~\ref{sec:prior}):
    generic laws and boundary-element physical models on morphed template meshes, with variability within and across
    episodes. Meshes, their partitions into patches by recursive inertial bisection (RIB) and their sampling distributions are
    prepared once; during pretraining, CPU workers generate new episodes and evaluate their fields only at the points
    the model requests, never materializing full fields, at a cost nearly independent of the number of mesh nodes. The reported run drew 8.0\,M episodes in 34.8\,h, about 64 episodes/s, the GPU utilization of its four H200 at
    85--100\% with 80 CPU cores generating (Section~\ref{sec:exp-plaid}).
  \item \textbf{A cost bounded by spatial budgets}: each mesh is read through 1{,}024 patches pooled from a
    stratified subsample of its nodes; only the partition of a new mesh and the decoding of its nodes grow with it.
  \item \textbf{One checkpoint evaluated in and out of its prior} (Section~\ref{sec:experiments}): on
    \heldout{all}{8}{tasks} tasks of its prior over geometries it never saw, its error is half that of per-dataset
    baselines at $K = 8$ (median ratio \heldout{all}{8}{ratio}), at parity by $K = 64$; on five PLAID
    datasets representative of engineering settings, its field error decreases with $K$; on AhmedML, the transfer
    holds on surface meshes with a median of 1.1 million nodes. A separate experiment fine-tunes it on a procedural prior of one dataset.
\end{itemize}

\section{Related work}
\label{sec:related}

\paragraph{Prior-data fitted networks and tabular foundation models.}
Prior-data fitted networks~\citep{muller2021pfn} train a transformer on tasks drawn from a prior so that it approximates
the posterior predictive distribution; TabPFN~\citep{hollmann2022tabpfn, hollmann2025tabpfnv2, grinsztajn2025tabpfn25} made the idea
practical for tables. TabICL~\citep{qu2025tabicl} scales it with a column-then-row embedding (set attention across rows per
column, attention within a row with CLS tokens) followed by an ICL transformer over row tokens; TabICLv2
\citep{qu2026tabiclv2} adds quantile regression heads and a query-aware scalable softmax. \physicl{} reuses these three
stages and these heads, with patches instead of columns. PFNs and tabular foundation models are already used or
evaluated as surrogates on scalar physical and engineering data~\citep{yu2026fire, regenwetter2026engineering, tenachi2026tfmphysics, sharma2025dva};
\physicl{} carries the recipe from tables to fields on meshes. In-context learning was demonstrated at scale in language
models~\citep{brown2020gpt3} and studied on synthetic function classes~\citep{garg2022incontext}; neural processes~\citep{garnelo2018cnp, garnelo2018np, kim2019anp, nguyen2022tnp} predict the value at a query point
from context points and their values, the principle of the node-level decoding of \physicl{} (Section~\ref{sec:model}).

\paragraph{Neural operators and mesh surrogates.}
DeepONet~\citep{lu2021deeponet} and FNO~\citep{li2020fno} learn operators on fixed grids or sensor sets.
MeshGraphNets~\citep{pfaff2021meshgraphnets}, GINO~\citep{li2023gino}, Transolver~\citep{wu2024transolver},
UPT~\citep{alkin2024upt}, GAOT~\citep{wen2025gaot}, and CORAL and AROMA~\citep{serrano2023coral, serrano2024aroma},
which encode geometries by neural fields, handle unstructured meshes and varying geometry, and
AB-UPT~\citep{alkin2025abupt}, DoMINO~\citep{ranade2025domino} and GeoTransolver~\citep{adams2025geotransolver}
scale to automotive meshes of millions of cells, and UniField~\citep{zou2025unifield} to several aerodynamic
domains. All are supervised surrogates trained on labeled simulations, UniField jointly on several domains, and
cross-domain pretraining also transfers zero-shot to held-out CFD datasets~\citep{zhou2026smartic}; \physicl{} is
instead given the task in context. Our per-dataset baselines follow MMGP~\citep{casenave2023mmgp}, which morphs meshes
to a common one, reduces fields and shapes by PCA and regresses with Gaussian processes; they replace the morphing by
a shared topology or a fixed grid of probes. They are fitted on the context itself, in seconds to minutes, which is
the setting of \physicl{}: no training, or a training short enough to run on each new task.

\paragraph{Foundation models and in-context learning for PDEs.}
Multiple physics pretraining~\citep{mccabe2023mpp}, DPOT~\citep{hao2024dpot} and Poseidon~\citep{herde2024poseidon}
pretrain on many PDE datasets on regular grids, then fine-tune. ICON~\citep{yang2023icon} and
DeepOSets~\citep{chiu2025deeposets} learn operators in context from example pairs, on grids or point sets of one
domain, and PROSE~\citep{liu2023prose} jointly predicts operators and symbolic expressions from data and equation
inputs; VICON~\citep{cao2024vicon} tokenizes dense 2D fields into patches, as \physicl{} does meshes, its examples
pairs of frames of one field at times $t$ and $t+\Delta t$ on one regular grid. Zebra~\citep{serrano2024zebra} generates the trajectories of time-dependent parametric
PDEs autoregressively, conditioned on context trajectories on a regular grid; \physicl{} predicts stationary fields,
each simulation on its own mesh. PDE-PFN~\citep{park2026pdepfn} pretrains on numerical solutions or, in a second variant, on
approximate solutions of physics-informed networks, and predicts in context from coordinate--value observations of a target
instance; its fixed-weight results cover a family of convection--diffusion--reaction equations, while its
irregular-mesh Airfoil experiment updates the weights on Airfoil training simulations. In two controlled steady-state CFD problems, \citet[§4.4]{zhou2026smartic} show that a solved
simulation sharing a hidden parameter with the query improves prediction, a setting we also test (Section~\ref{sec:exp-plaid}).
In their joint SMART-IC experiment across six CFD datasets, a same-dataset context simulation did not improve
validation error (Appendix~C.5 of that work). ICM~\citep{li2026icm}, trained on the
governing equations without stress labels, infers in one forward pass a hyperelastic constitutive relation from observed
fields, generalizing to new materials, geometries and loadings; that relation then runs in a finite-element solver,
whereas \physicl{} predicts the simulation's field itself. HyPINO~\citep{bischof2025hypino} maps a PDE to the weights of a physics-informed network, trained on
manufactured solutions and on residual losses. PDEformer-2~\citep{ye2025pdeformer2} reads the PDE as a computational
graph and predicts zero-shot on equations that resemble its pretraining ones. \citet{medranonavarro2026gmp} pretrain unstructured 3D solvers without stored data (geometric
descriptors for steady problems, online synthetic PDE data for transient ones) as an initialization, then train on each target dataset; AREALS~\citep{chlon2026areals} pretrains
self-supervised on surface meshes, with geometric targets and no CFD field labels, then fine-tunes on simulations. To our knowledge, \physicl{} provides the first demonstration that a single model pretrained solely on
procedural tasks can use multiple labeled simulations to predict fields directly on the query mesh in unseen
engineering studies, without task-specific weight updates. Geometries and meshes can vary within a task, and each
output field is predicted in one forward pass.

\section{Problem setting}
\label{sec:setting}

A task is a hidden law $\mathcal{L}$ that maps a simulation's inputs to its outputs. A simulation $s$ carries a
mesh $\mathcal{M}_s = (X_s, E_s)$ with nodes $X_s \in \R^{N_s \times d}$ and elements $E_s$, input scalars
$\theta_s \in \R^{I}$, input fields $f_s : X_s \to \R^{F_i}$, and outputs: scalars $y_s \in \R^{O}$ and fields
$u_s : X_s \to \R^{F_o}$, with $(y_s, u_s) = \mathcal{L}(\mathcal{M}_s, \theta_s, f_s)$. Node counts $N_s$, connectivity and
shape differ between simulations of a task. An episode gives the context
$\mathcal{C} = \{(\mathcal{M}_k, \theta_k, f_k, y_k, u_k)\}_{k=1}^{K}$ and $Q$ query simulations
$\{(\mathcal{M}_q, \theta_q, f_q)\}_{q=1}^{Q}$. \physicl{} returns quantiles $\hat q_\tau(u_{q,j}(x))$ of one output field $j$ at
any requested nodes $x$ of each query, one output field per forward pass; scalar outputs $y_q$ are regressed on its row
tokens (Section~\ref{sec:model}).
Variable counts vary by task and are unbounded; variables are anonymous (no names are read). We consider only
stationary tasks.

\section{Model}
\label{sec:model}

\begin{figure*}[t]
  \centering
  \resizebox{\linewidth}{!}{\begin{tikzpicture}[
    font=\scriptsize, node distance=3mm and 4mm, >={Latex[length=1.6mm]},
    block/.style={draw=black!70, rounded corners=2pt, align=center, minimum height=6mm, inner sep=2pt, fill=white},
    trunk/.style={block, fill=blue!7},
    field/.style={block, fill=teal!9},
    rowe/.style={block, fill=violet!8},
    data/.style={block, draw=black!40, dashed, fill=black!2},
    head/.style={block, fill=orange!10},
    lab/.style={font=\tiny, text=black!65, align=center},
    flow/.style={->, black!70},
    label/.style={->, orange!75!black, thick},
  ]
  \node[data] (ctx) {context: $K$ meshes $\mathcal{M}_k$\\ $\theta_k$, $f_k$, labels $u_k$ (one field)};
  \node[data, right=of ctx, xshift=28mm] (qry) {queries: $Q$ meshes $\mathcal{M}_q$\\ $\theta_q$, $f_q$ (no labels)};
  \node[data, right=of qry, fill=violet!4] (codes) {random codes $r_j \sim \mathcal{N}(0, I_{256})$\\
    one per variable, drawn per episode\\ (fixed seed at inference)};
  \node[block, fill=black!4, below=4mm of ctx, xshift=21mm, minimum width=50mm] (scale)
    {scales fitted on the context only: frame $(c, s)$, $\mu, \sigma$ of $\theta$, $f$, $u$};

  \node[trunk, below=of scale, minimum width=50mm] (enc) {patch encoder (shared): SinCos $+$ pooled input fields $\to$ MLP\\
    $\to$ attentive pooling per RIB patch};
  \node[trunk, below=of enc, minimum width=50mm] (col) {$\mathrm{TF}_\mathrm{col}$: induced attention across simulations\\
    \textcolor{orange!75!black}{$+$ pooled patch moments of $u_k$, label pooling, projected row encoding}};
  \node[trunk, below=of col, minimum width=50mm] (row) {$\mathrm{TF}_\mathrm{row}$: blocks over [CLS, patches]
    $\to$ one row token};
  \node[trunk, below=of row, minimum width=50mm] (icl) {$\mathrm{TF}_\mathrm{icl}$: rows read [null] $\cup$ context rows\\
    \textcolor{orange!75!black}{$+$ row encoding; context: global label pooling}};

  \node[rowe, anchor=north east] (rowenc) at ([xshift=-6mm, yshift=-1mm]row.south west) {\textbf{row encoder} (scalar variables):\\
    one cell per variable of $\theta$, \textcolor{orange!75!black}{$\mu, \sigma$ of $u_k$}:\\
    $\mathrm{Lin}(v_{sj}) + W r_j + e_\mathrm{role}$\\
    $\to$ ISAB per variable (inducing: context only)\\
    $\to$ inter-variable attention $\to$ CLS readout};

  \node[rowe, right=6mm of col] (pool) {\textbf{variable pooling} (nodes, patches):\\
    tokens $t_j = W_x x_j + W_r r_j + \sum_f x_{jf} V_f r_j$\\ one learned query per head\\
    $\to$ $\sum_j \alpha_j t_j$};

  \node[data, below=4mm of rowenc] (shead) {\textbf{scalar outputs} (outside the model, no head):\\
    row tokens of context and queries, encoded\\ with no output observed\\
    $\to$ PCA fitted on the $K$ context tokens (99.99\%)\\
    $\to$ Kumo-Tabular in context on $[\theta, \text{scores}] \mapsto y$};

  \node[field, below right=5mm and -16mm of icl] (picl) {patch ICL: query patches\\ read context patches};
  \node[field, below=of picl] (nicl) {node ICL, per decoded node:\\ reads [null] $\cup$ anchors $\to$ MLP};
  \node[head, below=of nicl] (fhead) {\textbf{output read}: query $+\, W_o r_j$ reads [null] $\cup$ anchors\\
    $\to$ readout: 19 quantiles of $u_{q}(x)$};

  \node[data, right=6mm of nicl] (anch) {context anchors\\ $(x, f, $ \textcolor{orange!75!black}{pooled $u$}$)$\\ at area draws};

  \draw[flow] (ctx) -- (ctx |- scale.north);
  \draw[flow] (scale) -- (enc);
  \draw[flow] (qry.south) |- ([yshift=1.5mm]enc.east);
  \draw[flow] (enc) -- (col);
  \draw[flow] (col) -- (row);
  \draw[flow] (row) -- (icl);
  \draw[flow] (rowenc.east |- icl.west) -- (icl.west);
  \draw[flow, violet!70!black] (codes.south) -- (codes.south |- pool.north);
  \draw[flow, violet!70!black, dashed] (pool.west) -- (col.east);
  \draw[flow, violet!70!black, dashed] (pool.north west) -- (enc.east);
  \draw[flow, dashed] (row.west) -| ([xshift=-3mm]rowenc.north west) |- (shead.west);
  \draw[flow] (icl.south) ++(6mm,0) -- (picl.north);
  \draw[flow] (col.south east) ++(0,1mm) -- ++(4mm,0) |- (picl.east);
  \draw[flow] (picl) -- (nicl);
  \draw[flow] (anch) -- (nicl);
  \draw[flow] (anch.south) |- (fhead.east);
  \draw[flow] (nicl) -- (fhead);
  \draw[flow, violet!70!black, dashed] (pool.south) -- (anch.north);

  \draw[label] ([yshift=1.5mm]ctx.west) -| (rowenc.north);
  \draw[label] ([yshift=-1.5mm]ctx.west) -- ++(-4mm,0) |- (col.west);
\end{tikzpicture}}
  \caption{\textbf{\physicl{}.} Context and queries share the patch encoder (blue). Variables are
    identified by episode-local random codes $r_j$, fused at nodes and patches by attention pooling over variable
    tokens, and in each simulation's scalars by the row encoder (violet). Context labels (orange text) enter per patch
    before \tfcol{}, per simulation through the row encoder, and through the anchors.
    The field path (green) reads context patches, then context anchors; the output field reads the anchors once more
    through a query carrying its code (orange block). Scalar outputs are not predicted by the network: a Kumo-Tabular~\citep{qu2026kumotabular} regression
    on $\theta$ and the context PCA of the row tokens, encoded with no output observed (dashed, outside the model).
    Scales are fitted on the context only; queries never attend to each other.}
  \label{fig:architecture}
\end{figure*}
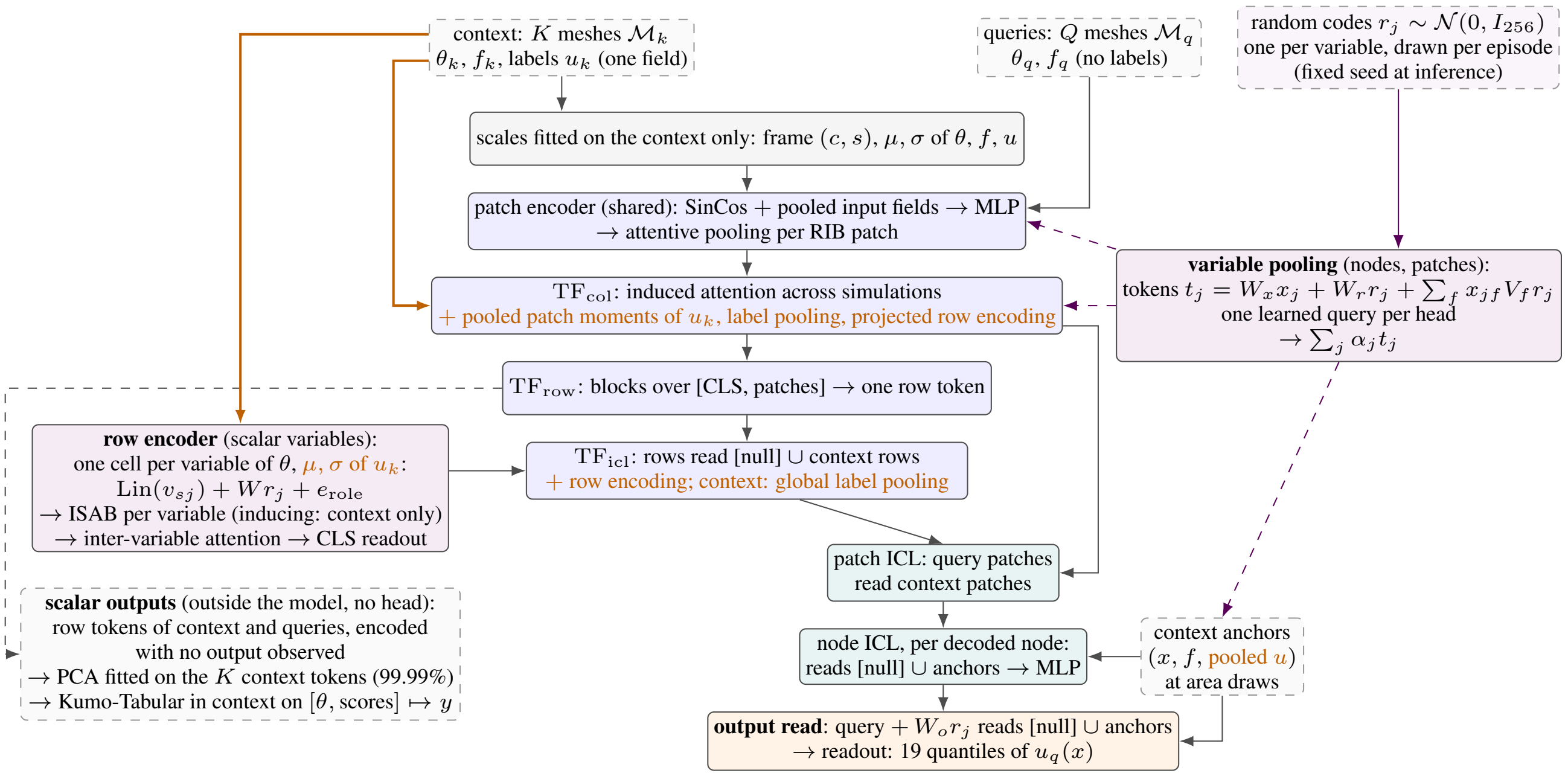

The model has 28.4M parameters (Table~\ref{tab:params}) and serves 2D and 3D meshes alike (planar points enter
with $z = 0$). Figure~\ref{fig:architecture} summarizes the forward pass, and
Appendix~\ref{sec:design} gives, for each choice, the alternative it was preferred to and the ablations behind it.
Patch tokens have width 256, row tokens 512 (two CLS tokens).

\paragraph{One output field per task.}
Every training task has one output field and no output scalar; at inference, each output field of a dataset is
decoded alone, with one encoding of the context per field. Decoding each field alone keeps a heavy-tailed field from
contaminating a smooth one through shared pooled paths; the cost is the information that the other outputs of the context
carry (Section~\ref{sec:discussion}). Input fields and input scalars keep a variable, unbounded count, identified by
episode-local random codes rather than by a table of slots, which would bound the count by training.

\paragraph{Context-fitted scales.}
Coordinates are mapped to the context's frame, $[-1, 1]^d$ from the axis ranges of the context nodes; scalars are
standardized over the context simulations, and fields by their area-weighted moments on context quadrature nodes,
with relative floors; predictions are mapped back. Predictions are thus invariant to translations and scalings of the
geometry and equivariant to affine unit changes of every feature, up to clips, floors and a constant-context fallback
(Appendix~\ref{app:scales}; unit-tested). Rotations are not removed: each training episode is drawn in one random
rotated frame, in which a vector or tensor component shown alone is a scalar.

\paragraph{Patch encoder and \tfcol{}.}
Each mesh is partitioned into 1{,}024 disjoint patches by recursive inertial bisection (RIB), as MeshMAE splits a
mesh into patches of faces~\citep{liang2022meshmae} and the Vi-Transf baseline of PLAID into METIS partitions, one
token each, a projection of the partition's nodal features~\citep{casenave2025plaid}. The encoder
sees a stratified subsample, up to 2{,}048 boundary nodes plus 8 nodes per patch, each weighted by its inverse
inclusion probability $w$. A node gets a sine-cosine embedding of its position~\citep{mildenhall2020nerf,
tancik2020fourier} through an MLP, plus a vector summarizing its input fields, an attention
pooling over one token per field, each identified by its random code (Appendix~\ref{sec:design}).
A patch token is a learned weighted mean of its node embeddings (attentive pooling,
RandLA-Net,~\citealp{hu2020randlanet}). Context patches add an encoding of
per-patch moments of the output field. These counts of encoder nodes per patch and of boundary nodes, with the
counts of anchors, quadrature and decoded nodes below, are the spatial budgets of a simulation
(Appendix~\ref{app:budgets}). \tfcol{} is the column-wise induced set attention block (ISAB) of
TabICL~\citep{qu2025tabicl, lee2018settransformer}, run across the patches of the task's simulations instead of the
rows of a column: as in TabICL, only the context feeds its 64 learned vectors, which every context and query patch
then reads back, so queries see a summary of the context, never each other; each patch's position is added to it before
the query and key projections, not the values, so that the learned vectors can specialize to regions.

\paragraph{\tfrow{} and \tficl{}.}
\tfrow{} follows the row-wise stage of TabICL, which prepends CLS tokens~\citep{dosovitskiy2021image} to a row's
column embeddings and concatenates their outputs into the row embedding~\citep{qu2025tabicl}, with patches in place
of columns: four blocks attend over two CLS tokens and the patches of each simulation, the two CLS outputs,
concatenated, form the row token $r_s \in \R^{512}$, and the patch outputs are the keys of patch ICL. \tficl{} is the ICL transformer of
TabICL: context rows add the row encoding of $\theta_k$ and of the per-simulation moments of the output field, as
TabICL adds a label embedding to its training rows, and query rows that of $\theta_q$. Six transformer blocks let
every row attend to a null token, a learned token that keeps the attention defined without
context and lets a query take nothing from it, and to the context rows only, the training-rows-only mask of TabPFN and
TabICL~\citep{hollmann2022tabpfn, qu2025tabicl}, with the query-aware scalable softmax of
TabICLv2~\citep{qu2026tabiclv2, nakanishi2025ssmax}, so that attention does not dilute as $K$ grows.

\paragraph{Field path.}
Patch ICL: query patches, keyed by their simulation's row token after \tficl{}, attend to every context patch
(four blocks). Node ICL: a decoded query node, from its position, input fields, patch and row tokens, attends
to a null token and to the context anchors, 64 area-drawn nodes per context simulation encoded with their
position, input and output fields (two blocks), then an MLP. A last read of the anchors, by a query to which the
output field's code $r_j$ is added, and a readout give 19 non-crossing quantiles. Nodes are decoded independently, in chunks, so inference memory is bounded
by the chunk.

\paragraph{Scalar outputs.} No scalar head: a tabular regression on the row tokens (Appendix~\ref{sec:design}).

\paragraph{Loss and training.}
The pinball loss~\citep{koenker1978quantile} over 19 evenly spaced levels, in units of the output's context
standard deviation. TabICLv2 trains its quantile head the same way (999 levels,~\citealp{qu2026tabiclv2}), whereas
TabPFN predicts a distribution over bins~\citep{muller2021pfn, hollmann2025tabpfnv2}. Query nodes are drawn in
proportion to area, so the mean loss estimates the area-weighted loss without bias. Point predictions are the mean
of the quantiles. Training draws $K \in [8, 64]$ context simulations and 4
queries per episode, 32 episodes per step on 4 H200 GPUs, for 250{,}000 steps (Appendix~\ref{app:exp-details}).

\section{Procedural prior}
\label{sec:prior}

\begin{figure*}[t]
  \centering
  \IfFileExists{figures/prior_current.pdf}{\includegraphics[width=\linewidth]{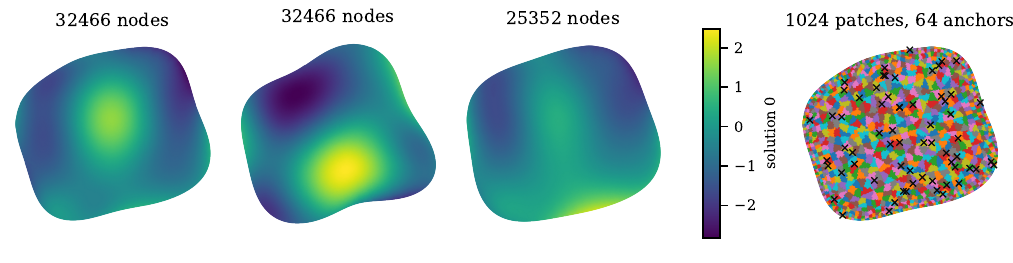}}{%
    \fbox{\parbox{0.9\linewidth}{\centering\todo{one episode of the pretraining configuration (paper/scripts/prior\_figure.py
    on a cluster bank): three context simulations of one task, one template morphed by its designs, three meshes;
    the RIB patches and the anchors of the first.}}}}
  \caption{\textbf{One task of the procedural prior.} Three context simulations of one task (a generic law): one
    template morphed by each simulation's design variables, on a mesh drawn among the template's 12 (the first two
    draw the same one); the output field varies with the drivers. Right: the RIB patches of the first mesh and its 64 anchors,
    the context nodes whose output values node ICL reads.}
  \label{fig:prior}
\end{figure*}

\physicl{} is pretrained only on a procedural prior: tasks generated on the fly while the model trains, by
generic laws and fast physical models on morphed template meshes, each case evaluated only at the points the model
samples and never stored. This design is what makes the pretraining possible: once its meshes are prepared, an episode costs seconds of one
CPU core whatever their size, and the data workers kept four GPUs busy
(Section~\ref{sec:exp-plaid}). The
prior library owns the laws and the variability within a task; \physicl{} owns the
curriculum, the spatial budgets and one random rotation per episode.

\paragraph{Geometries and prepared banks.}
Seven generators each draw tasks on one family of geometries: in 2D, star-shaped unions of polygons and
superellipses, non-star bodies (airfoils, ellipses, polygons with holes) and annular domains; on 3D surfaces and in
3D volumes, unions of convex primitives morphed by spherical harmonics, and constructive solids, wings, twisted blades
and ducts; each as an interior domain or as a body in a box. What is precomputed is the geometry, not the fields:
each generator has a bank of 1{,}000 templates $\times$ 12 meshes of 20{,}000 to 50{,}000 nodes, each mesh stored with
its wall tags, its RIB partition into 1{,}024 patches and its area or volume distribution. The seven banks were built
once, in 64\,min on 80 CPU cores (181\,GB), and are shared by every run. A morph applies the design variables (boundary harmonics on star
shapes, rigid motions and stretches per body otherwise, diffeomorphic by construction).

\paragraph{Laws.}
Every task has one output field and draws its law, which fixes how the outputs depend on the inputs. On the
geometries that allow one (2D wall loops, wings and blades), the law is a physical model with probability 1/2,
else a generic law. A generic law combines modes of one of eleven basis families (spectral, Gaussian-process
random features, random MLP, harmonic, Poisson, elasticity, fronts, advection--diffusion, Helmholtz, singular,
wavelet), ambient or attached to the wall and the shape through the signed distance, with coefficients linear in the
drivers (free conditions, design variables, input-field latents, shape features) and a power-law spectrum. A
physical model solves a linear PDE on the drawn domain by a boundary-element (panel) method and adds classical
engineering corrections: potential flow with compressibility, shocks and boundary layers, half of it with no-slip
viscous layers and an eddy viscosity; plane elasticity; Darcy flow with wells; torsion; heat conduction; Stokes flow;
and, on 3D wings and blades, strip theory with a lifting line. Their fields are physically shaped, not CFD- or
FEM-accurate (Appendix~\ref{app:prior}). The laws are deterministic: every output is a function of what the model
sees.

\paragraph{Variability within and across episodes.}
Within an episode, the template and the law, with its coefficients, stay fixed, as do the wall condition (free,
or forced to zero at the wall), which inputs vary and over which sub-ranges, and the rotation. Each simulation applies
its own morph and draws one of the template's 12 precomputed meshes, so the shape, the node count and the connectivity
vary within a context (Figure~\ref{fig:prior}). Each simulation also draws its own values of three kinds of
variables: one to three operating conditions, independent or correlated in half of the tasks; the design variables of
its morph, given to the model as input
scalars in a fifth of the tasks and otherwise seen only through the resulting shape; and the latent coefficients from
which its input fields are generated. Across episodes, the generator, the template and the law are drawn anew, and so are the counts of input scalars (up to 12, plus 4
design variables) and input fields (up to 12, plus the signed distance or the normals). The mix weighs the three 2D generators 0.2 each and the four 3D ones 0.1 each, with
$K \in [8, 64]$ context simulations and 4 queries per episode.

\paragraph{A cost independent of the mesh size.}
An episode is lazy: its fields are closed forms evaluated only at the $m$ points the model samples (about 2{,}000 to
20{,}000 per simulation, set by the spatial budgets), and the morph and the selection cost $O(m)$. A generic field
costs $O(m)$ per mode; a physical model $O(N_w^2)$ for its matrix and $O(N_w^3/3)$ for its dense solve, with $N_w$ from 64
to a few hundred wall panels, then $O((m + 256)\,N_w)$ for its fields, at the $m$ points and at 256 points that calibrate the
field's scale. After preparation, at fixed spatial budgets and panel count, none depends on
the $N$ nodes of the mesh, only on the sampled points: from $2\cdot10^4$ to $10^6$ nodes, an episode's cost grows by a
factor 1.01 to 1.20 (preparation, cold opening, validity checks and rejection fallbacks are counted apart,
Appendix~\ref{app:budgets}). A case costs 10--150\,ms on one core,
an episode at $K = 64$ 0.1--1\,s for generic laws and 0.5--12\,s for physical models. During training, 72 data workers, 18 per GPU, on 80 CPU cores build the
next episodes while the model trains on the current ones.

\section{Experiments}
\label{sec:experiments}

We evaluate one pretrained checkpoint, unchanged: its last step, which also has the best validation score on
tasks of the prior (Appendix~\ref{app:exp-details}). It is compared with the baselines of Section~\ref{sec:exp-baselines} and predicts in context, first in
distribution,
on tasks of its own prior over geometries it never saw (Section~\ref{sec:exp-heldout}), then out of distribution, on
five PLAID datasets of engineering simulations (Section~\ref{sec:exp-plaid}) and on large 3D surface meshes
(Section~\ref{sec:exp-ahmed}). A distinct experiment fine-tunes it on a procedural prior of one dataset
(Section~\ref{sec:exp-finetune}). Every
evaluation decodes each output field alone and regresses the scalar outputs on the row tokens.

\paragraph{Metrics.} The NMSE of an output field on a query is its area-weighted squared error over every node of the
query mesh, divided by the field's area-weighted variance over the context, so the context mean scores about 1; a
reported value is the median over the queries. Scalar outputs are scored by $R^2$ over the queries. A method's time
is split into fit, its work from the context alone, and query, that of one more query, the partition of
a new mesh included; a method is on the accuracy--time front if no other is both faster and more accurate.

\subsection{Baselines}
\label{sec:exp-baselines}

Every dataset or task is also fitted on the same $K$ context simulations, without training a neural surrogate on the task, by
per-dataset baselines: the
context mean; a PCA of the node coordinates as shape features when the meshes share one topology
(\texttt{pca\_*}), else a PCA of a signed distance on a fixed grid of probes (\texttt{probe\_*}); each with a
Gaussian-process (\texttt{\_gp}) or Kumo-Tabular~\citep{qu2026kumotabular} (\texttt{\_kumo}) regression of the PCA
scores of every output field, and the nearest context simulation in the inputs (\texttt{\_nn}) as a lookup control.
The baselines are tuned on each dataset or task: their settings (PCA variance, Kumo size and estimators, GP training budget)
are swept and the best one kept, selected on the queries themselves (an oracle selection, which favors the
baselines), up to $K = 64$ and kept beyond; the probe grid is checked against finer grids by
their PCA floor, the projection of the query truth on the context modes (Appendix~\ref{app:comparisons}); their fits are
parallelized and cached. \physicl{} is never tuned per dataset: its scalar regression has one setting, fixed in
advance, and its architecture was frozen before pretraining. Appendix~\ref{app:comparisons} gives their
definitions, these choices and their limits; a supervised reference trained per dataset, Transolver-3, is compared
on the PLAID competitions in Table~\ref{tab:competitions}.

\subsection{Tasks of the prior on unseen geometries}
\label{sec:exp-heldout}

We first measure the model in distribution, on tasks of its own prior over geometries it never saw. We draw 32 new templates per pretraining generator, with another seed than the
1{,}000 of pretraining. On them, 28 tasks are fixed before any prediction: four for each of seven groups of physical
models (plate, heat, and flow or Darcy, each on the radial and shapes generators, and strip theory on 3D surfaces).
Each task is evaluated like a dataset of Section~\ref{sec:exp-plaid}, with $K$ from 8 to 64, 64 queries and the
baselines of Section~\ref{sec:exp-baselines}, tuned per task; \physicl{} reads one RIB partition per mesh ($M = 1$; Section~\ref{sec:exp-plaid} averages $M$ partitions).

\begin{figure}[tb]
  \centering
  \IfFileExists{figures/heldout_k.pdf}{\includegraphics[width=\linewidth]{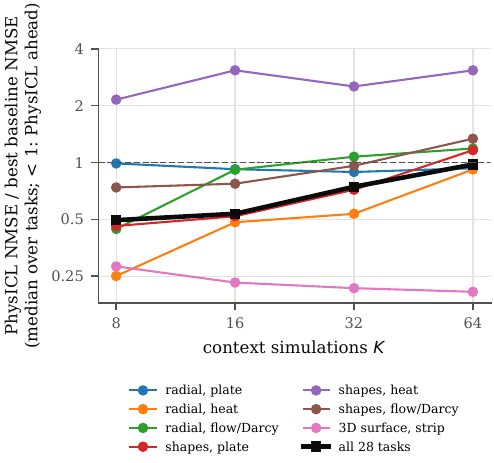}}{\todo{run paper/scripts/heldout.py}}
  \caption{\textbf{Tasks of the prior on unseen geometries}: \physicl{}'s median NMSE over that of the best
    Gaussian-process or Kumo setting (chosen on the queries, per task and $K$), median over each group's four
    tasks, and over all 28 (black); below 1, \physicl{} ($M = 1$, one setting) is more accurate.}
  \label{fig:heldout}
\end{figure}

\physicl{}'s error is half that of the best Gaussian-process or Kumo setting at $K = 8$ (median ratio
\heldout{all}{8}{ratio} over the 28 tasks), ahead on \heldout{all}{8}{ahead} of them and on the accuracy--query-time
front of \heldout{all}{8}{onfront}; the baselines close the gap as $K$ grows, to parity at $K = 64$ (median ratio
\heldout{all}{64}{ratio}, ahead on \heldout{all}{64}{ahead}, on the front of \heldout{all}{64}{onfront};
Figure~\ref{fig:heldout}, Table~\ref{tab:heldout}; predicted fields in Figure~\ref{fig:heldout-fields}). The groups differ: strip theory on 3D surfaces stays about five
times more accurate (\heldout{shapes_surface_strip_strip}{64}{ratio} at $K = 64$), while heat on shapes domains
favors the baselines at every $K$ (\heldout{shapes_heat}{64}{ratio}). A pretrained model thus brings most where the
context is small, the regime of a new design study.

\subsection{In-context prediction on PLAID datasets}
\label{sec:exp-plaid}

We then measure the model out of distribution, on five PLAID datasets~\citep{casenave2025plaid} of
engineering simulations, none seen in pretraining.

\paragraph{Setting.} %
The pretraining ran 250{,}000 steps (8.0\,M episodes) in 34.8\,h on 4 H200 GPUs (139 GPU-hours), about 64 episodes/s;
in this run the GPU utilization, read on the cluster's monitoring, stayed at 85--100\% while 80 CPU cores generated
the episodes (training curves in Appendix~\ref{app:training-runs}). The five datasets (Table~\ref{tab:datasets}) are
2D\_profile (flow around airfoils), 2D\_Multiscale\_Hyperelasticity (2D\_MSHE, perforated cells under finite strain),
Tensile2d (plates in tension), VKI-LS59 (a turbine cascade) and Rotor37 (compressor blade surfaces, 3D); none was seen
in pretraining, though the prior covers similar geometries and fields (Appendix~\ref{app:prior}). Contexts and
queries come from the train split, as PLAID hides its test labels: $K$ from 8 to 64, the training range, and 125 and
250 in extrapolation, with the same queries for every $K$ and method: 250, or 150 on 2D\_profile, whose train split
of about 300 simulations allows $K$ up to 125 only.

\begin{figure*}[t]
  \centering
  \IfFileExists{figures/icl_k.pdf}{\includegraphics[width=\linewidth]{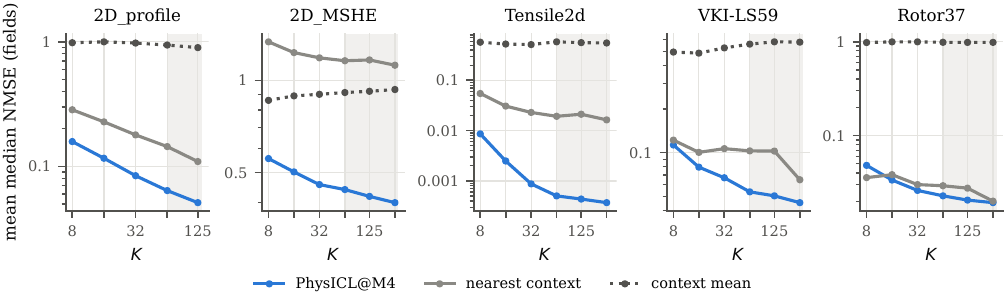}}{\todo{figure}}
  \caption{\textbf{In-context learning on five PLAID datasets.} Mean over the output fields of the median NMSE of
    \physicl{} ($M = 4$ partitions), of the nearest context simulation in the inputs (its best setting) and of the
    context mean, against the number $K$ of context simulations; no weight is updated. Shaded, $K > 64$: beyond the
    training context sizes.}
  \label{fig:icl-k}
\end{figure*}

\paragraph{In-context learning.} \physicl{} averages its predictions over $M = 4$ RIB partitions of each mesh, which removes the
per-patch offsets of a single partition for about $M$ times the time ($M = 1$ and 8 also appear on the fronts). Figure~\ref{fig:icl-k} is the main evidence of the transfer: the same checkpoint,
never updated, sees only its context grow. Its mean NMSE over the output fields decreases monotonically from $K = 8$ to 64 on \plaid{all}{64}{decreasing} of the \plaid{all}{64}{datasets}
datasets (Figure~\ref{fig:icl-k}), and from $K = 32$ it is below the nearest context simulation on every dataset and each
of three context draws (13 of 15 dataset--draw pairs at $K = 16$):
\physicl{} infers more from its context than which simulation the query resembles. Appendix~\ref{app:per-dataset} shows the predicted fields of one query per dataset.

\paragraph{Does it use the pairing of inputs and outputs in its context?} A larger $K$ brings more input--output
examples, but also a wider geometric coverage and more accurate context statistics. To isolate the examples, we
shuffle one part of each context simulation among the context simulations, and keep everything else unchanged
(Table~\ref{tab:context-control}): with permuted outputs, each context mesh carries the output field of another
context simulation, which needs meshes of one shared topology (Rotor37 and VKI-LS59); with permuted inputs, each
context simulation carries the input scalars of another. The set of context fields, its statistics, the meshes,
the queries and the weights stay the same; only which output goes with which input changes. Permuting the outputs
multiplies \physicl{}'s error by \ctrl{Rotor37}{outputs}{8} and \ctrl{VKI-LS59}{outputs}{8} at $K = 8$, and by
\ctrl{Rotor37}{outputs}{250} and \ctrl{VKI-LS59}{outputs}{250} at $K = 250$; permuting the inputs, by
\ctrl{Rotor37}{inputs}{8} and \ctrl{VKI-LS59}{inputs}{8} at $K = 8$, and by \ctrl{Rotor37}{inputs}{250} and
\ctrl{VKI-LS59}{inputs}{250} at $K = 250$: \physicl{} uses the pairing. The ratio grows with $K$ for permuted outputs
only: more examples help much more when their outputs are correctly paired.

\begin{figure}[t]
  \centering
  \IfFileExists{figures/heading_wake.pdf}{\includegraphics[width=\linewidth]{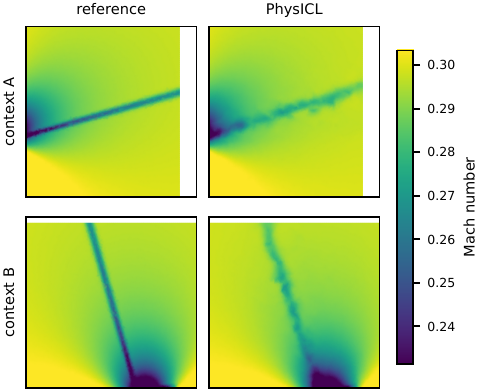}}{\todo{run
  paper/scripts/heading\_swap.py}}
  \caption{\textbf{Hidden free-stream direction.} One query (incidence $7.7^\circ$, $K = 8$, \physicl{} at $M = 8$ here; $M = 1$ in Appendix~\ref{app:heading-cp}), with the contexts
  of laws A (top) and B (bottom), whose hidden free-stream direction differs by $90^\circ$; same geometry, incidence
  and checkpoint, one Mach scale.}
  \label{fig:heading-wake}
\end{figure}

\paragraph{Does it read a hidden law from its context?} Two contexts A and B of $K = 8$ flow simulations on one
procedural geometry differ only in the free-stream direction, turned by $90^\circ$ and shown by no input;
in this qualitative test of one query, the query, the context geometry, incidences and meshes, and the checkpoint are the same, so the contexts' Mach fields
are the only model input that changes. \physicl{} turns its predicted wake with its context, with a more diffuse
Mach deficit than the reference (Figure~\ref{fig:heading-wake}). On the pressure coefficient $C_p$
(Appendix~\ref{app:heading-cp}), a quantitative test of the same intervention on eight templates, every method,
\physicl{} included, follows the hidden free-stream direction on every query; the appendix compares their accuracy.

\begin{figure*}[t]
  \centering
  \IfFileExists{figures/fronts_fields_K64.pdf}{\includegraphics[width=\linewidth]{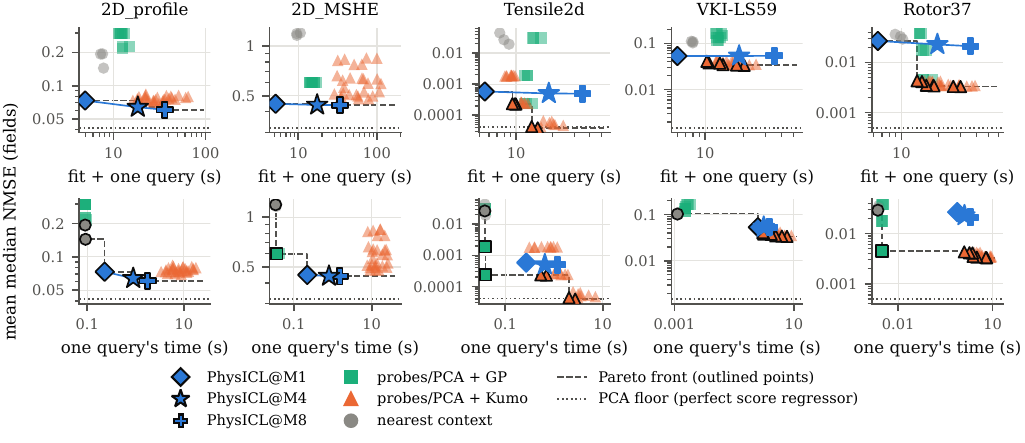}}{\todo{figure}}
  \caption{\textbf{Accuracy against time} at $K = 64$. Mean over the output fields of the median NMSE, against the
    fit plus one query, the cost of a new task (top), and against the time of one query (every output; bottom). Pale: every swept baseline setting; outlined: the Pareto front; \physicl{} at $M = 1$, 4 and 8 partitions,
    linked in $M$ order; dotted, the PCA floor of the most accurate truncation. Per-field numbers and predicted
    fields of every dataset: Appendix~\ref{app:per-dataset}.}
  \label{fig:fronts}
\end{figure*}

\paragraph{Comparison with per-dataset baselines.} Figure~\ref{fig:fronts} and Table~\ref{tab:plaid} place
\physicl{} against the baselines of Section~\ref{sec:exp-baselines} at $K = 64$. The comparison favors the baselines:
their settings are swept and selected on the queries themselves, their PCA is fitted on each dataset's own fields,
and they read every context field whole (every node, or $2^{14}$ probes), where \physicl{}, at one fixed setting,
reads it through its spatial budgets. On 2D\_profile and 2D\_MSHE, whose fields change shape with the geometry, \physicl{}'s error is
\plaid{2D_profile}{64}{ratio} and \plaid{2D_MSHE}{64}{ratio} times that of the best baseline; on 2D\_MSHE the gain comes from the energy density (Table~\ref{tab:fields}). On Tensile2d
and Rotor37, whose outputs lie close to a low-dimensional linear space, a reduced-order model is
\plaid{Tensile2d}{64}{ratio} and \plaid{Rotor37}{64}{ratio} times more accurate, and VKI-LS59 lies between
(\plaid{VKI-LS59}{64}{ratio}). \physicl{} answers a query in about a second, faster than the best baseline on every
dataset at this $K$. A Transolver-3~\citep{zhou2026transolver3} trained separately on each dataset is a supervised
reference: it is \competition{all}{min} to \competition{all}{max} times more accurate than \physicl{} with fixed
weights on the PLAID competitions (Table~\ref{tab:competitions}). In-context learning is thus on par with tuned reduced-order models where the fields are far from a
low-dimensional linear space, and behind them where such a space is nearly exact.

\paragraph{Beyond the training context sizes.} At $K = 125$ and 250 (the baselines at their best setting of $K = 64$),
its error still decreases, more slowly, and it stays on the front of \plaid{all}{250}{onfront} of
\plaid{all}{250}{datasets} datasets at $K = 250$.

\begin{figure}[tb]
  \centering
  \IfFileExists{figures/calibration_K64.pdf}{\includegraphics[width=0.49\linewidth]{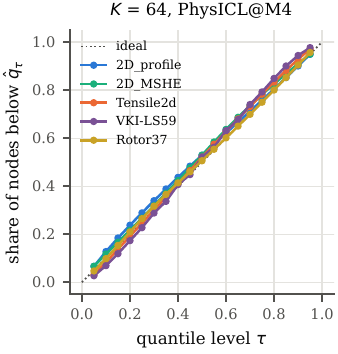}}{\todo{PLAID}}\hfill
  \IfFileExists{figures/finetune_calibration.pdf}{\includegraphics[width=0.49\linewidth]{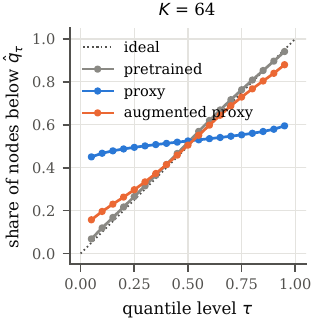}}{\todo{2D\_MSHE}}
  \caption{\textbf{Calibration} at $K = 64$: share of nodes whose truth is at most each of \physicl{}'s 19 field
    quantiles (area-weighted, pooled over the nodes, output fields and queries: a marginal calibration); the diagonal
    is ideal. Left: the PLAID
    datasets; right: 2D\_MSHE, pretrained and its two fine-tunings (Section~\ref{sec:exp-finetune}).}
  \label{fig:calibration}
\end{figure}

\paragraph{Calibration.} Pooled over nodes, output fields and queries, \physicl{}'s field quantiles are close to
calibrated on every PLAID dataset: the share of nodes below each of its 19 quantiles follows the quantile
level (Figure~\ref{fig:calibration}, left), and its 5--95\% band covers \plaid{all}{64}{covmin} to
\plaid{all}{64}{covmax} of the nodes at $K = 64$ per dataset, \plaid{all}{64}{covfieldmin} to
\plaid{all}{64}{covfieldmax} per output field, for a nominal 0.9.

\begin{figure}[tb]
  \centering
  \IfFileExists{figures/scaling.pdf}{\includegraphics[width=\linewidth]{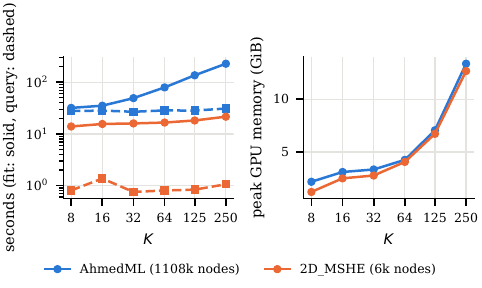}}{\todo{run paper/scripts/scaling.py}}
  \caption{\textbf{Cost against $K$}, AhmedML surfaces against 2D\_MSHE (same model, same spatial budgets): fit and one
    query's time (left) and peak GPU memory of \physicl{}'s predictions (right).}
  \label{fig:scaling}
\end{figure}

\subsection{Scaling to large 3D surface meshes: AhmedML}
\label{sec:exp-ahmed}

The AhmedML surfaces~\citep{ashton2024ahmedml} (pressure, pressure coefficient, wall shear stress and $y^+$ fields) test
a transfer with fixed weights along two axes. The resolution gap is measured: query meshes of
\plaid{AhmedML}{64}{knodes}k nodes (median), against 20k to 50k in pretraining. The geometric gap is qualitative and
concerns the study as a whole: the vehicle shapes vary within it, whereas the shapes of a pretraining episode derive
from one template. A query may still have a close neighbor in its context; the nearest-context control, which reuses
the field of the closest simulation, has a mean field NMSE of \plaid{AhmedML}{64}{nn} at $K = 64$, against
\plaid{AhmedML}{64}{physicl} for \physicl{}. The peak GPU memory of its predictions barely depends on the mesh: at $K = 64$,
\plaid{AhmedML}{64}{peak}\,GiB on AhmedML against \plaid{2D_MSHE}{64}{peak}\,GiB on the \plaid{2D_MSHE}{64}{knodes}k
nodes of 2D\_MSHE (Figure~\ref{fig:scaling}). The fit takes \plaid{AhmedML}{64}{fit}\,s and grows with $K$; a query,
partition and decoding included, takes about \plaid{AhmedML}{64}{query}\,s at every $K$. At
$K = 64$, \physicl{} is the most accurate method on every output field, with a shorter fit and query than every timed baseline
(nearest context, GP, Kumo): its mean field NMSE is
\plaid{AhmedML}{64}{ratio} times that of the best baseline (Kumo on probes; $\pm$\plaid{AhmedML}{64}{spreadpct}\% over three draws), whose query takes
\plaid{AhmedML}{64}{bestquery}\,s. At $K = 250$, that baseline becomes more accurate (\physicl{}'s error is \plaid{AhmedML}{250}{ratio} times its) but fits in \plaid{AhmedML}{250}{bestfit}\,s against \plaid{AhmedML}{250}{fit}\,s, and \physicl{} stays
on the front. Per field: Table~\ref{tab:fields}, Figures~\ref{fig:fields-AhmedML} and~\ref{fig:ahmed-3d}.

\subsection{Fine-tuning on 2D\_Multiscale\_Hyperelasticity}
\label{sec:exp-finetune}

\paragraph{Why and how.} Within an episode, the pretraining prior only deforms one template smoothly, whereas the
5--8 equal holes of 2D\_MSHE appear and move from one simulation to the next; the pretrained model predicts its stresses flat around the holes, where they peak.
A procedural prior built for one dataset, its proxy, is cheap to make (Appendix~\ref{app:finetune-prior}): the \texttt{perforated}
generator and its \texttt{rve} model take the dataset's geometry, loads (the imposed loading $C$, given as input scalars) and kinematics, but solve a linear problem
where the dataset is hyperelastic. The proxy thus encodes knowledge of the dataset (its geometry, its kinematics
and the ranges of its loads, measured on the train split) but fits no parameter to its outputs. From the pretrained weights, we
fine-tune for 50{,}000 steps on one H200, with larger spatial budgets since the stress peaks cover less than 0.1\% of
the domain (Appendix~\ref{app:exp-details}). We report the checkpoint that scores best on the proxy's own
validation tasks, not on 2D\_MSHE.

\paragraph{Models and protocol.} (a) The pretrained model; (b) the model fine-tuned on the proxy alone; (b$'$) the model
fine-tuned on an augmented proxy, the same geometries with the \texttt{rve} model in 3/8 of the tasks, the other physical
laws of the prior in 3/8 and generic laws in a quarter; (b$'$) keeps the context informative (see Results). The protocol is that of Section~\ref{sec:exp-plaid},
with three context draws, each model with the wall inputs it was trained with, every model at $M = 4$; each fine-tuning is one
fine-tuning run, so the spreads reported cover the context and query draws, not the training seed. A near-hole NMSE,
on the nodes within half a hole radius of a hole wall and normalized by the query's variance there (about 1 for a
flat prediction), shows the region the fine-tuning targets; the coverage of the 5--95\% band measures its
calibration.

\paragraph{Results.} Table~\ref{tab:finetune} and Figures~\ref{fig:finetune-k-fields} and~\ref{fig:calibration}
(right) give the results, and Figure~\ref{fig:fields-2D_MSHE-arms} shows the fields of one query for each model.
Both fine-tunings divide the stress error at $K = 64$, by \ft{proxy}{64}{stressgain} on the proxy alone and
\ft{augmentedproxy}{64}{stressgain} on the augmented proxy, and more near the holes, where the pretrained model predicts flat
stresses (near-hole NMSE \ft{pretrained}{64}{stressnear}): by \ft{proxy}{64}{stressneargain} and
\ft{augmentedproxy}{64}{stressneargain}. Both fine-tunings fall below the best baseline (\ft{best}{64}{stress}, mean over the four
stress components) and below the PCA floor (\ft{floor}{64}{stress}), the oracle projection of the query's truth on the $K$
context modes: the holes move within the dataset, so a linear space of the context fields cannot place a new query's stress
peaks. The two fine-tunings differ in how they use the context (Figure~\ref{fig:finetune-k-fields}). On the proxy
alone, the map from the geometry and $C$ to each field is the same in every task, so the context only tells which of the proxy's fields is
predicted, and its scales; the model learns that map in its weights and ignores the context: its stress error stays near \ft{proxy}{64}{stress} whatever $K$. That map is the
linear proxy, not the dataset's hyperelastic law, so the remaining error cannot be corrected from the context. The
augmented proxy mixes other laws on the same geometries, so the context again tells which law applies, and the
model keeps reading it: its stress error keeps falling with $K$, from \ft{augmentedproxy}{16}{stress} to
\ft{augmentedproxy}{250}{stress}, the dataset's own simulations correcting the proxy's law. The proxy alone also loses the calibration: its 5--95\% band covers
\ft{proxy}{64}{fieldscov} of the nodes, against \ft{pretrained}{64}{fieldscov} before fine-tuning
(Figure~\ref{fig:calibration}, right), a deterministic proxy leaving no spread to learn. The augmented proxy keeps the
stress bands calibrated (\ft{augmentedproxy}{64}{stresscov}) but narrows those of the displacements
(\ft{augmentedproxy}{64}{displacementcov}), on which the best baseline overtakes it at $K = 250$. Each fine-tuning costs about a
tenth of the pretraining's GPU-hours (training details in Appendix~\ref{app:exp-details}).

\section{Limitations and perspectives}
\label{sec:discussion}

\paragraph{Limitations.} \physicl{} decodes each output field alone and does not use the other output fields of the context. It delegates scalar outputs to a tabular regressor on its tokens, which the per-dataset
baselines beat on the five PLAID datasets, though not on AhmedML (Table~\ref{tab:scalars}). It reads each context
simulation through spatial budgets (Section~\ref{sec:model}, Appendix~\ref{app:budgets}),
and patch ICL is quadratic in the number of patches. Vector and tensor outputs are learned one component at a time,
in a rotated frame. The prior's physical models are panel solvers and analytic laws, not finite-element solutions;
in 3D, only wings and blades have one (strip theory), every other 3D task drawing a generic law; and its generic
laws solve no boundary-value problem. Within an episode, shapes vary only by smooth morphs of one template. All results come from a single pretraining run;
their spread over pretraining seeds is not measured. A Transolver-3~\citep{zhou2026transolver3}
trained separately on each dataset is \competition{all}{min} to \competition{all}{max} times more accurate
than \physicl{} with fixed weights on the PLAID competitions' test splits, and \competition{2D_MSHE}{min} to
\competition{2D_MSHE}{max} times on 2D\_MSHE after its fine-tuning (Table~\ref{tab:competitions}; the competitions'
metric, and its choices selected on those scores), at a training cost per dataset outside the setting of
\physicl{}.

\paragraph{Perspectives.} Three directions follow. First, a decoder that predicts several output fields jointly,
with no bound on their number, would let each field benefit from the others in the context. Second, the prior can be
enriched with other physical laws, in particular in 3D, and with larger shape variations within an episode; the
fine-tuning on 2D\_MSHE, whose holes move between simulations, suggests that a prior closer to the target
geometry and loads helps. Third, pretraining on many more episodes, distributed over several GPU nodes, would show how in-context
learning of fields scales with data and compute, and how much of the gap to supervised models it closes.

\section{Conclusion}

In-context learning of physical fields on unstructured meshes is feasible. \physicl{} combines four properties that,
to our knowledge, no previous method combines: several labeled simulations, each a mesh with its fields, serve as
context and are measurably used; the geometry and the mesh vary within a task; fields are predicted at the query's
nodes in one forward pass, with no weight update, on unseen engineering studies; and it is pretrained on a procedural
prior only. On five PLAID datasets, its
error decreases as its context grows and, from $K = 32$, is below that of the nearest context simulation on every
dataset and context draw. Its peak GPU memory at $K = 64$ is \plaid{AhmedML}{64}{peak}\,GiB on the
surfaces of AhmedML, with a median of 1.1 million nodes, against \plaid{2D_MSHE}{64}{peak}\,GiB on the
\plaid{2D_MSHE}{64}{knodes}{,}000-node meshes of 2D\_MSHE. For a dataset far from the prior, fine-tuning on procedural
tasks that reproduce its geometry and loads, without any of its outputs, divides the stress error of
2D\_Multiscale\_Hyperelasticity by \ft{augmentedproxy}{64}{stressgain}.

\section*{Acknowledgements}
We thank Brian Staber (Safran) for the annulus geometry of the prior library.

\section*{Impact Statement}
This work aims to advance machine learning for engineering simulation. Fast and confident surrogates can replace
simulations in design loops, and an in-context surrogate used outside its prior, or miscalibrated, could mislead an
engineering decision. Its quantiles, whose calibration we report, and its explicit prior are meant to make this
visible; neither guarantees its validity on a new dataset, which must be checked against simulations.

\bibliographystyle{icml2026}
\bibliography{references}

\appendix
\counterwithin{figure}{section}\counterwithin{table}{section}

\section{Design choices}
\label{sec:design}

This appendix gives, for each choice of Section~\ref{sec:model}, its motivation and the
related work (Appendices~\ref{app:design-codes}--\ref{app:design-decoder}), then the ablations behind the
architecture choices (Appendix~\ref{app:design-choices},
Table~\ref{tab:choices}). Correctness properties (padding
invariance, cache reuse, counts beyond the training maximum) are unit-tested; they are not accuracy claims.

\subsection{Random variable codes}
\label{app:design-codes}
TabICL tells the columns of a table apart by their order, through rotary embeddings~\citep{qu2025tabicl}. The
variables of a simulation have no natural order and no fixed meaning across the tasks of a prior, so \physicl{} gives
each input of an episode a Gaussian code $r_j \sim \mathcal{N}(0, I_{256})$, drawn afresh for
every training episode and from a fixed seed at inference, and projected by learned linear maps; the output field
gets one too, which tags its context labels and the decoder query. TabPFN v2 also
identifies features, or groups of features in its grouped configuration, by random vectors drawn with a fixed
seed~\citep{hollmann2025tabpfnv2}, and TabICLv2 tried random
identifiers, in the variant used by LimiX, with unclear benefit~\citep{qu2026tabiclv2}. Since a code means nothing beyond its episode, the model reads
a variable's role from the context rather than from a slot, and accepts more variables than it saw in training,
provided pretraining varies their counts.

\subsection{The row encoder of the scalar variables}
\label{app:design-row}
Each simulation also has scalar variables: its input scalars and, in the context, the mean and the standard
deviation of its output field over the simulation. They are encoded by the pipeline of TabICL~\citep{qu2025tabicl} and TabICLv2~\citep{qu2026tabiclv2}, which TabPFN-3
adopted~\citep{grinsztajn2026tabpfn3} and TabPFN-3.5 keeps~\citep{jager2026tabpfn35}. As in a table, each simulation
of the episode, context or query, is a row and each of its scalar variables a column; the cell of simulation $s$ and
variable $j$ is the token $\mathrm{Linear}(v_{sj}) + W r_j + e_\text{role}$, with $v_{sj}$ the value of $j$ in $s$,
standardized over the context, and $e_\text{role}$ a learned vector that tells an input scalar from the mean or
the standard deviation of the output field; a
query row has no output cells. One induced set attention
block~\citep{lee2018settransformer} per variable then runs across the $K + Q$ simulations; its inducing vectors read
the context cells only, as the distribution embedder of TabPFN-3 does, so queries never inform each other or the
context. Three attention blocks then mix the cells of each simulation, its input scalars and, on the context,
the mean and standard deviation of its output field, and two CLS tokens read them out into the row encoding.

\subsection{Node and patch pooling over variable tokens}
\label{app:design-pooling}
Field variables live at mesh nodes. At encoder nodes, anchors and decoded nodes, the input fields,
and on the context the output-field labels, are fused by a multi-head attention pooling over
linear variable tokens: $t_j = W_x x_j + W_r r_j + \sum_f x_{jf} V_f r_j$, weights
$\alpha_{jh} = \mathrm{softmax}_j(q_h \cdot t_j)$ with one learned query per head $h$, and output $\sum_j \alpha_{jh} t_j$,
the 4 heads splitting
the code width. At patches, the context output-field moments ($3 + d$ values) use
the same pooling, and its output is added to the patch token; the row encoding of each simulation
(Appendix~\ref{app:design-row}), projected and broadcast, reaches its patches, in the context and the queries alike, so that $\theta$
enters \tfcol{} on both sides. Its structure is the variable aggregation of ClimaX~\citep{nguyen2023climax},
Stormer~\citep{nguyen2023stormer} and MORPH~\citep{rautela2025morph}: a cross-attention whose query is learned and
whose keys and values are the variables at one position. It pools every variable of a node into one token by softmax weights, which sum to one
(each value is diluted by about $1/J$ for $J$ pooled variables) and depend on the values, and with identities given by the random codes $r_j$ of
Appendix~\ref{app:design-codes} instead of learned per variable.

\subsection{The output-query decoder}
\label{app:design-decoder}
The output field reads the context through its own query, the decoder principle of Perceiver
IO~\citep{jaegle2021perceiverio}: output queries carry the semantics of the output and attend to a shared memory; we
take this principle, not the rest of that architecture. After node ICL and an MLP, a decoded node $h$ forms the query
$\mathrm{proj}(h) + W_o r_j$, with $r_j$ the code of the output field; it reads a null token and the fused anchor keys
in one block of the patch width (256), and a readout, a ratio-2 MLP and a linear layer, gives 19 quantiles. This last
read lets each decoded node gather from the anchors what its prediction needs; without it, the validation pinball
rises by 0.005 (Table~\ref{tab:choices}). Quantile count:
with 19 levels $i/(n+1)$, $n = 19$, the median is a predicted level and their mean already estimates the predictive mean, so more
levels would refine only the tails (beyond 5\% and 95\%), at a memory cost that grows with the number of levels.

\begin{table}[h]
  \centering
  \caption{Parameters per stage of \physicl{} (width 256, 2D and 3D); the counts do not depend on the number of
    variables.}
  \label{tab:params}
  \small
  \begin{tabular}{lr}
    \toprule
    Stage & Parameters \\
    \midrule
    Patch encoder & 0.72\,M \\
    Context labels in patches & 1.05\,M \\
    Row encoder & 2.94\,M \\
    \tfcol{} (1 block) & 1.10\,M \\
    \tfrow{} (4 blocks) & 4.03\,M \\
    \tficl{} (6 blocks, width 512) & 12.87\,M \\
    Patch ICL (4 blocks) & 2.34\,M \\
    Node ICL (2 blocks) & 1.70\,M \\
    Output read and head & 1.68\,M \\
    \midrule
    \textbf{Total} & \textbf{28.44\,M} \\
    \bottomrule
  \end{tabular}
\end{table}

\subsection{Architecture choices}
\label{app:design-choices}
The model is trained on tasks with one output field and no output scalar, and each field is decoded alone at
inference. It has no scalar head; scalars are a regression on the per-simulation row tokens, read
from a context and queries encoded with no output observed: by default Kumo-Tabular on the scalar inputs $\theta$ and
the scores of a PCA of the context tokens at 99.99\% of their variance. The model has 28.4M
parameters (Table~\ref{tab:params}), the final size. Table~\ref{tab:choices} gives one
justification per architecture choice: a principle, the literature, or an ablation. The ablations use a 7.2M-parameter model (width 128) on the generators and banks of the pretraining configuration, at
15k steps, one seed per variant against a two-seed baseline, which has the values kept except 16 encoder nodes per patch.
They measure the fixed validation, a fixed set of prior tasks scored by the field pinball loss. Changes of
at most 0.001, five times the gap between the two baseline seeds, are ties, on which the default is kept. Each variant was also scored on three PLAID datasets (VKI-LS59, 2D\_MSHE and
Rotor37) as a check. The final size is an arbitration between capacity and batch: it lets a
batch of 8 episodes per GPU fit with contexts of up to $K = 64$ simulations (95 of 127\,GB in the worst case). The depths of the
five attention stages and the field-quantile count come with that sizing and were not ablated. These choices were
frozen before the final pretraining.
\begin{table*}[t]
 \centering\scriptsize
 \caption{Architecture choices of the single-output model. $\Delta$: change in fixed-validation field pinball of the variant
 named against the two-seed baseline of the small model (7.2M parameters) at 15k steps; $\Delta > 0$: the variant is
 worse.}
 \label{tab:choices}
 \begin{tabular}{@{}>{\raggedright\arraybackslash}p{2.4cm}>{\raggedright\arraybackslash}p{1.9cm}>{\raggedright\arraybackslash}p{3.9cm}>{\raggedright\arraybackslash}p{4.3cm}@{}}\toprule
 Choice & Value & Source & Ablation\\\midrule
 Variable identity & random codes & Appendix~\ref{app:design-codes}; cf.\ ClimaX (learned variable embeddings)~\citep{nguyen2023climax} & ---\\
 Outputs & one field per task, decoded alone & avoids cross-field contamination (Section~\ref{sec:model}) & ---\\
 Output read (a query carrying the output's code) & on & Perceiver IO & off: $+0.005$\\
 Variable pooling & value-coded attention & ClimaX aggregation, Appendix~\ref{app:design-pooling} & sum pooling: $+0.013$\\
 Encoder nodes per patch & 8 & inverse-inclusion estimator & 8 against 16: $-0.001$, 1.47$\times$ the steps/s; 32: $+0.002$, slower\\
 Decoded nodes per query & 1{,}024 & loss estimator & 512: $0.000$; 2{,}048: $+0.002$\\
 Anchors per context simulation & 64 & --- & 32: $+0.001$; 128: $+0.0007$\\
 Quadrature nodes & 1{,}024 & --- & 512: $-0.0004$; 2{,}048: $+0.003$\\
 Optimizer, lr & Muon, $3\cdot10^{-4}$ & Muon~\citep{jordan2024muon}, TabICLv2~\citep{qu2026tabiclv2} & AdamW: $+0.039$; lr $1.5\cdot10^{-4}$: $+0.008$; $6\cdot10^{-4}$: $+0.001$\\
 \midrule
 \multicolumn{4}{@{}l}{Not ablated: set with the final size}\\[2pt]
 Choice & Value & \multicolumn{2}{>{\raggedright\arraybackslash}p{\dimexpr8.2cm+2\tabcolsep\relax}@{}}{Source}\\\midrule
 \tfcol{} & 1 block & \multicolumn{2}{>{\raggedright\arraybackslash}p{\dimexpr8.2cm+2\tabcolsep\relax}@{}}{ISAB across simulations; Set Transformer, TabPFN-3, TabICL~\citep{lee2018settransformer,grinsztajn2026tabpfn3,qu2025tabicl}}\\
 \tfrow{} & 4 blocks & \multicolumn{2}{>{\raggedright\arraybackslash}p{\dimexpr8.2cm+2\tabcolsep\relax}@{}}{patches of a simulation}\\
 \tficl{} & 6 blocks & \multicolumn{2}{>{\raggedright\arraybackslash}p{\dimexpr8.2cm+2\tabcolsep\relax}@{}}{row tokens (CLS outputs) across simulations; TabICL~\citep{qu2025tabicl}}\\
 Patch ICL & 4 blocks & \multicolumn{2}{>{\raggedright\arraybackslash}p{\dimexpr8.2cm+2\tabcolsep\relax}@{}}{Perceiver IO~\citep{jaegle2021perceiverio}}\\
 Node ICL & 2 blocks & \multicolumn{2}{>{\raggedright\arraybackslash}p{\dimexpr8.2cm+2\tabcolsep\relax}@{}}{Perceiver IO}\\
 Field quantiles & 19 & \multicolumn{2}{>{\raggedright\arraybackslash}p{\dimexpr8.2cm+2\tabcolsep\relax}@{}}{Appendix~\ref{app:design-decoder}}\\
 \bottomrule\end{tabular}
\end{table*}

\subsection{Scale conventions}
\label{app:scales}
\physicl{} and the baselines share these conventions. Every output is standardized over the context as
$(v - \mu_C)/s_C$, with $\mu_C$ and $s_C$ its mean and standard deviation over the context simulations
(area-weighted for fields). A constant, nonempty context output, equal to $c$, takes
\begin{equation}
 \mu_C = c,\qquad s_C = \begin{cases}|c|,&c \ne 0,\\ 1,&c = 0,\end{cases}
\end{equation}
constancy being tested by the equality of the valid minimum and maximum, before the relative standard-deviation
floors; for fields, only positive-weight quadrature observations enter. A constant context cannot identify its
task's output spread, so this fallback is a convention: it preserves multiplicative unit scalings for $c \ne 0$, not
general affine translations. Coverage is the area-weighted share of nodes with $\hat Q_{0.05} \le u \le \hat Q_{0.95}$,
the pinball loss $\rho_\tau(e) = e(\tau - \boldsymbol{1}_{e < 0})$ with $e = y - \hat Q_\tau$.

\section{The procedural prior}
\label{app:prior}

\subsection{Generators}
Every generator prepares a bank of 1{,}000 templates $\times$ 12 meshes of 20{,}000--50{,}000 nodes (log-spaced),
each mesh with its wall tags, its RIB partition into 1{,}024 patches and its area or volume distribution; fixed
quadratures (1{,}024 points, 512 in volumes) calibrate the laws. Each simulation of an episode draws one of the
template's meshes and morphs it with its design variables (strength 0.15).
\begin{center}\small
\begin{tabular}{@{}>{\centering\arraybackslash}m{1.5cm}>{\raggedright\arraybackslash}m{6.3cm}@{}}
\includegraphics[width=1.45cm,height=1.45cm,keepaspectratio]{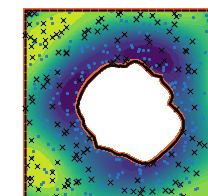} & \texttt{radial} (2D): star-shaped unions of 1--4 polygons or superellipses, interior or exterior (an obstacle
    in a channel); Delaunay meshes refined at the wall; designs move the radius through boundary harmonics 2--6. \\
\includegraphics[width=1.45cm,height=1.45cm,keepaspectratio]{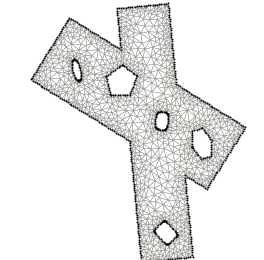} & \texttt{shapes} (2D, non-star): one body in a box (NACA 4-digit airfoil, ellipse or blob) or an L, T, U,
    notched or rectangle-union polygon with up to 8 holes; designs are rigid motions and scalings per body, blended to
    the identity, a diffeomorphism with a closed-form Jacobian. \\
\includegraphics[width=1.45cm,height=1.45cm,keepaspectratio]{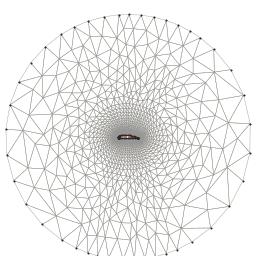} & \texttt{annulus} (2D): one body (NACA 4- or 5-digit or CST airfoil, turbine blade, superellipse, polygon or
    blob) in an annular far field of 4 to 7 body sizes, on O-grid or Delaunay meshes; designs vary 1--3 shape
    parameters of the body's family, with up to 3 small wall bumps or smooth wall modes. \\
\end{tabular}
\end{center}
\begin{center}\small
\begin{tabular}{@{}>{\centering\arraybackslash}m{1.5cm}>{\raggedright\arraybackslash}m{6.3cm}@{}}
\includegraphics[width=1.45cm,height=1.45cm,keepaspectratio]{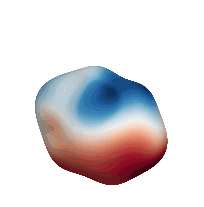} & \texttt{radial\_surface} (3D surfaces): unions of 1--4 convex primitives (ellipsoids, cuboids, prisms);
    designs map $x \mapsto x\,e^{m(x/|x|)}$ with $m$ a mix of spherical harmonics of degree 2--6; normals as inputs. \\
\includegraphics[width=1.45cm,height=1.45cm,keepaspectratio]{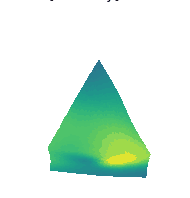} & \texttt{radial\_volume} (3D tetrahedra): the same bodies, interior or in a box, morphed along rays. \\
\includegraphics[width=1.45cm,height=1.45cm,keepaspectratio]{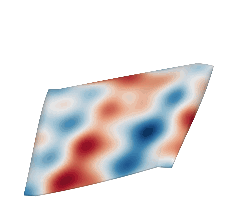} & \texttt{shapes\_surface} (3D surfaces): constructive solids (tori, bored solids, genus 1--3), wings (lofted
    NACA sections with taper, sweep, dihedral and washout), blades twisted by 15--40$^\circ$, and ducts. \\
\includegraphics[width=1.45cm,height=1.45cm,keepaspectratio]{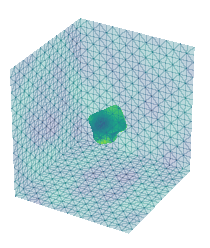} & \texttt{shapes\_volume} (3D tetrahedra): these bodies in a box, or non-star solids with up to 2 holes, meshed
    by isosurface stuffing; no inverted tetrahedron in 14{,}400 validation cases. \\
\end{tabular}
\end{center}

\subsection{Generic laws}
A generic field is a constant plus modes times coefficients, calibrated to unit RMS on the template's quadrature.
The eleven basis families:
\begin{center}\small
\begin{tabular}{@{}>{\centering\arraybackslash}m{1.5cm}>{\raggedright\arraybackslash}m{6.3cm}@{}}
\includegraphics[width=1.45cm]{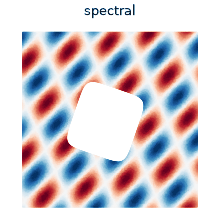} & \texttt{spectral}: cosines $\cos(f\cdot x + \varphi)$, $f = 2\pi\,U(0.4, 3)\,\mathcal{N}(0, I)$, amplitudes $\max(|f|, 1)^{-p}$, $p \sim U(0.35, 1.5)$ \\
\includegraphics[width=1.45cm]{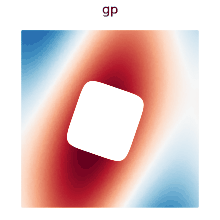} & \texttt{gp}: random Fourier features, length scales $\exp U(\log 0.2, \log 1.5)$, RBF or Student-$t$ kernel \\
\includegraphics[width=1.45cm]{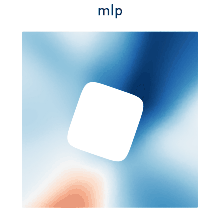} & \texttt{mlp}: a random network $W_2\,\sigma(W_1 x + b_1)$, width 32, $\sigma$ tanh, sine or softplus \\
\includegraphics[width=1.45cm]{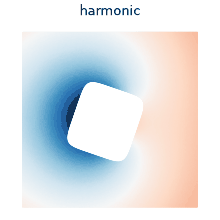} & \texttt{harmonic}: coordinates, monopoles ($\log r$, $1/r$) and dipoles, sources outside the domain or inside the obstacle \\
\includegraphics[width=1.45cm]{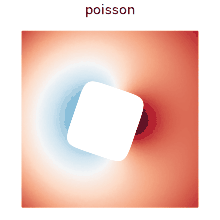} & \texttt{poisson}: harmonic terms plus $|x - x_0|^2/4$ in 2D, $/6$ in 3D (a constant source) \\
\includegraphics[width=1.45cm]{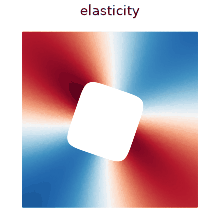} & \texttt{elasticity}: Kelvin point-force solutions, $\nu \sim U(0.1, 0.45)$, 2 or 3 components per source \\
\end{tabular}
\end{center}
\begin{center}\small
\begin{tabular}{@{}>{\centering\arraybackslash}m{1.5cm}>{\raggedright\arraybackslash}m{6.3cm}@{}}
\includegraphics[width=1.45cm]{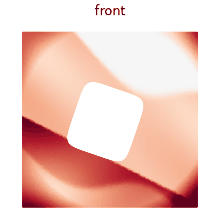} & \texttt{front}: sharp steps $\sigma((s - \psi)/\varepsilon)$ (80\%) and kinks (20\%) on plane, distance or Fourier level sets $\psi$ \\
\includegraphics[width=1.45cm]{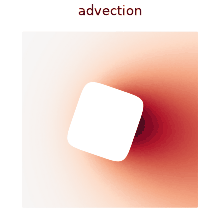} & \texttt{advection}: exact steady advection--diffusion point sources, one flow direction and P\'eclet number per task \\
\includegraphics[width=1.45cm]{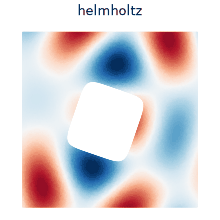} & \texttt{helmholtz}: exact point sources ($\cos(kr + \varphi)/r$, $J_0$, $Y_0$) and plane waves, one wavenumber per task \\
\includegraphics[width=1.45cm]{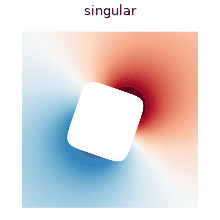} & \texttt{singular}: coordinates and fractional-power singularities $r^\lambda$ just outside the domain \\
\includegraphics[width=1.45cm]{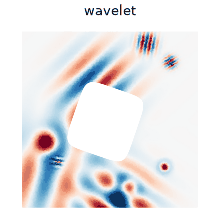} & \texttt{wavelet}: Ricker and Morlet wavelets, scales over 4 octaves \\
\end{tabular}
\end{center}
Modes are ambient,
wall-attached ($e^{-\mathrm{sdf}/\ell}$ times an angular harmonic) or shape-attached ($\cos(p\pi\,\mathrm{sdf})$ times
an angular harmonic); laws are dense or sparse in the drivers (weights 2 and 1). The wall condition multiplies a field
by $1 - e^{-\mathrm{sdf}/\delta}$, $\delta$ log-uniform in $[0.02, 0.2]$, in half of the tasks.

\subsection{Physical models}
Each model solves a linear PDE of the case's domain by boundary elements, exactly up to the wall discretization, then
adds classical engineering approximations. Their fields are physically shaped and consistent with each other, not
CFD- or FEM-accurate. Figures~\ref{fig:gallery-physical}--\ref{fig:gallery-strip} show their fields.
\begin{itemize}[leftmargin=*, itemsep=1pt, topsep=2pt]
  \item \texttt{flow}: potential flow by Hess--Smith panels with a Kutta condition, Karman--Tsien compressibility,
    shock jumps and a Thwaites/Head boundary layer; half of the tasks show no-slip layers (Pohlhausen laminar,
    Reichardt and Coles turbulent profiles, reversed past separation, a free shear layer at stall) with Mach 0 on the
    wall; every task has an eddy viscosity (Cebeci--Smith in the layers, the trailing edge's value in the wakes).
    Outputs, one per task: Mach, pressure coefficient, density, temperature, velocity, eddy viscosity.
  \item \texttt{plate}: plane elasticity (Kolosov--Muskhelishvili): von Mises stress, displacement, stress tensor,
    strain-energy density and the asymmetric first Piola stress $P = (I + \lambda\nabla u)\,\sigma$, with a per-task
    strain $\lambda$ log-uniform in $[0.01, 0.3]$.
  \item \texttt{darcy}: anisotropic Laplace with 1--4 wells (interior domains): head and Darcy velocity.
  \item \texttt{torsion}: $\nabla^2\phi = -2$ with Bredt's condition (interior domains): stress function and shear stress.
  \item \texttt{heat}: Laplace or screened Laplace with Robin, Dirichlet or insulated walls: temperature and heat flux.
  \item \texttt{stokes}: Stokeslets inside, Brinkman flow outside: velocity, pressure and vorticity.
  \item \texttt{strip} (3D wings and blades): the \texttt{flow} fields of 8 spanwise sections, coupled by Prandtl's
    lifting line and, on blades, a blade-element inflow set by a tip-speed ratio; the mid-span pressure matches the 2D
    model at the induced incidence within 1.6--1.8\% of its range.
\end{itemize}
Validity. Each model bounds a few values (e.g.\ Mach in $[0, 1.6]$); a case whose solve raises, or whose values
are not finite or leave the bounds, is redrawn deterministically on its own random streams (at most 8 times), then clipped.

\subsection{Label quality of the procedural prior}
The laws are calibrated on scrambled-Sobol quadratures of each template, not on the mesh nodes, so that the labels
do not depend on the mesh resolution. On 16 tasks per family (GP, spectral and harmonic), Sobol points reduce the
median integral error from \journalIntegralUniform{} to \journalIntegralSobol{} label standard deviations against
uniform points at the same budget, with no measurable cost.

\subsection{What the spatial budgets bound}
\label{app:budgets}
Let $n_p$ be the number of prepared patches, $n_e$ the encoder nodes per patch, $B_\partial$ the boundary budget, $d$
the dimension, $n_v$ the quadrature budget and $n_d$ the decoded-node budget. The node sets sampled from a
simulation, its encoder nodes $\mathcal{N}_\mathrm{enc}$, its quadrature nodes $\mathcal{N}_\mathrm{quad}$ and its decoded nodes
$\mathcal{N}_\mathrm{dec}$, then satisfy
\begin{equation}
 \begin{gathered}
 |\mathcal{N}_\mathrm{enc}|\leq n_pn_e+B_\partial+2d,\\
 |\mathcal{N}_\mathrm{enc}\cup \mathcal{N}_\mathrm{quad}\cup \mathcal{N}_\mathrm{dec}|\leq n_pn_e+B_\partial+2d+n_v+n_d,
 \end{gathered}
\end{equation}
where $2d$ counts the extreme wall nodes, a minimum and a maximum along each axis, that fix the context frame, so
the $m$ points of Section~\ref{sec:prior} number at most $n_pn_e+B_\partial+2d+n_v+n_d$. The boundary nodes belong
to $\mathcal{N}_\mathrm{enc}$, and the anchors are drawn among $\mathcal{N}_\mathrm{quad}$. Preparation, cold opening, validity checks,
rejection fallbacks and the search for the extreme nodes are counted separately. Decoding every query node costs
work proportional to the number of requested nodes.

\subsection{The proxy and the augmented proxy of 2D\_MSHE}
\label{app:finetune-prior}
Used only to fine-tune, never in pretraining. The pretraining prior never reproduces 2D\_MSHE's geometry: its 5--8
equal holes (radius 0.05 of the unit cell, gaps and clearances at least 0.05) appear and move within the dataset,
which no morph produces. The \texttt{perforated} generator draws such perforated cells, one per simulation (12{,}000 cells of
one mesh each, about 5{,}400--6{,}100 nodes, RIB 1{,}024 patches). Its \texttt{rve} model is linear plane elasticity on the cell, with traction-free holes and a displacement of the
outer wall set by the loading $C$; $C_{11}, C_{22}$ are uniform on $[0.8, 1.2]$ and $C_{12}$ on $[-0.2, 0.2]$,
independent, as measured on the train split. Its outputs, one per task, are the displacement, the first Piola stress
and the energy density; its input fields are the signed distance and the direction to the nearest wall. The model is
linear where the dataset is hyperelastic: against the dataset (21 samples, an affine map per field), its fields
correlate at 0.98 (displacement), 0.92 (energy density) and 0.63--0.73 (Piola stress), and it misses the bands of
stress that join nearby holes. No parameter is fitted to the dataset's outputs. The augmented proxy keeps the
cells and inputs but draws \texttt{rve} in 3/8 of the tasks, the other physical models of Appendix~\ref{app:prior}
(flow, plate, Darcy, torsion, heat, Stokes) in another 3/8 and generic laws with up to four input fields in the last
quarter.

\section{Gallery of the procedural prior}
\label{app:gallery}

Figures~\ref{fig:gallery-geometry}--\ref{fig:gallery-rve} show the procedural prior of
Section~\ref{sec:prior} and the proxy of 2D\_MSHE (Appendix~\ref{app:finetune-prior}), from the documentation of
our prior library. Each figure was drawn on a documentation bank, coarser than the training banks.

\newcommand{\galleryfigure}[3]{%
  \begin{figure*}[h]
    \centering
    #1
    \caption{#2}
    \label{#3}
  \end{figure*}}

\galleryfigure{\includegraphics[width=\linewidth]{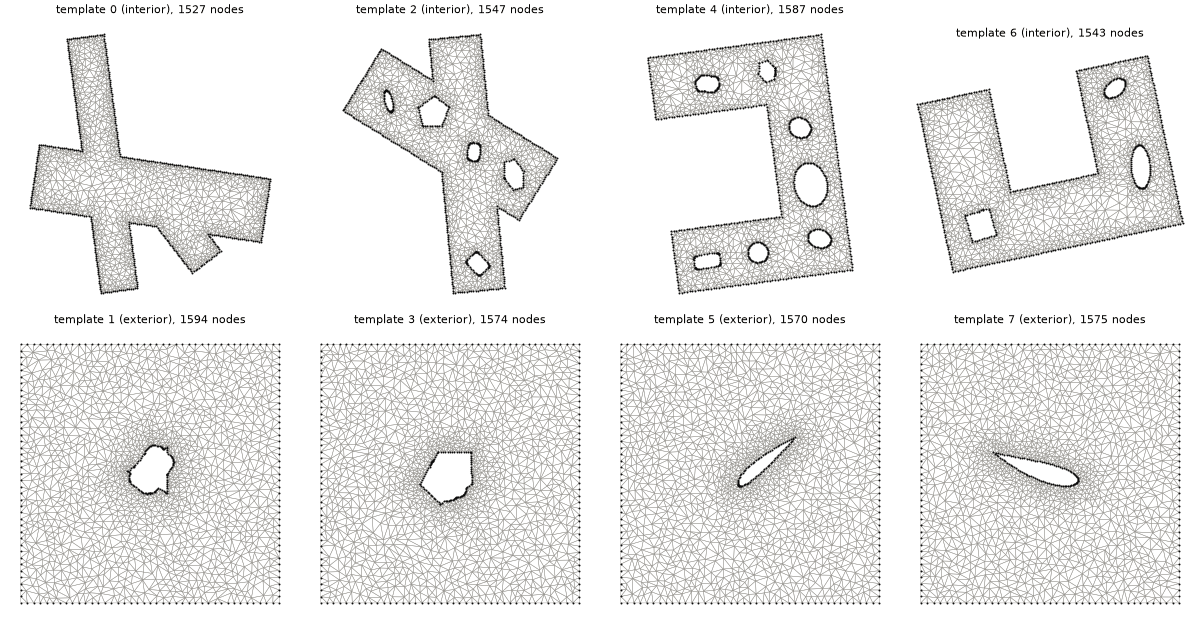}}
  {\textbf{2D geometries} of \texttt{shapes}: polygons with holes and bodies in a box, with their meshes.}{fig:gallery-geometry}
\galleryfigure{\includegraphics[width=0.85\linewidth]{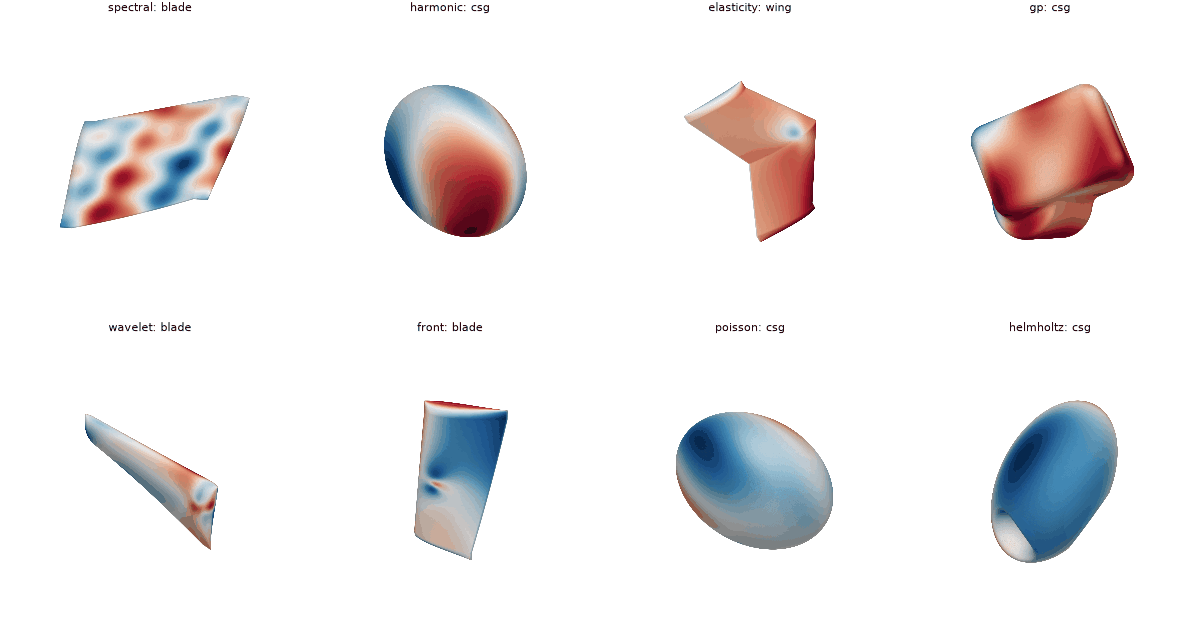}}
  {\textbf{3D surface geometries} of \texttt{shapes\_surface}: constructive solids, wings, blades and ducts.}{fig:gallery-surfaces}
\galleryfigure{\includegraphics[width=0.85\linewidth]{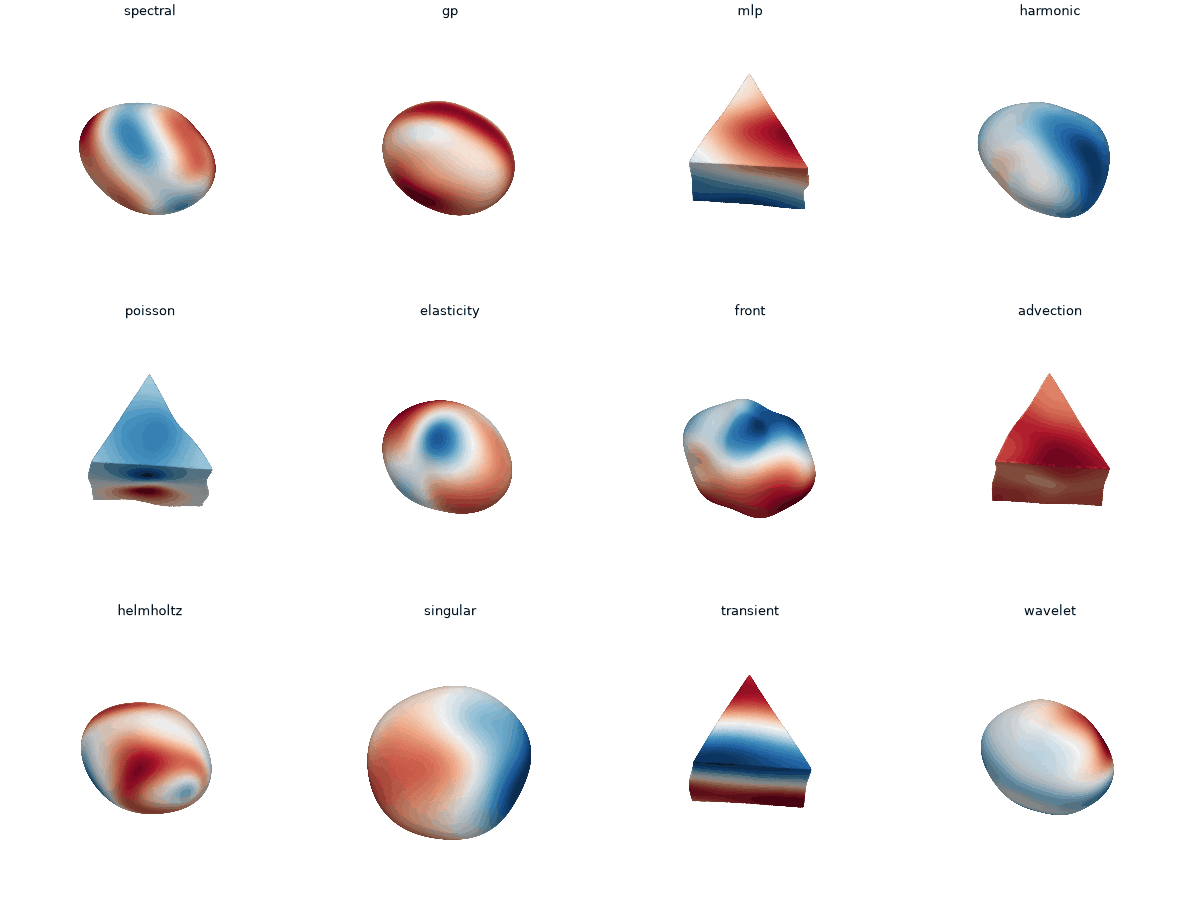}}
  {\textbf{Star surfaces} of \texttt{radial\_surface}, morphed by spherical harmonics, one generic law each.}{fig:gallery-radial-surface}
\galleryfigure{\includegraphics[width=0.85\linewidth]{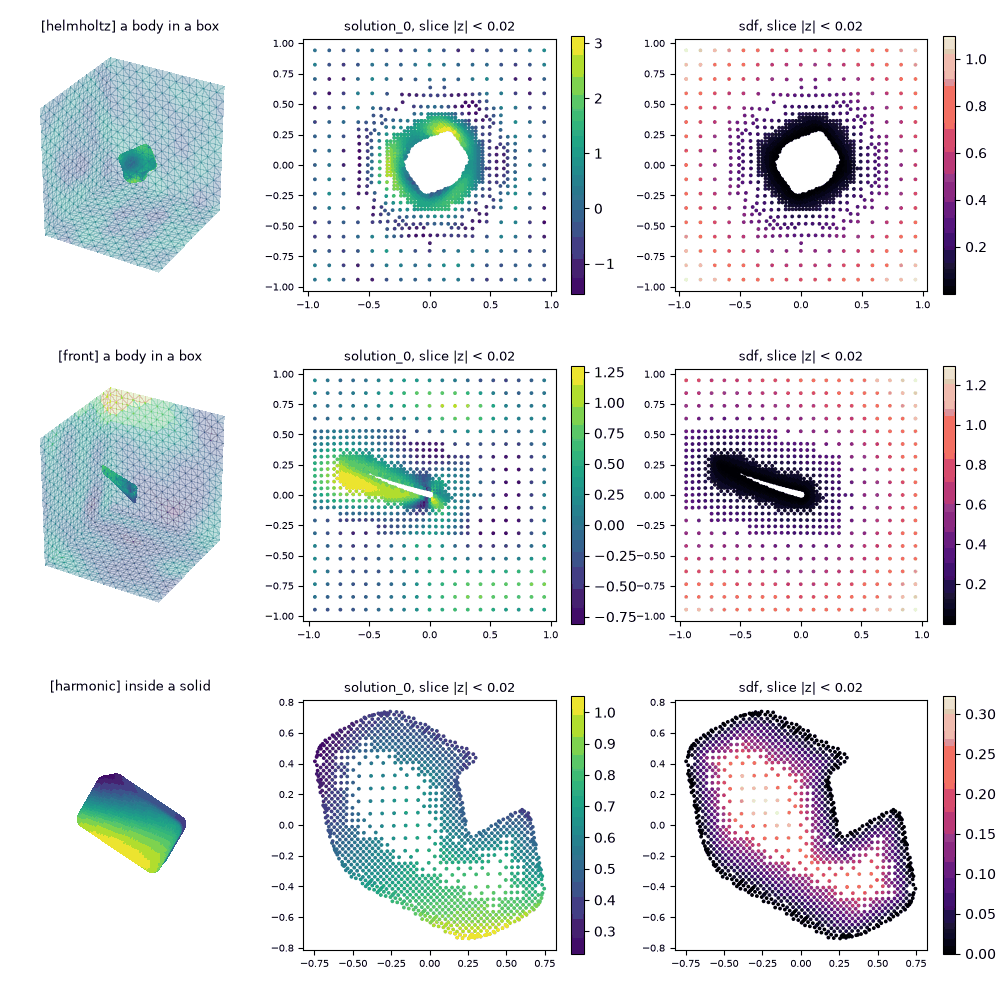}}
  {\textbf{Volume tasks} of \texttt{shapes\_volume}: the wall, a mid-plane slice of the field and of the signed
    distance.}{fig:gallery-volumes}
\galleryfigure{\includegraphics[width=0.24\linewidth]{figures/docs/family_harmonic.png}\hfill
  \includegraphics[width=0.24\linewidth]{figures/docs/family_elasticity.png}\hfill
  \includegraphics[width=0.24\linewidth]{figures/docs/family_front.png}\hfill
  \includegraphics[width=0.24\linewidth]{figures/docs/family_singular.png}}
  {\textbf{Generic laws}: four of the eleven basis families around one obstacle (harmonic, elasticity, fronts,
    singular).}{fig:gallery-generic}
\galleryfigure{\includegraphics[width=0.8\linewidth]{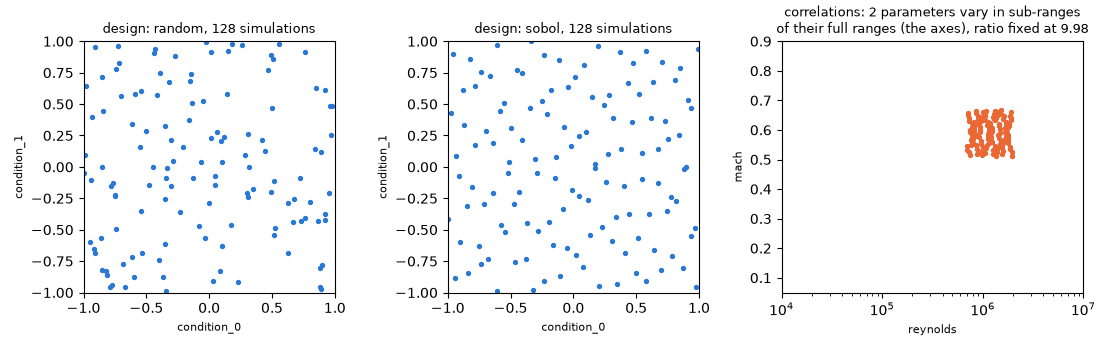}}
  {\textbf{Design variables} drawn at random or along a scrambled Sobol sequence, and conditions varying over
    sub-ranges.}{fig:gallery-design}
\galleryfigure{\includegraphics[width=\linewidth]{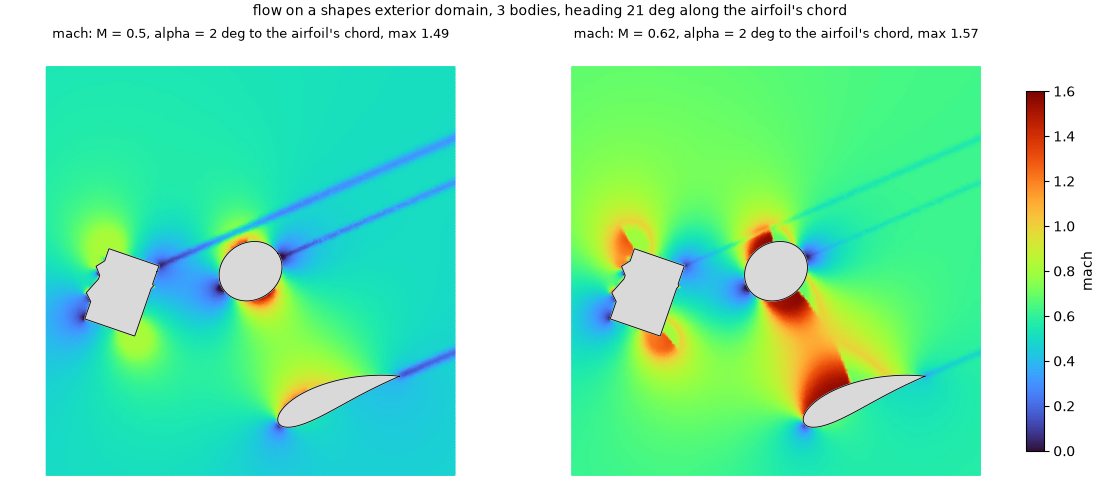}}
  {\textbf{Physical model \texttt{flow}} around bodies in a channel (Mach number).}{fig:gallery-physical}
\galleryfigure{\includegraphics[width=\linewidth]{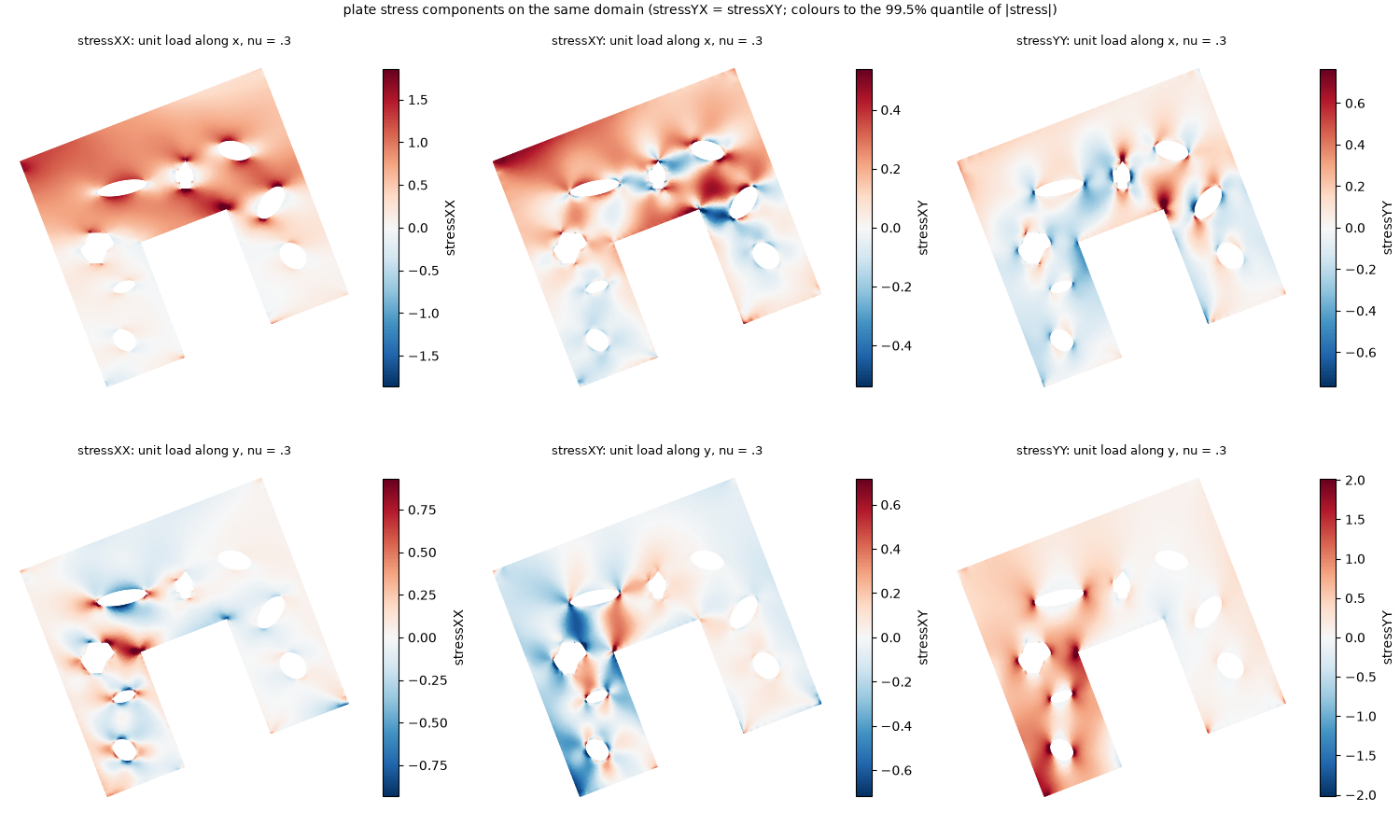}}
  {\textbf{Physical model \texttt{plate}}: stresses on a domain with holes.}{fig:gallery-plate}
\galleryfigure{\includegraphics[width=\linewidth]{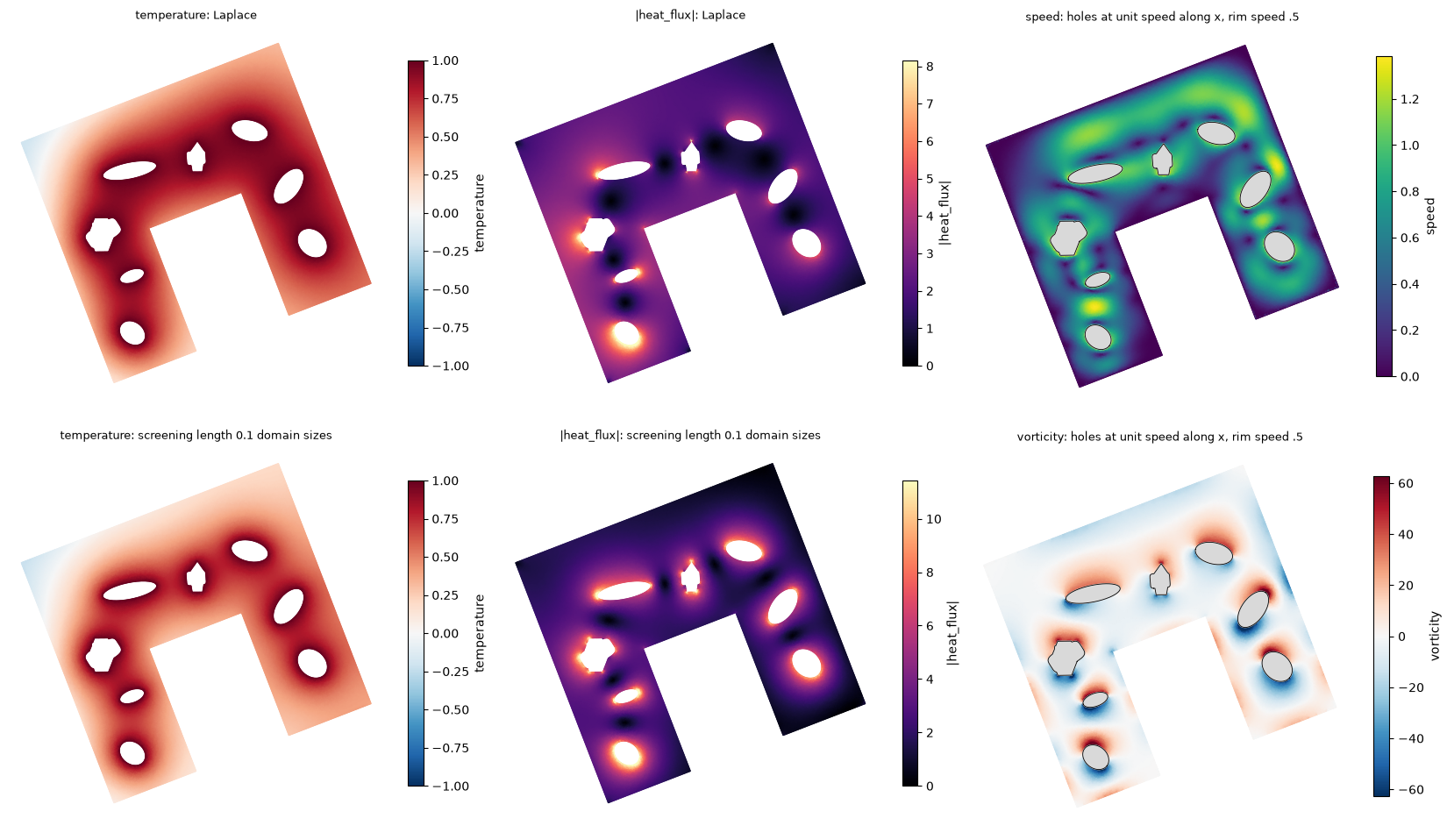}}
  {\textbf{Physical models \texttt{heat}} (temperature and heat flux, without and with a screening length, left and
    middle) and \textbf{\texttt{stokes}} (speed and vorticity, right).}{fig:gallery-heat-stokes}
\galleryfigure{\includegraphics[width=0.75\linewidth]{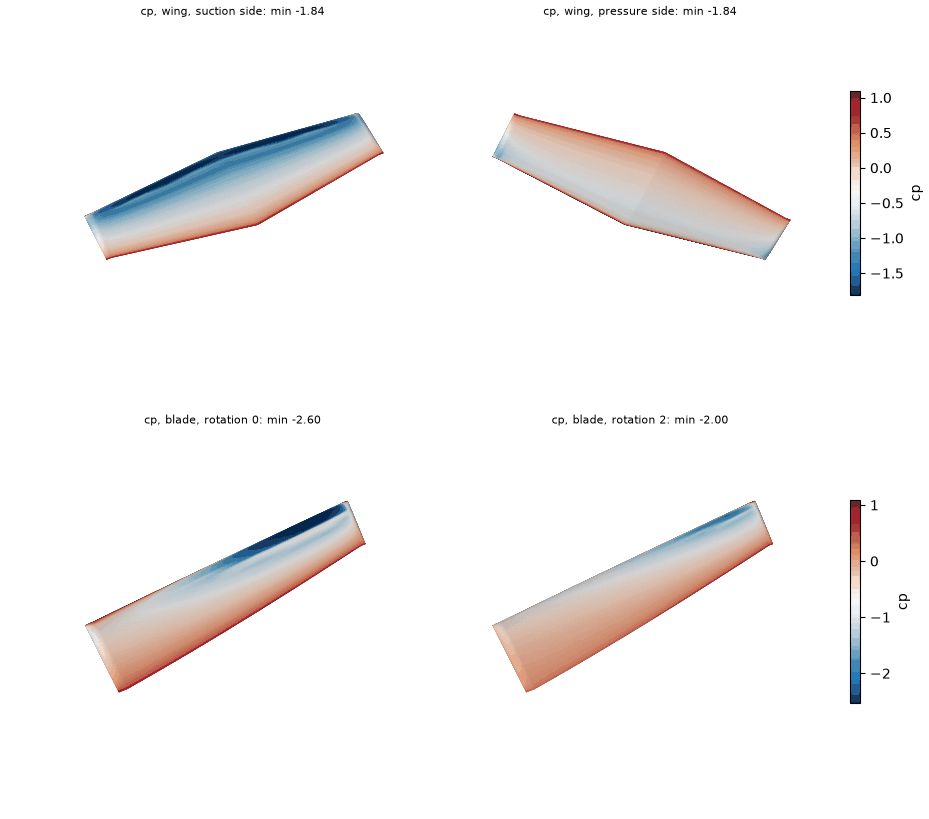}}
  {\textbf{Physical model \texttt{strip}} on a 3D wing and blade (pressure coefficient).}{fig:gallery-strip}
\galleryfigure{\includegraphics[width=\linewidth]{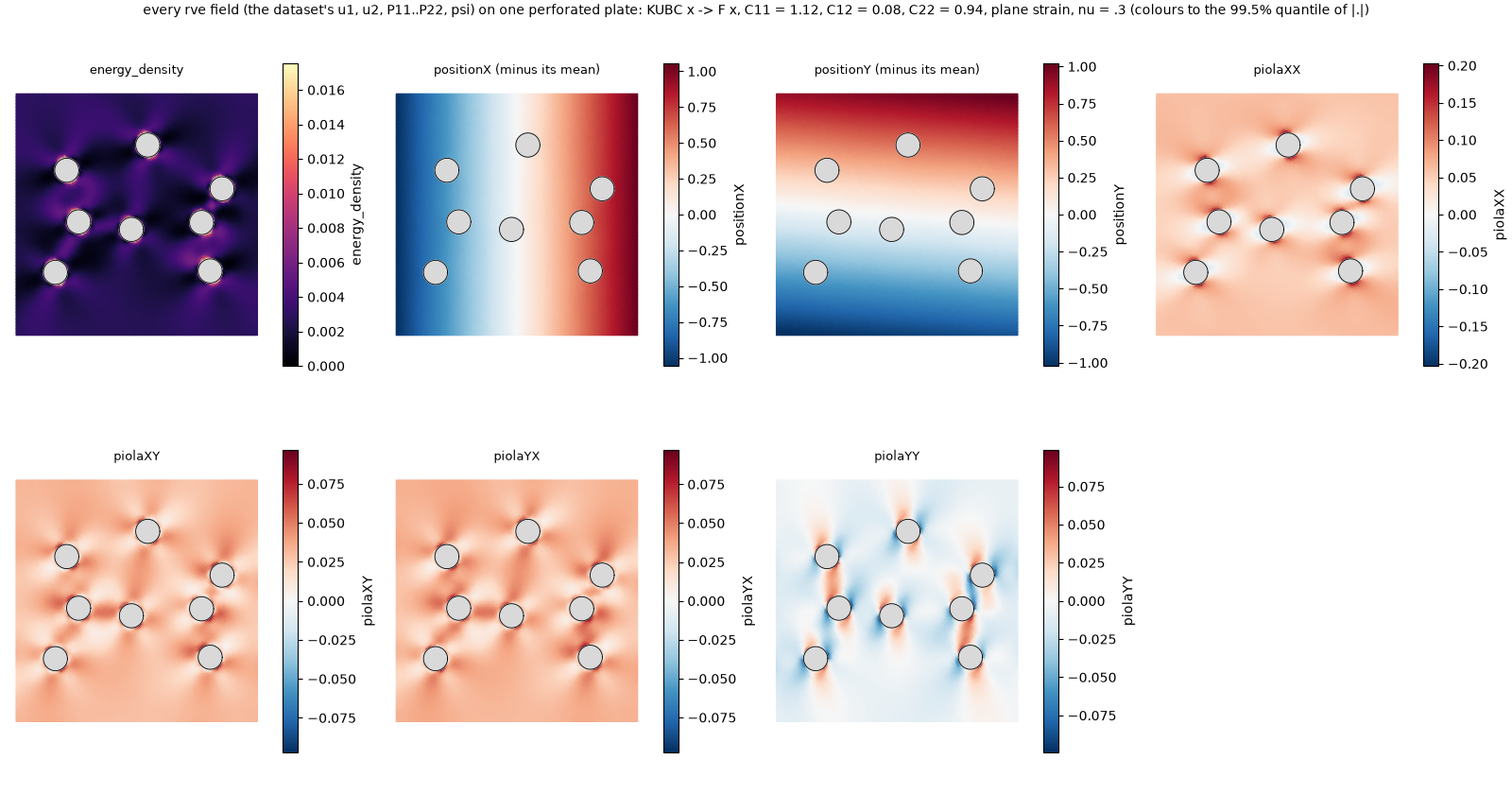}}
  {\textbf{The proxy of 2D\_MSHE} (fine-tuning only): the \texttt{rve} model on one perforated cell,
    displacement, Piola stress and energy density. Each simulation draws its own cell and load $C$.}{fig:gallery-rve}

\section{Baselines}
\label{app:comparisons}

The baselines are fitted per dataset on the same $K$ context simulations as \physicl{}, with no pretraining on the
prior. Write $\theta_s$ for the scalar inputs of simulation $s$, $X_s$ for its node coordinates and $u_{s,c}$ for its
output field $c$. Every fitted quantity (PCAs, regressors, scales) uses the $K$ context simulations only; query labels enter
only the scores and the floors, which are explicit oracles.

\subsection{Shape coordinates and output PCA}
Each baseline regresses the PCA scores of the context output fields on $z_s = [\theta_s, a^\mathrm{shape}_s]$, the
scalar inputs and shape coordinates. For a common-support matrix $F_\mathcal{C} \in \R^{K \times n}$, with $n$ the nodes or probes of the common support,
whose row $f_s \in \R^n$ is the field of simulation $s$ on that support,
\begin{equation}
 \begin{gathered}
 \bar f = K^{-1}\textstyle\sum_{s \in \mathcal{C}} f_s,\qquad F_\mathcal{C} - \boldsymbol{1}\bar f^\top = U\Sigma V^\top,\\
 a_s = (f_s - \bar f)^\top V_{n_r},\qquad \hat f_q = \bar f + V_{n_r} \hat a_q,
 \end{gathered}
 \label{eq:baseline-pca}
\end{equation}
with $V_{n_r}$ the first $n_r$ right singular vectors, $\hat a_q$ the scores that a regressor predicts for the query
(next subsection), and $n_r$ the fewest modes reaching a fraction of the context variance (0.99, 0.999 or 0.9999, swept), at most $K - 1$;
each output field has its own PCA. Shared topology (\texttt{pca\_*}): when every mesh of a dataset has the same
connectivity and node count (Rotor37, VKI-LS59), the fields stay on their node indices and the shape coordinates are
the scores of a PCA of $\mathrm{vec}(X_s)$. Varying meshes (\texttt{probe\_*}): every simulation is resampled
on a rectilinear grid $G$ of $2^{14}$ probes ($2^{18}$ on AhmedML), each axis at quantiles of the context meshes'
node coordinates, so the grid reads no query mesh. A field is transferred to $G$ by barycentric interpolation in the
cell containing a probe, else from the nearest node; the shape is the signed distance to the boundary nodes, negative
inside a cell, and its PCA gives the shape coordinates. A predicted grid is interpolated back to the query nodes
(multilinear, $R_q$); for output field $c$, with its own PCA, $\hat u_{q,c} = R_q(\bar f_c + V_{n_{r,c}}\hat a_{q,c})$.

\subsection{Regressors and controls}
Gaussian process (\texttt{\_gp}): on standardized features, an anisotropic Mat\'ern-$5/2$ kernel plus a
learned white noise, one multi-output GP per field sharing its hyperparameters across the field's normalized scores;
its posterior mean, mapped back to target units, is the prediction. The hyperparameters maximize the log marginal
likelihood by L-BFGS-B from the default start and 2, 8 or 32 seeded log-uniform draws of the signal variance and
length scales in $[0.1, 10]$ and of the noise in $[10^{-6}, 10^{-2}]$ (budgets \texttt{small}, \texttt{medium},
\texttt{large}), the best likelihood kept: scikit-learn's own restarts, drawn across the whole bounds, mostly land on
the plateau of very large length scales and stop within a few iterations. Kumo-Tabular (\texttt{\_kumo},~\citealp{qu2026kumotabular}): a pretrained tabular model, without gradient updates, of size small, medium or
large and with 1, 4 or 8 estimators, reading the raw features; each PCA score is a separate target, and the point
prediction is the mean of the library's predicted quantiles. Nearest context (\texttt{\_nn}): the outputs of
the context simulation closest to the query in the standardized features, a lookup that any learner must beat.
Context mean: the area-weighted mean field of the context simulations. PCA floor (\texttt{pca\_floor},
\texttt{probe\_floor}): the true query field projected onto the context output modes,
$R_q[\bar f + V_{n_r} V_{n_r}^\top(T_q u_q - \bar f)]$, i.e.\ a perfect regressor of the scores; it uses the query truth and
shows what the context PCA can represent, not a strict lower bound (transfer and truncation errors can partly cancel).
Its error has two sources: the transfer to and from the probe grid ($T_q$, $R_q$; \texttt{probe\_floor} only), and,
above all, the truncation to the span of the $K$ context fields, at most $K - 1$ modes. The latter is large at small
$K$ by construction and falls as $K$ grows: a field whose structures move with the geometry, such as stress peaks at
holes that move, lies outside the span of a few context fields.

\subsection{Settings, selection and speed}
Every combination of the PCA variance, the Kumo size and estimator count and the GP training budget runs at every $K$ of the
training range ($K \le 64$), and each regressor keeps its best setting on each dataset, on the queries themselves. Beyond $K = 64$, each
regressor runs the setting of least error over all outputs at $K = 64$; its variance fraction still sets a number of
modes that grows with $K$. \physicl{}'s scalar regression has one setting, fixed in advance. The probe grid was
checked against finer ones by its floor ($K = 64$, 20 queries): on 2D\_MSHE and 2D\_profile, $2^{16}$ and $2^{18}$
probes cut the round trip $R_q T_q u_q$ by 2.6 to 30 times but leave the floor unchanged, so the context PCA, not the
grid, limits these baselines; on the AhmedML surfaces the floor drops by 30 to 40\% from $2^{14}$ to $2^{18}$
probes, so the AhmedML baselines
use $2^{18}$ probes. The baselines are implemented as fast as we found without changing their predictions:
Kumo's model is loaded once per process, outside the times; each target's context is read once and serves every
query, its estimators run in one forward pass, and its cache is moved back to the GPU once after the fit; the GP fits
(every target and start) run in parallel processes, capped at the job's CPU quota, and a GP query is a kernel row
times precomputed weights; the probe projections of the context meshes run in parallel. As for \physicl{}, the
\texttt{fit} time is a task's work from its context, and the \texttt{query} time that of one more query.

\subsection{Limits of the baselines}
The probe count is not swept on top of their other settings on AhmedML. Beyond $K = 64$ their settings are those of
$K = 64$, and their floors are computed at those settings' variances only. The nearest-context distance is dominated by the shape coordinates when they outnumber
the scalar inputs. Kumo runs in float32, \physicl{} in bfloat16.

The scale conventions that the baselines share with \physicl{} are in Appendix~\ref{app:scales}.

\section{Details of the experiments}
\label{app:exp-details}

\paragraph{Checkpoint.} The pretrained checkpoint is the last step, 250{,}000, of the pretraining run, which also
has its best fixed validation score (0.122). It is selected on that validation, on tasks of the
prior.

\paragraph{Pretraining.} The seven generators of Section~\ref{sec:prior}
on their prepared banks; $K \in [8, 64]$ context simulations and 4 queries per episode, 8 episodes per GPU on 4 H200
GPUs (32 per step); Muon with learning rate $3\cdot10^{-4}$, weight decay 0.01, gradient clipping at 1, 2{,}000 warmup
steps and a linear decay over the last fifth of 250{,}000 steps; bfloat16 and regional compilation; budgets of 8
encoder nodes per patch, 2{,}048 boundary nodes, 1{,}024 quadrature nodes, 64 anchors and 1{,}024 decoded nodes per
simulation; one random rotation per episode; 72 data workers, 18 per GPU, on 80 CPU cores.

\paragraph{Timings.} Fit counts everything that depends on the context only: for \physicl{}, the patch
indices of the $K$ context meshes, their node selections and the context encoding; for a baseline, its PCAs, probe
projections and regression. Query counts the cost of one more query: for \physicl{}, the partition of the new
mesh, its node selection and its decoding; for a baseline, its shape projection, prediction and reconstruction.
Every method is timed on the same node, one H200 GPU and its CPU cores. Reading the dataset and computing the
geometric inputs are not timed. With $M$ partitions, \physicl{}'s times count
its $M$ members.

\paragraph{Against a model trained per dataset.} The PLAID competitions score each dataset's test split
(Table~\ref{tab:competitions}; per output field, the root mean over the test samples of the squared error over the
nodes divided by their count and the truth's squared maximum). \physicl{} reads the whole train split as its
context, $M = 8$, with no gradient step on the dataset; on 2D\_MSHE, the augmented-proxy fine-tuning of
Section~\ref{sec:exp-finetune}, which saw none of the dataset's outputs. Transolver-3, trained on each train split (below), is
more accurate on every field: by \competition{all}{min} to \competition{all}{max} times in direct
inference, and by \competition{2D_MSHE}{min} to \competition{2D_MSHE}{max} on 2D\_MSHE, where the gap is smallest on
the stresses and the energy.

Transolver-3~\citep{zhou2026transolver3} is a supervised reference, the scaled-up version of
Transolver~\citep{wu2024transolver}, trained separately on each dataset's train split: four transformer blocks of width 128 with
eight attention heads and 64 physics-attention slices, a two-layer MLP decoder of the output fields, and an encoder of
the scalar inputs. Its mesh inputs carry dataset-specific nodal geometric and topological features, among them a
signed distance on 2D\_MSHE. It minimizes the relative squared error in physical units with AdamW for at most 2{,}000
epochs on one A30 GPU: a 50-epoch warmup from $10^{-6}$ to $5 \times 10^{-4}$, the learning rate reduced on plateau
from epoch 500, and early stopping on an exponential moving average of the loss on 20\% of the train samples, held
out for validation (the same train split \physicl{} reads as its context). This configuration is the outcome of ten cumulative development stages on these five
datasets, varying the loss (relative squared error against mean squared error), the model size, the output heads
(FiLM-conditioned point heads against MLPs), stochastic depth, the geometric preprocessing and the optimization
schedule, each stage kept or reverted on the competitions' test scores: like the baselines' settings
(Section~\ref{sec:exp-baselines}), its architecture and training choices were selected on the evaluated samples
themselves, the 20\% of the train samples serving only the early stopping.

\begin{table}[tb]
  \centering\footnotesize
  \caption{\textbf{PLAID competitions}, test-split scores (the competitions' metric; lower is better) of Transolver-3,
    trained per dataset, and of \physicl{} with no gradient step on the dataset ($M = 8$, the train split as context):
    the pretrained checkpoint, its weights fixed, on Tensile2d, 2D\_profile, VKI-LS59 and Rotor37; on 2D\_MSHE,
    the checkpoint fine-tuned on the augmented proxy of Section~\ref{sec:exp-finetune}, which saw none of
    the dataset's outputs. Ratio: \physicl{} over Transolver-3.}
  \label{tab:competitions}
  \IfFileExists{results/competitions_table.tex}{\begin{tabular}{llccc}
\toprule
Dataset & Field & Transolver-3 & \physicl{} & ratio \\
\midrule
2D\_MSHE & u1 & 0.0092 & 0.05 & 5.5 \\
 & u2 & 0.0097 & 0.052 & 5.4 \\
 & P11 & 0.027 & 0.057 & 2.1 \\
 & P12 & 0.041 & 0.11 & 2.6 \\
 & P22 & 0.028 & 0.06 & 2.1 \\
 & P21 & 0.041 & 0.11 & 2.5 \\
 & psi & 0.029 & 0.047 & 1.6 \\
\midrule
2D\_profile & Mach & 0.021 & 0.15 & 7.4 \\
 & Pressure & 0.015 & 0.098 & 6.6 \\
 & Velocity-x & 0.023 & 0.23 & 9.9 \\
 & Velocity-y & 0.02 & 0.097 & 4.9 \\
\midrule
Rotor37 & Density & 0.0095 & 0.052 & 5.4 \\
 & Pressure & 0.0094 & 0.053 & 5.6 \\
 & Temperature & 0.0024 & 0.0068 & 2.8 \\
\midrule
Tensile2d & U1 & 0.00084 & 0.0065 & 7.8 \\
 & U2 & 0.00091 & 0.0063 & 6.9 \\
 & sig11 & 0.0017 & 0.01 & 6.1 \\
 & sig22 & 0.0011 & 0.0048 & 4.4 \\
 & sig12 & 0.0013 & 0.0073 & 5.5 \\
\midrule
VKI-LS59 & nut & 0.042 & 0.14 & 3.3 \\
 & mach & 0.018 & 0.055 & 3.1 \\
\bottomrule
\end{tabular}
}{\todo{run paper/scripts/competitions.py}}
\end{table}

\begin{table}[tb]
  \centering\footnotesize
  \caption{\textbf{Tasks of the prior on unseen geometries} (Section~\ref{sec:exp-heldout}): per group (four tasks),
    the median ratio of \physicl{}'s median NMSE ($M = 1$) to the best Gaussian-process or Kumo setting's, and the
    tasks where \physicl{} is on the accuracy--query-time front.}
  \label{tab:heldout}
  \IfFileExists{results/heldout_table.tex}{\begin{tabular}{lcccc}
\toprule
 & \multicolumn{2}{c}{$K = 8$} & \multicolumn{2}{c}{$K = 64$} \\
Group & ratio & front & ratio & front \\
\midrule
radial, plate & 0.99 & 3/4 & 0.93 & 2/4 \\
radial, heat & 0.25 & 4/4 & 0.92 & 3/4 \\
radial, flow/Darcy & 0.44 & 3/4 & 1.2 & 3/4 \\
shapes, plate & 0.46 & 4/4 & 1.2 & 3/4 \\
shapes, heat & 2.2 & 1/4 & 3.1 & 2/4 \\
shapes, flow/Darcy & 0.74 & 4/4 & 1.3 & 4/4 \\
3D surface, strip & 0.28 & 4/4 & 0.21 & 4/4 \\
\midrule
all tasks & 0.5 & 23/28 & 0.98 & 21/28 \\
\bottomrule
\end{tabular}
}{\todo{run paper/scripts/heldout.py}}
\end{table}

\begin{figure*}[tb]
  \centering
  \IfFileExists{figures/fields/heldout_strip.jpg}{\includegraphics[width=0.49\linewidth]{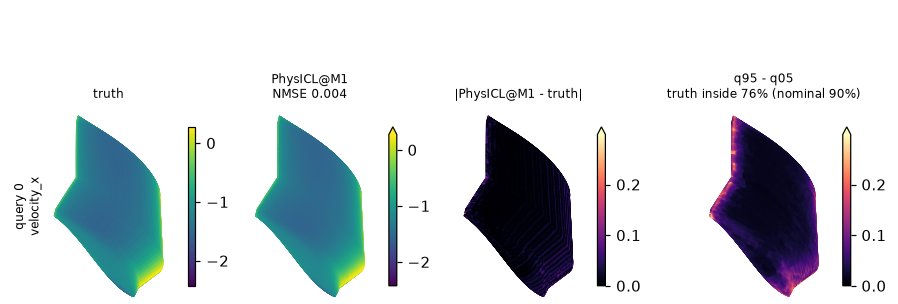}}{\todo{strip}}\\[2pt]
  \IfFileExists{figures/fields/heldout_shapes_plate.jpg}{\includegraphics[width=0.65\linewidth]{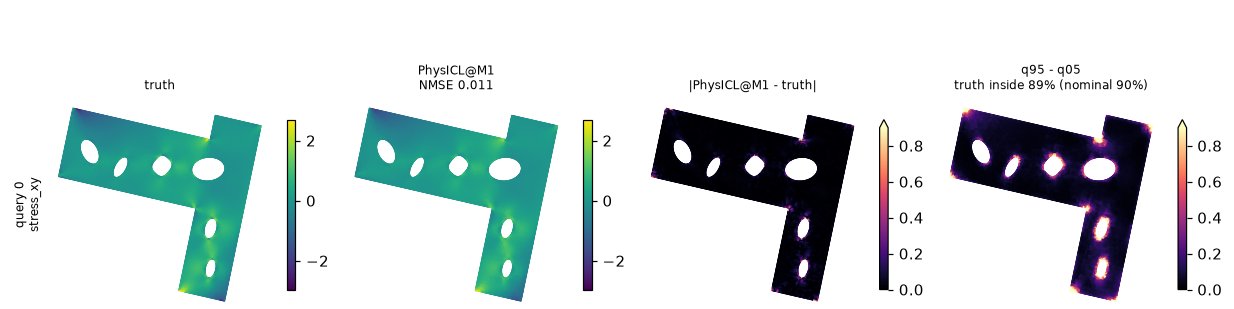}}{\todo{plate}}\\[2pt]
  \IfFileExists{figures/fields/heldout_radial_flow.jpg}{\includegraphics[width=0.6\linewidth]{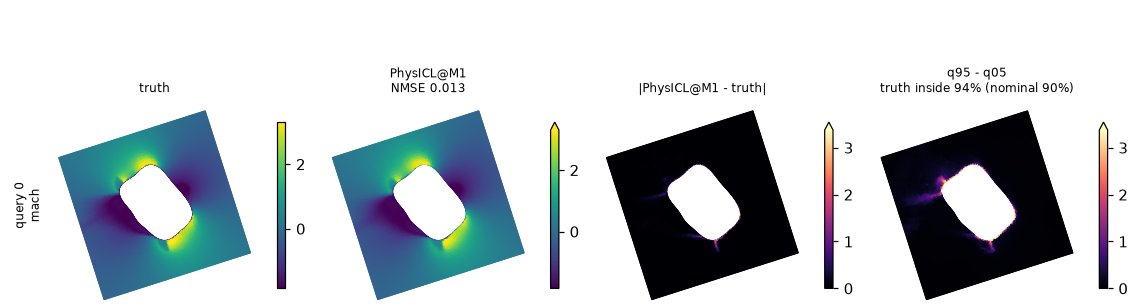}}{\todo{flow}}
  \caption{\textbf{Tasks of the prior on unseen geometries}, $K = 64$, the first query: truth, \physicl{} ($M = 1$),
    its absolute error and its 5--95\% band width. Top to bottom: strip theory on a 3D surface (a velocity
    component), a plate on a shapes domain (a shear stress), flow on a radial domain (Mach number). These three tasks
    are illustrations chosen after the results, each the task of its group where \physicl{} leads the best baseline
    most at $K = 64$; the counts of Section~\ref{sec:exp-heldout} are over the 28 tasks fixed beforehand.}
  \label{fig:heldout-fields}
\end{figure*}

\paragraph{Datasets.} Contexts and queries come from the PLAID train split. The geometric inputs are the dataset's,
else computed from the meshes: the distance to the walls on volume meshes (to the airfoil on 2D\_profile), outward
normals on surfaces.
\begin{table}[h]
 \centering\footnotesize
 \caption{Datasets of the evaluation: the five PLAID datasets of Section~\ref{sec:exp-plaid} and AhmedML; dimension, median nodes per query mesh, whether the meshes share one
   topology, output fields and output scalars.}
 \label{tab:datasets}
 \setlength{\tabcolsep}{4pt}\begin{tabular}{lccccc}\toprule
 Dataset & Dim. & Nodes & Shared & Fields & Scalars\\\midrule
 2D\_profile & 2 & \plaid{2D_profile}{64}{knodes}k & no & 4 & 0\\
 2D\_MSHE & 2 & \plaid{2D_MSHE}{64}{knodes}k & no & 7 & 1\\
 Tensile2d & 2 & \plaid{Tensile2d}{64}{knodes}k & no & 5 & 3\\
 VKI-LS59 & 2 & 36k & yes & 2 & 6\\
 Rotor37 & 3 surf. & \plaid{Rotor37}{64}{knodes}k & yes & 3 & 3\\
 AhmedML & 3 surf. & \plaid{AhmedML}{64}{knodes}k & no & 6 & 4\\
 \bottomrule\end{tabular}
\end{table}

\begin{table*}[h]
  \centering\footnotesize
  \caption{\textbf{Context control}, mean over the output fields of \physicl{}'s median NMSE with the correct context,
    with its output fields and with its input scalars permuted among its simulations (in parentheses: the ratio to the
    correct context). For reference at $K = 64$, the context mean, which neither permutation changes, scores
    \plaid{Rotor37}{64}{mean} and \plaid{VKI-LS59}{64}{mean}, and the nearest context, correctly paired,
    \plaid{Rotor37}{64}{nn} and \plaid{VKI-LS59}{64}{nn}.}
  \label{tab:context-control}
  \setlength{\tabcolsep}{3pt}
  \IfFileExists{results/context_control_table.tex}{\begin{tabular}{@{}llcccccc@{}}
\toprule
Dataset & Context & $K = 8$ & $K = 16$ & $K = 32$ & $K = 64$ & $K = 125$ & $K = 250$ \\
\midrule
Rotor37 & correct & 0.0482 & 0.0336 & 0.0261 & 0.0228 & 0.0206 & 0.0193 \\
 & outputs permuted & 0.101 ($\times$2.1) & 0.0836 ($\times$2.5) & 0.0723 ($\times$2.8) & 0.0653 ($\times$2.9) & 0.0587 ($\times$2.9) & 0.0584 ($\times$3.0) \\
 & inputs permuted & 0.14 ($\times$2.9) & 0.089 ($\times$2.6) & 0.0701 ($\times$2.7) & 0.0654 ($\times$2.9) & 0.0613 ($\times$3.0) & 0.058 ($\times$3.0) \\
VKI-LS59 & correct & 0.113 & 0.0794 & 0.067 & 0.0536 & 0.0503 & 0.0452 \\
 & outputs permuted & 0.221 ($\times$2.0) & 0.182 ($\times$2.3) & 0.198 ($\times$2.9) & 0.198 ($\times$3.7) & 0.193 ($\times$3.8) & 0.169 ($\times$3.7) \\
 & inputs permuted & 0.2 ($\times$1.8) & 0.128 ($\times$1.6) & 0.105 ($\times$1.6) & 0.0857 ($\times$1.6) & 0.0851 ($\times$1.7) & 0.0798 ($\times$1.8) \\
\bottomrule
\end{tabular}
}{\todo{run paper/scripts/context\_control.py}}
\end{table*}

\begin{table*}[h]
  \centering\footnotesize
  \caption{\textbf{PLAID datasets}, mean over the output fields of the median NMSE at $K = 16$ and 64: \physicl{}
    ($M = 4$, first draw; $\pm$ half the range over three draws of the context and queries, the baselines on the first), the nearest
    context (its best setting), the best swept baseline and the PCA floor; fit / one query's times in seconds of
    \physicl{} and of that baseline. Bold, on the first draw: the most accurate of the three methods (the floor is an
    oracle), and the faster fit and query.}
  \label{tab:plaid}
  \setlength{\tabcolsep}{3.5pt}
  \IfFileExists{results/plaid_table.tex}{\begin{tabular}{@{}lrcccccc@{}}
\toprule
Dataset & $K$ & \physicl{} & Nearest & Best & Floor & \physicl{} (s) & Best (s) \\
\midrule
2D\_profile & 16 & \textbf{0.116}\,$\pm$\,0.0088 & 0.228 & 0.129 & 0.101 & 15.5 / \textbf{0.95} & \textbf{7.88} / 3.78 \\
2D\_MSHE & 16 & \textbf{0.502}\,$\pm$\,0.019 & 1.23 & 0.628 & 0.505 & 15.5 / \textbf{1.37} & \textbf{11.2} / 3.57 \\
Tensile2d & 16 & 0.00249\,$\pm$\,0.00067 & 0.0308 & \textbf{0.000534} & 0.000122 & 16.1 / 1.18 & \textbf{4.96} / \textbf{0.913} \\
VKI-LS59 & 16 & 0.0794\,$\pm$\,0.083 & 0.1 & \textbf{0.0702} & 0.00917 & 16.3 / 0.999 & \textbf{1.6} / \textbf{0.478} \\
Rotor37 & 16 & 0.0336\,$\pm$\,0.0015 & 0.0383 & \textbf{0.00761} & 0.00282 & 16.2 / \textbf{0.906} & \textbf{4.32} / 2.15 \\
2D\_profile & 64 & \textbf{0.064}\,$\pm$\,0.0065 & 0.144 & 0.0693 & 0.0416 & \textbf{17.5} / \textbf{0.895} & 22.3 / 8.87 \\
2D\_MSHE & 64 & \textbf{0.44}\,$\pm$\,0.018 & 1.16 & 0.48 & 0.319 & \textbf{16.6} / \textbf{0.803} & 63.4 / 15.1 \\
Tensile2d & 64 & 0.000504\,$\pm$\,9e-05 & 0.0193 & \textbf{3.94e-05} & 4.15e-05 & 17.9 / \textbf{0.648} & \textbf{12.3} / 2.73 \\
VKI-LS59 & 64 & 0.0536\,$\pm$\,0.013 & 0.103 & \textbf{0.0337} & 0.00149 & 17.6 / \textbf{0.983} & \textbf{14.5} / 5.91 \\
Rotor37 & 64 & 0.0228\,$\pm$\,0.00082 & 0.0292 & \textbf{0.00331} & 0.000497 & \textbf{18.2} / \textbf{1.28} & 23.4 / 6.46 \\
\bottomrule
\end{tabular}
}{\todo{run paper/scripts/fronts.py}}
\end{table*}

\paragraph{Fine-tuning.} Both fine-tunings start from the pretrained weights (fresh optimizer and schedule), with the
pretraining's model and the proxy and augmented proxy of Appendix~\ref{app:finetune-prior} (12{,}000 perforated cells of one mesh each, RIB 1{,}024
patches, one output field among the dataset's seven, the three components of $C$ as input scalars, the signed
distance and the direction to the nearest wall as input fields), 50{,}000 steps of Muon at learning rate $10^{-4}$
(warmup 500, linear decay over the last 10{,}000) on one H200, and larger spatial budgets: 2{,}048 quadrature nodes,
1{,}024 anchors, 4{,}096 decoded nodes and 4{,}096 boundary nodes. Validation uses the proxy's tasks; the
reported checkpoint, the best on that validation, is also near the end of each run, at step 49{,}000 for (b) and
50{,}000 for (b$'$). The fine-tunings took 13.3 and 10.7 GPU-hours, against 139 for the pretraining. Each of the 12{,}000 perforated cells is
seen about 1{,}300 times in 50{,}000 steps, and the validation tasks draw new loads and laws on perforated cells of the same bank.
The near-hole NMSE takes the nodes within 0.025 of a hole wall in the unit cell, the
cell's sides excluded.

\paragraph{What the two fine-tunings do not separate.} No control run continues the pretraining for as many steps, so the gain
of (b) and (b$'$) is not separated from the extra steps.

\begin{table*}[p]
  \centering\footnotesize
  \caption{\textbf{Fine-tuning on 2D\_MSHE.} Median NMSE per output field, on every node and near the holes, of the
    pretrained model (a) and the two fine-tunings (b, b$'$), all at $M = 4$, against the best swept baseline and the
    PCA floor (a perfect regressor of the PCA scores). Mean over three context draws, $\pm$ half their range.
    $K = 250$ is beyond the training context sizes. Bold: the lowest NMSE of the models and the best baseline.}
  \label{tab:finetune}
  \IfFileExists{results/finetune_table.tex}{\begin{tabular}{llccccc}
\toprule
& $K$ & pretrained & proxy & augmented proxy & best baseline & PCA floor \\
\midrule
\multicolumn{7}{l}{median NMSE, all nodes} \\
P11 & 16 & 0.775\,$\pm$\,0.020 & 0.371\,$\pm$\,0.022 & \textbf{0.314\,$\pm$\,0.044} & 0.866 & 0.656 \\
P11 & 64 & 0.674\,$\pm$\,0.023 & 0.355\,$\pm$\,0.016 & \textbf{0.267\,$\pm$\,0.020} & 0.632 & 0.459 \\
P11 & 250 & 0.585\,$\pm$\,0.006 & 0.349\,$\pm$\,0.003 & \textbf{0.220\,$\pm$\,0.004} & 0.544 & 0.249 \\
P12 & 16 & 0.873\,$\pm$\,0.018 & 0.378\,$\pm$\,0.015 & \textbf{0.337\,$\pm$\,0.007} & 0.811 & 0.685 \\
P12 & 64 & 0.771\,$\pm$\,0.027 & 0.365\,$\pm$\,0.008 & \textbf{0.278\,$\pm$\,0.014} & 0.644 & 0.432 \\
P12 & 250 & 0.690\,$\pm$\,0.016 & 0.358\,$\pm$\,0.005 & \textbf{0.230\,$\pm$\,0.002} & 0.506 & 0.205 \\
P21 & 16 & 0.819\,$\pm$\,0.027 & 0.352\,$\pm$\,0.002 & \textbf{0.310\,$\pm$\,0.049} & 0.766 & 0.656 \\
P21 & 64 & 0.759\,$\pm$\,0.031 & 0.340\,$\pm$\,0.009 & \textbf{0.270\,$\pm$\,0.019} & 0.610 & 0.417 \\
P21 & 250 & 0.679\,$\pm$\,0.003 & 0.340\,$\pm$\,0.002 & \textbf{0.222\,$\pm$\,0.002} & 0.487 & 0.201 \\
P22 & 16 & 0.754\,$\pm$\,0.054 & 0.373\,$\pm$\,0.038 & \textbf{0.300\,$\pm$\,0.014} & 0.829 & 0.645 \\
P22 & 64 & 0.664\,$\pm$\,0.034 & 0.364\,$\pm$\,0.013 & \textbf{0.264\,$\pm$\,0.013} & 0.643 & 0.465 \\
P22 & 250 & 0.585\,$\pm$\,0.016 & 0.353\,$\pm$\,0.010 & \textbf{0.227\,$\pm$\,0.003} & 0.543 & 0.247 \\
psi & 16 & 0.257\,$\pm$\,0.026 & 0.125\,$\pm$\,0.012 & \textbf{0.118\,$\pm$\,0.009} & 0.954 & 0.837 \\
psi & 64 & 0.235\,$\pm$\,0.016 & 0.126\,$\pm$\,0.009 & \textbf{0.106\,$\pm$\,0.009} & 0.779 & 0.448 \\
psi & 250 & 0.205\,$\pm$\,0.007 & 0.127\,$\pm$\,0.003 & \textbf{0.104\,$\pm$\,0.003} & 0.561 & 0.152 \\
u1 & 16 & 0.060\,$\pm$\,0.005 & 0.026\,$\pm$\,0.002 & \textbf{0.020\,$\pm$\,0.002} & 0.053 & 0.027 \\
u1 & 64 & 0.043\,$\pm$\,0.002 & 0.022\,$\pm$\,0.001 & \textbf{0.015\,$\pm$\,0.000} & 0.021 & 0.007 \\
u1 & 250 & 0.034\,$\pm$\,0.001 & 0.022\,$\pm$\,0.001 & 0.014\,$\pm$\,0.001 & \textbf{0.010} & 0.002 \\
u2 & 16 & 0.067\,$\pm$\,0.002 & 0.026\,$\pm$\,0.002 & \textbf{0.020\,$\pm$\,0.002} & 0.066 & 0.032 \\
u2 & 64 & 0.048\,$\pm$\,0.002 & 0.024\,$\pm$\,0.001 & \textbf{0.016\,$\pm$\,0.001} & 0.023 & 0.007 \\
u2 & 250 & 0.036\,$\pm$\,0.001 & 0.023\,$\pm$\,0.000 & 0.014\,$\pm$\,0.001 & \textbf{0.010} & 0.002 \\
\midrule
\multicolumn{7}{l}{median NMSE near the holes} \\
P11 & 16 & 1.092\,$\pm$\,0.024 & 0.453\,$\pm$\,0.008 & \textbf{0.291\,$\pm$\,0.022} & -- & -- \\
P11 & 64 & 1.014\,$\pm$\,0.028 & 0.440\,$\pm$\,0.009 & \textbf{0.281\,$\pm$\,0.010} & -- & -- \\
P11 & 250 & 0.908\,$\pm$\,0.009 & 0.430\,$\pm$\,0.001 & \textbf{0.264\,$\pm$\,0.003} & -- & -- \\
P12 & 16 & 1.202\,$\pm$\,0.020 & 0.344\,$\pm$\,0.009 & \textbf{0.314\,$\pm$\,0.003} & -- & -- \\
P12 & 64 & 1.154\,$\pm$\,0.031 & 0.349\,$\pm$\,0.007 & \textbf{0.302\,$\pm$\,0.007} & -- & -- \\
P12 & 250 & 1.057\,$\pm$\,0.001 & 0.351\,$\pm$\,0.003 & \textbf{0.294\,$\pm$\,0.001} & -- & -- \\
P21 & 16 & 1.207\,$\pm$\,0.019 & 0.350\,$\pm$\,0.013 & \textbf{0.291\,$\pm$\,0.017} & -- & -- \\
P21 & 64 & 1.182\,$\pm$\,0.017 & 0.344\,$\pm$\,0.010 & \textbf{0.288\,$\pm$\,0.003} & -- & -- \\
P21 & 250 & 1.055\,$\pm$\,0.009 & 0.340\,$\pm$\,0.002 & \textbf{0.278\,$\pm$\,0.006} & -- & -- \\
P22 & 16 & 1.065\,$\pm$\,0.012 & 0.486\,$\pm$\,0.009 & \textbf{0.321\,$\pm$\,0.016} & -- & -- \\
P22 & 64 & 0.986\,$\pm$\,0.022 & 0.463\,$\pm$\,0.011 & \textbf{0.315\,$\pm$\,0.009} & -- & -- \\
P22 & 250 & 0.890\,$\pm$\,0.007 & 0.455\,$\pm$\,0.008 & \textbf{0.299\,$\pm$\,0.001} & -- & -- \\
psi & 16 & 0.907\,$\pm$\,0.037 & \textbf{0.449\,$\pm$\,0.013} & 0.464\,$\pm$\,0.024 & -- & -- \\
psi & 64 & 0.776\,$\pm$\,0.033 & 0.434\,$\pm$\,0.008 & \textbf{0.419\,$\pm$\,0.015} & -- & -- \\
psi & 250 & 0.647\,$\pm$\,0.017 & 0.424\,$\pm$\,0.007 & \textbf{0.406\,$\pm$\,0.008} & -- & -- \\
u1 & 16 & 0.253\,$\pm$\,0.004 & 0.064\,$\pm$\,0.003 & \textbf{0.045\,$\pm$\,0.002} & -- & -- \\
u1 & 64 & 0.211\,$\pm$\,0.005 & 0.055\,$\pm$\,0.001 & \textbf{0.036\,$\pm$\,0.002} & -- & -- \\
u1 & 250 & 0.179\,$\pm$\,0.003 & 0.053\,$\pm$\,0.000 & \textbf{0.033\,$\pm$\,0.001} & -- & -- \\
u2 & 16 & 0.277\,$\pm$\,0.012 & 0.070\,$\pm$\,0.005 & \textbf{0.050\,$\pm$\,0.005} & -- & -- \\
u2 & 64 & 0.232\,$\pm$\,0.005 & 0.063\,$\pm$\,0.003 & \textbf{0.039\,$\pm$\,0.002} & -- & -- \\
u2 & 250 & 0.192\,$\pm$\,0.003 & 0.061\,$\pm$\,0.002 & \textbf{0.036\,$\pm$\,0.001} & -- & -- \\
\bottomrule
\end{tabular}
}{\todo{run paper/scripts/finetune.py}}
\end{table*}
\begin{figure*}[p]
  \centering
  \IfFileExists{figures/finetune_k.pdf}{\includegraphics[width=\linewidth]{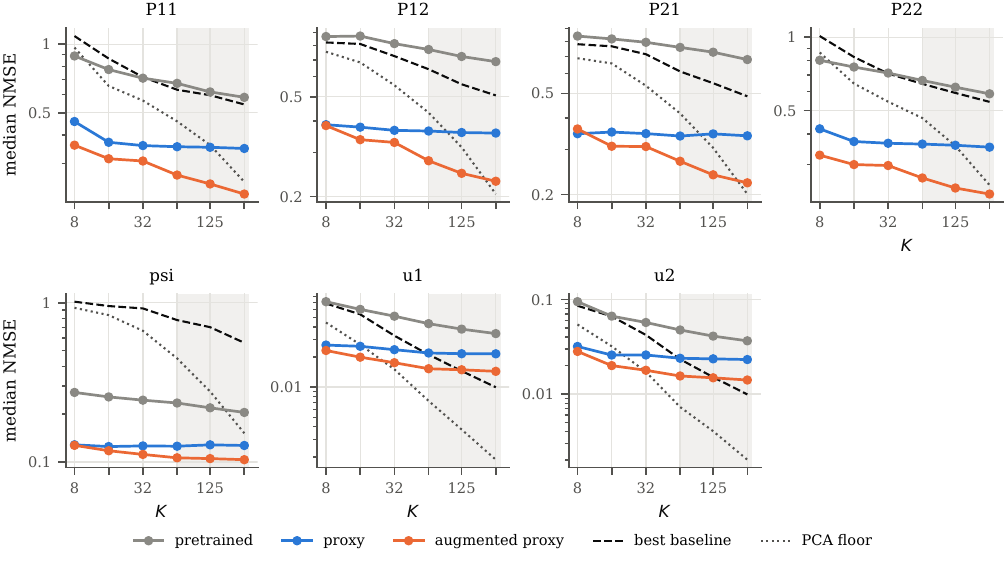}}{\todo{figure}}
  \caption{\textbf{Fine-tuning against $K$.} Median NMSE per output field of the pretrained model and the two
    fine-tunings, the best swept baseline and the PCA floor; shaded, $K > 64$, beyond the training context sizes.}
  \label{fig:finetune-k-fields}
\end{figure*}

\section{Training curves}
\label{app:training-runs}

\begin{figure*}[h]
  \centering
  \IfFileExists{figures/training_runs.pdf}{\includegraphics[width=\linewidth]{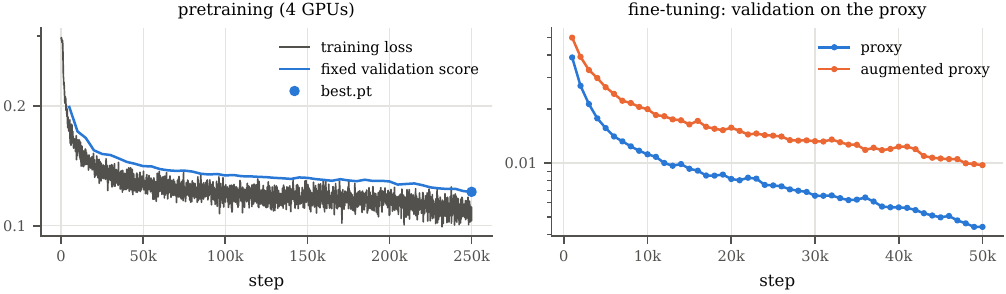}}{\todo{run
    paper/scripts/export_mlflow.py}}
  \caption{\textbf{Training curves.} Left: the pretraining's training loss and fixed validation score
    (tasks of the procedural prior; the best-validation checkpoint at the last step). Right: the fine-tunings' validation score on the proxy's
    tasks, whose geometries come from the training bank; it ranks the two fine-tunings the other way from 2D\_MSHE.}
  \label{fig:training-runs}
\end{figure*}

Figure~\ref{fig:training-runs} shows the training of the reported checkpoint and of its two fine-tunings; their
costs are given in Section~\ref{sec:exp-plaid} and Section~\ref{sec:exp-finetune}.

\section{Per-dataset results}
\label{app:per-dataset}

Table~\ref{tab:fields} gives the median NMSE of every output field at $K = 64$, and Table~\ref{tab:scalars} the
scalar outputs. Figures~\ref{fig:fields-2D_profile}--\ref{fig:fields-AhmedML} show the first query of each dataset
at $K = 64$.

\newcommand{\perdatasetfields}[3]{%
  \begin{figure*}[p]
    \centering
    \IfFileExists{figures/fields/#1.jpg}{\includegraphics[width=\linewidth, height=0.8\textheight, keepaspectratio]{figures/fields/#1.jpg}}{\todo{run paper/scripts/field\_figures.py on #2}}
    \caption{\textbf{#2}, first query at $K = 64$#3.}
    \label{fig:fields-#1}
  \end{figure*}}

\begin{table*}[t]
  \centering\footnotesize
  \caption{\textbf{Field outputs}: median NMSE at $K = 64$ of \physicl{} ($M = 4$) and of the best setting of each
    baseline family, selected on the queries, with the PCA floor (an oracle); first context draw; Table~\ref{tab:finetune} gives the mean over three draws for
    2D\_MSHE. Bold: the best of the row.}
  \label{tab:fields}
  \setlength{\tabcolsep}{3pt}
  \IfFileExists{results/fields_table.tex}{\begin{tabular}{@{}llccccc@{}}
\toprule
Dataset & Field output & PhysICL & probes/PCA + GP & probes/PCA + Kumo & nearest context & PCA floor \\
\midrule
2D\_profile & Mach & \textbf{0.069} & 0.252 & 0.0824 & 0.166 & 0.0499 \\
2D\_profile & Pressure & \textbf{0.0573} & 0.198 & 0.0603 & 0.133 & 0.0393 \\
2D\_profile & Velocity-x & \textbf{0.0705} & 0.236 & 0.074 & 0.162 & 0.0467 \\
2D\_profile & Velocity-y & 0.059 & 0.178 & \textbf{0.0569} & 0.116 & 0.0304 \\
2D\_MSHE & P11 & 0.643 & 0.773 & \textbf{0.632} & 1.52 & 0.459 \\
2D\_MSHE & P12 & 0.759 & 0.787 & \textbf{0.644} & 1.33 & 0.432 \\
2D\_MSHE & P21 & 0.723 & 0.757 & \textbf{0.61} & 1.3 & 0.417 \\
2D\_MSHE & P22 & \textbf{0.639} & 0.791 & 0.643 & 1.54 & 0.465 \\
2D\_MSHE & psi & \textbf{0.223} & 0.925 & 0.779 & 2.18 & 0.448 \\
2D\_MSHE & u1 & 0.0437 & 0.0735 & \textbf{0.0211} & 0.107 & 0.00727 \\
2D\_MSHE & u2 & 0.0477 & 0.104 & \textbf{0.0233} & 0.137 & 0.00728 \\
Tensile2d & U1 & 0.000637 & 0.000157 & \textbf{2.36e-05} & 0.0395 & 1.87e-05 \\
Tensile2d & U2 & 0.000221 & 2.44e-05 & \textbf{2.06e-05} & 0.0276 & 1.47e-05 \\
Tensile2d & sig11 & 0.000464 & 0.000302 & \textbf{3.76e-05} & 0.00346 & 4.02e-05 \\
Tensile2d & sig12 & 0.000645 & 0.000395 & \textbf{6.84e-05} & 0.00567 & 7.9e-05 \\
Tensile2d & sig22 & 0.000553 & 0.00027 & \textbf{4.7e-05} & 0.0123 & 5.51e-05 \\
VKI-LS59 & mach & 0.0187 & 0.028 & \textbf{0.0112} & 0.0475 & 0.00106 \\
VKI-LS59 & nut & 0.0886 & 0.173 & \textbf{0.0546} & 0.157 & 0.00192 \\
Rotor37 & Density & 0.0295 & 0.00557 & \textbf{0.00384} & 0.0188 & 0.000433 \\
Rotor37 & Pressure & 0.032 & 0.00565 & \textbf{0.004} & 0.0193 & 0.000457 \\
Rotor37 & Temperature & 0.00689 & 0.00225 & \textbf{0.00208} & 0.0496 & 0.000601 \\
AhmedML & pMean & \textbf{0.0548} & 0.693 & 0.0652 & 0.198 & 0.0226 \\
AhmedML & static(p)\_coeffMean & \textbf{0.0562} & 0.693 & 0.0652 & 0.198 & 0.0226 \\
AhmedML & wallShearStressMean\_0 & \textbf{0.0679} & 0.538 & 0.0889 & 0.225 & 0.043 \\
AhmedML & wallShearStressMean\_1 & \textbf{0.0403} & 0.593 & 0.06 & 0.166 & 0.0315 \\
AhmedML & wallShearStressMean\_2 & \textbf{0.0405} & 0.579 & 0.063 & 0.159 & 0.033 \\
AhmedML & yPlusMean & \textbf{0.0658} & 0.388 & 0.0927 & 0.226 & 0.0581 \\
\bottomrule
\end{tabular}
}{\todo{run paper/scripts/output\_tables.py}}
\end{table*}
\begin{table*}[t]
  \centering\footnotesize
  \caption{\textbf{Scalar outputs}: $R^2$ at $K = 64$ of \physicl{} (from one partition, $M = 1$) and of the best
    setting of each baseline family, selected on the queries. Bold: the best of the row.}
  \label{tab:scalars}
  \setlength{\tabcolsep}{3pt}
  \IfFileExists{results/scalars_table.tex}{\begin{tabular}{@{}llcccc@{}}
\toprule
Dataset & Scalar output & PhysICL & probes/PCA + GP & probes/PCA + Kumo & nearest context \\
\midrule
2D\_MSHE & effective\_energy & 0.433 & 0.854 & \textbf{0.922} & -1.26 \\
Tensile2d & max\_U2\_top & 0.943 & 0.999 & \textbf{1} & 0.848 \\
Tensile2d & max\_sig22\_top & \textbf{0.999} & \textbf{0.999} & \textbf{0.999} & 0.477 \\
Tensile2d & max\_von\_mises & \textbf{0.992} & 0.991 & \textbf{0.992} & 0.931 \\
VKI-LS59 & Pr & 0.637 & 0.871 & \textbf{0.924} & 0.556 \\
VKI-LS59 & Q & 0.996 & \textbf{0.999} & \textbf{0.999} & 0.71 \\
VKI-LS59 & Tr & 0.943 & 0.98 & \textbf{0.993} & 0.709 \\
VKI-LS59 & angle\_out & 0.981 & 0.992 & \textbf{0.994} & 0.675 \\
VKI-LS59 & eth\_is & 0.805 & 0.866 & \textbf{0.945} & 0.618 \\
VKI-LS59 & power & 0.833 & 0.966 & \textbf{0.98} & 0.576 \\
Rotor37 & Compression\_ratio & 0.962 & \textbf{0.988} & 0.987 & 0.069 \\
Rotor37 & Efficiency & 0.667 & 0.813 & \textbf{0.833} & -0.255 \\
Rotor37 & Massflow & 0.952 & \textbf{0.987} & 0.986 & 0.0266 \\
AhmedML & cd & \textbf{0.813} & -0.0371 & 0.767 & 0.336 \\
AhmedML & cd\_varref & \textbf{0.579} & 0.505 & 0.544 & -0.0567 \\
AhmedML & cl & \textbf{0.863} & -0.0485 & 0.859 & -0.432 \\
AhmedML & cl\_varref & \textbf{0.868} & -0.0314 & 0.852 & -0.388 \\
\bottomrule
\end{tabular}
}{\todo{run paper/scripts/output\_tables.py}}
\end{table*}

\perdatasetfields{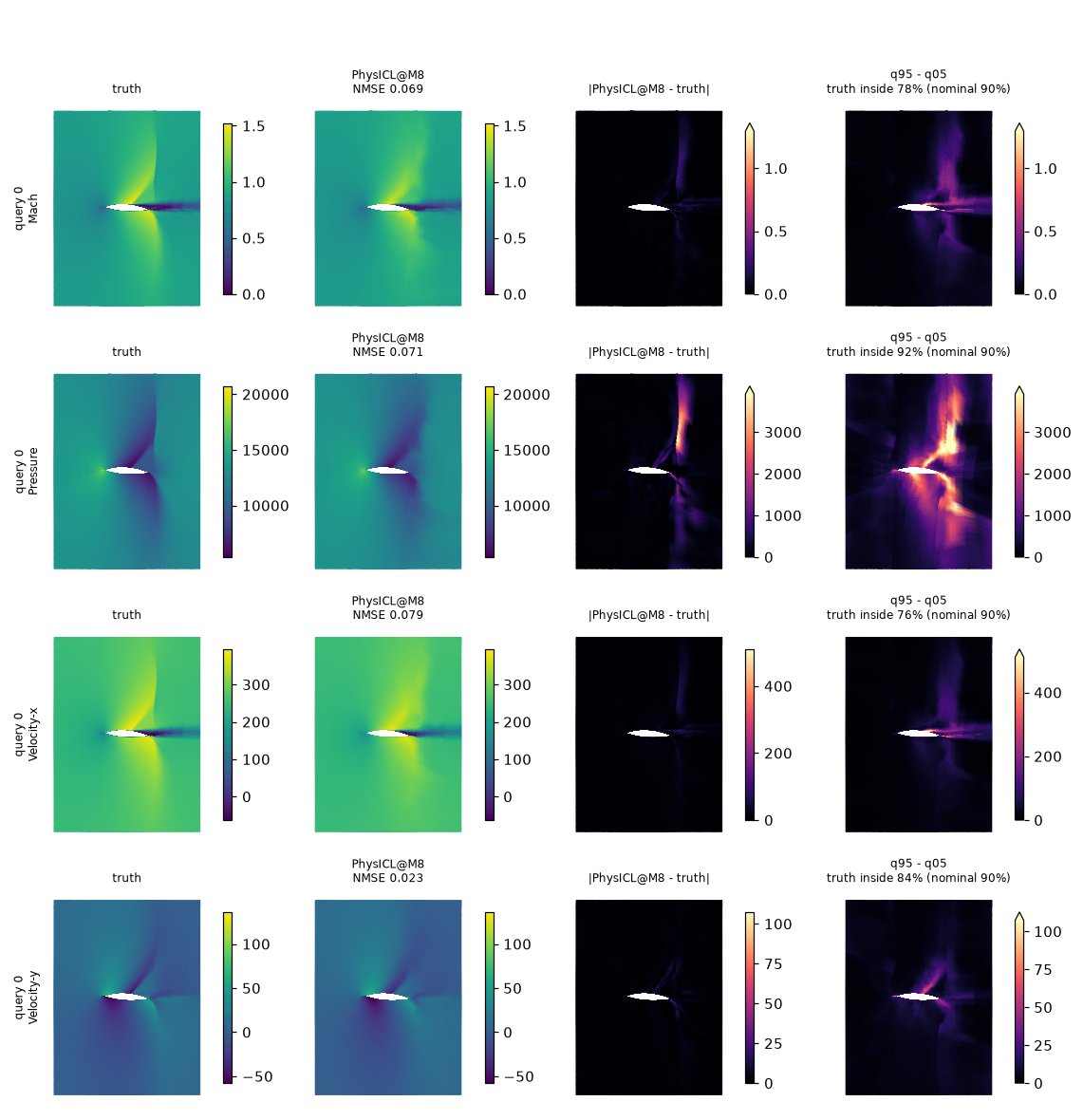}{2D\_profile}{}
\perdatasetfields{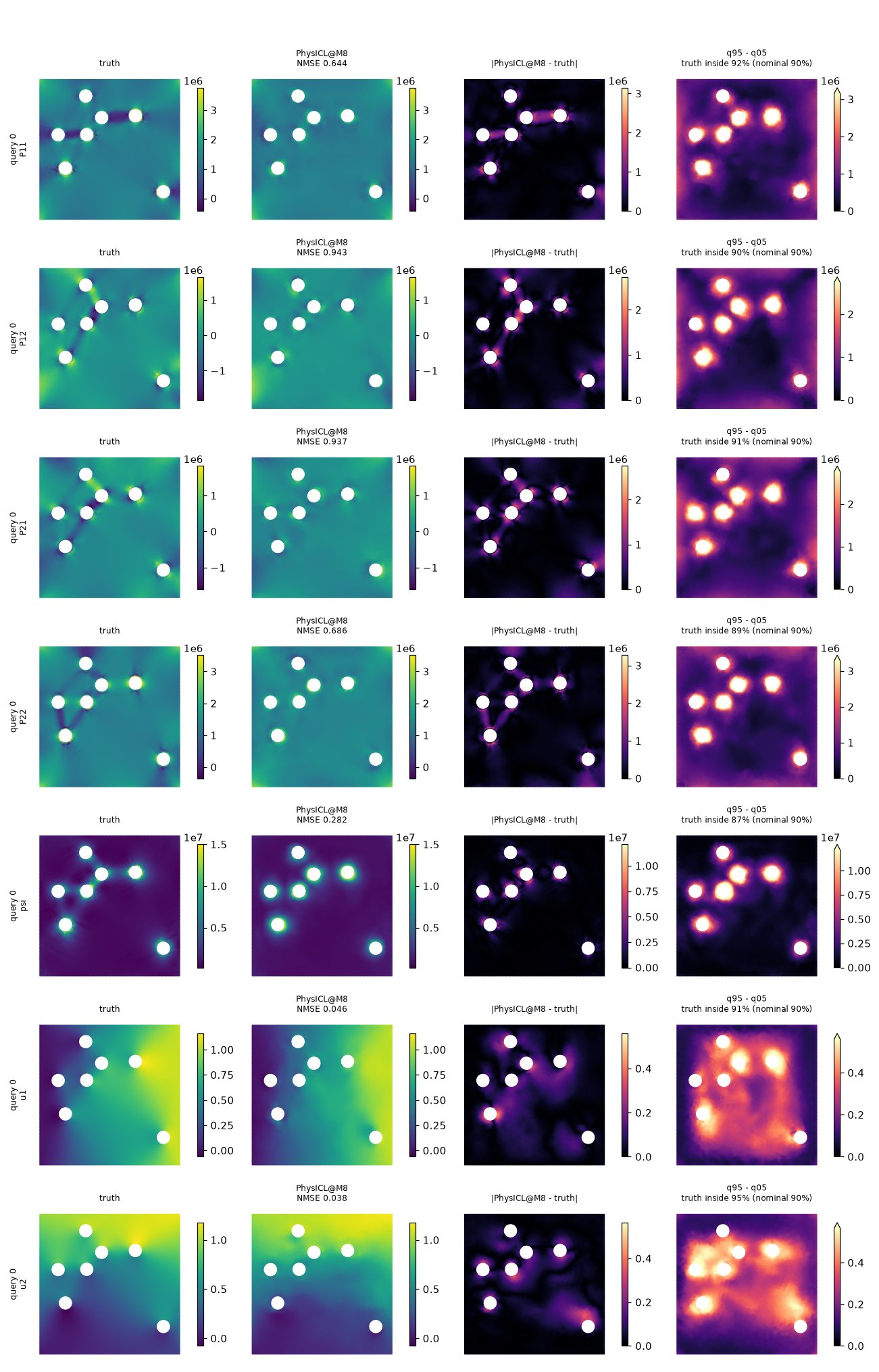}{2D\_MSHE}{, pretrained model}
\perdatasetfields{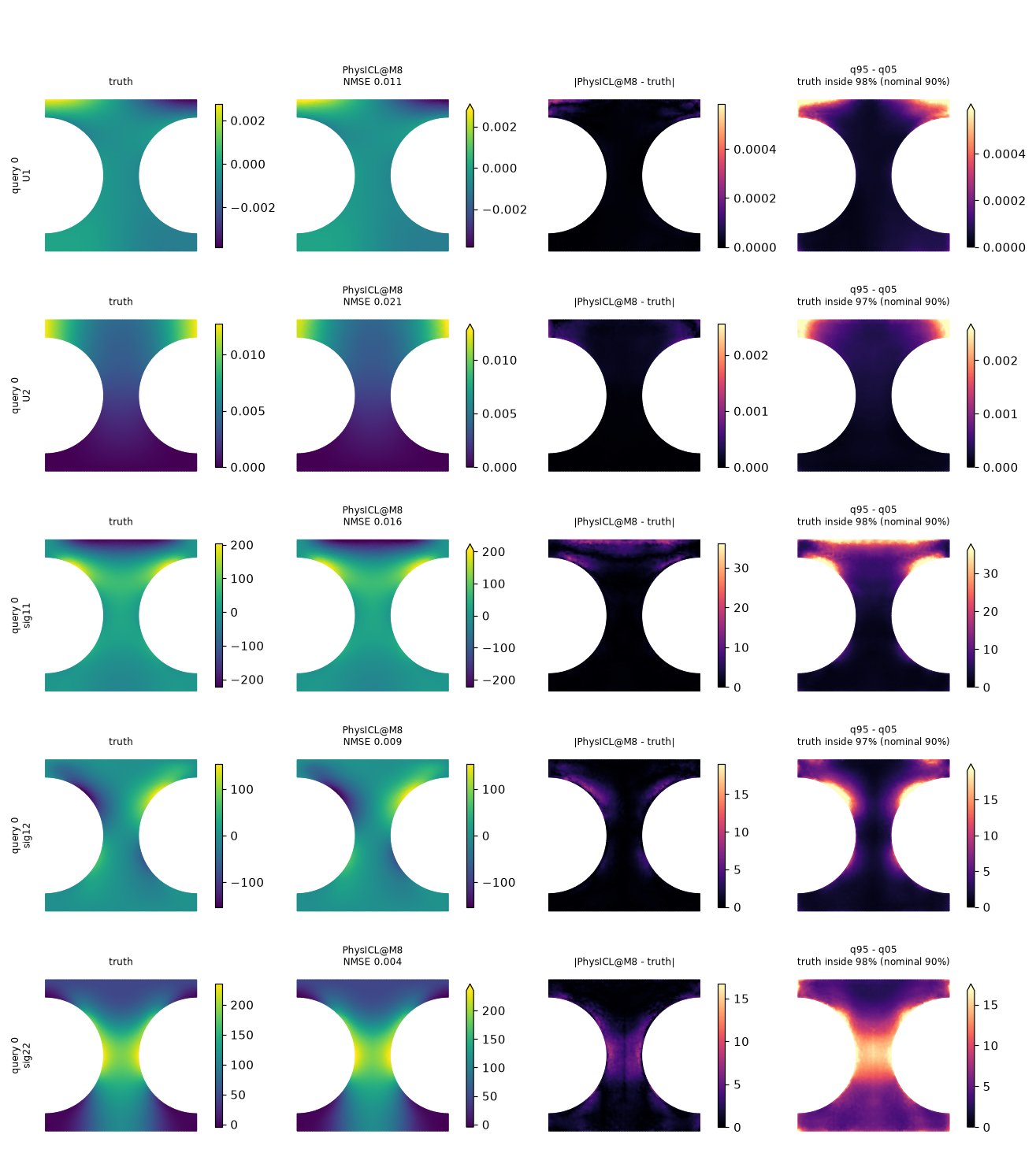}{Tensile2d}{}
\perdatasetfields{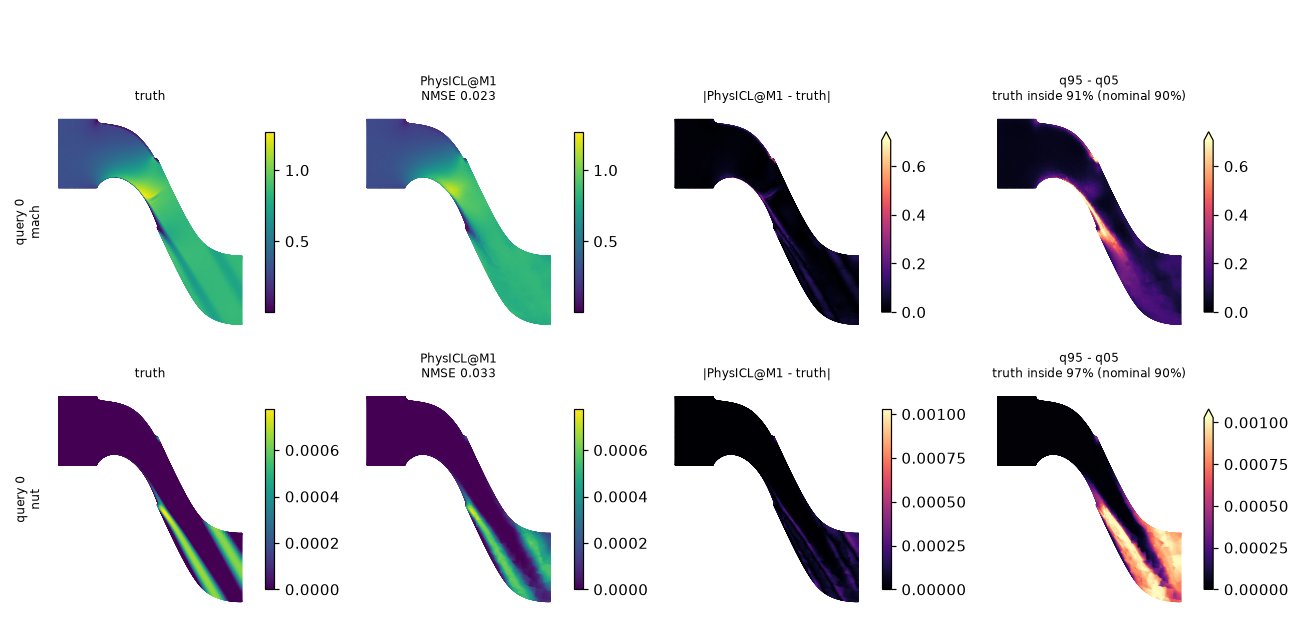}{VKI-LS59}{}
\perdatasetfields{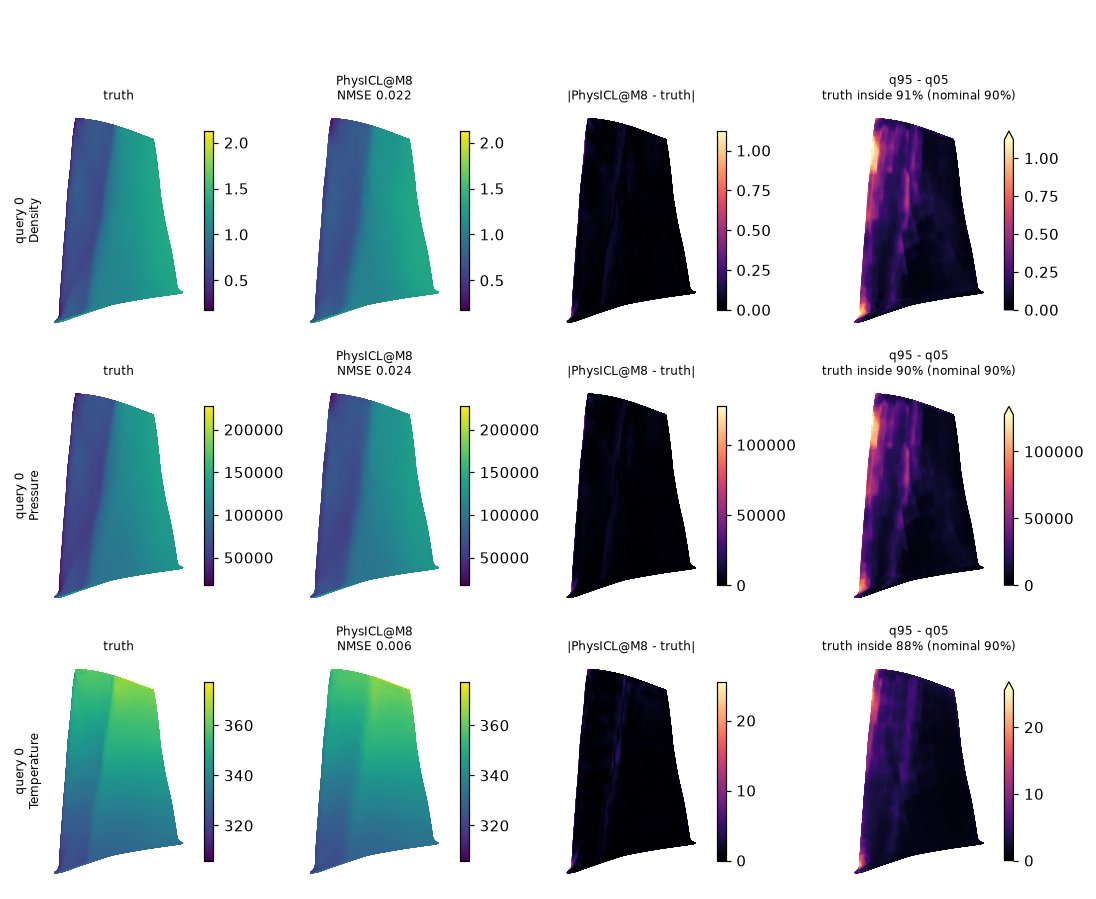}{Rotor37}{}
\perdatasetfields{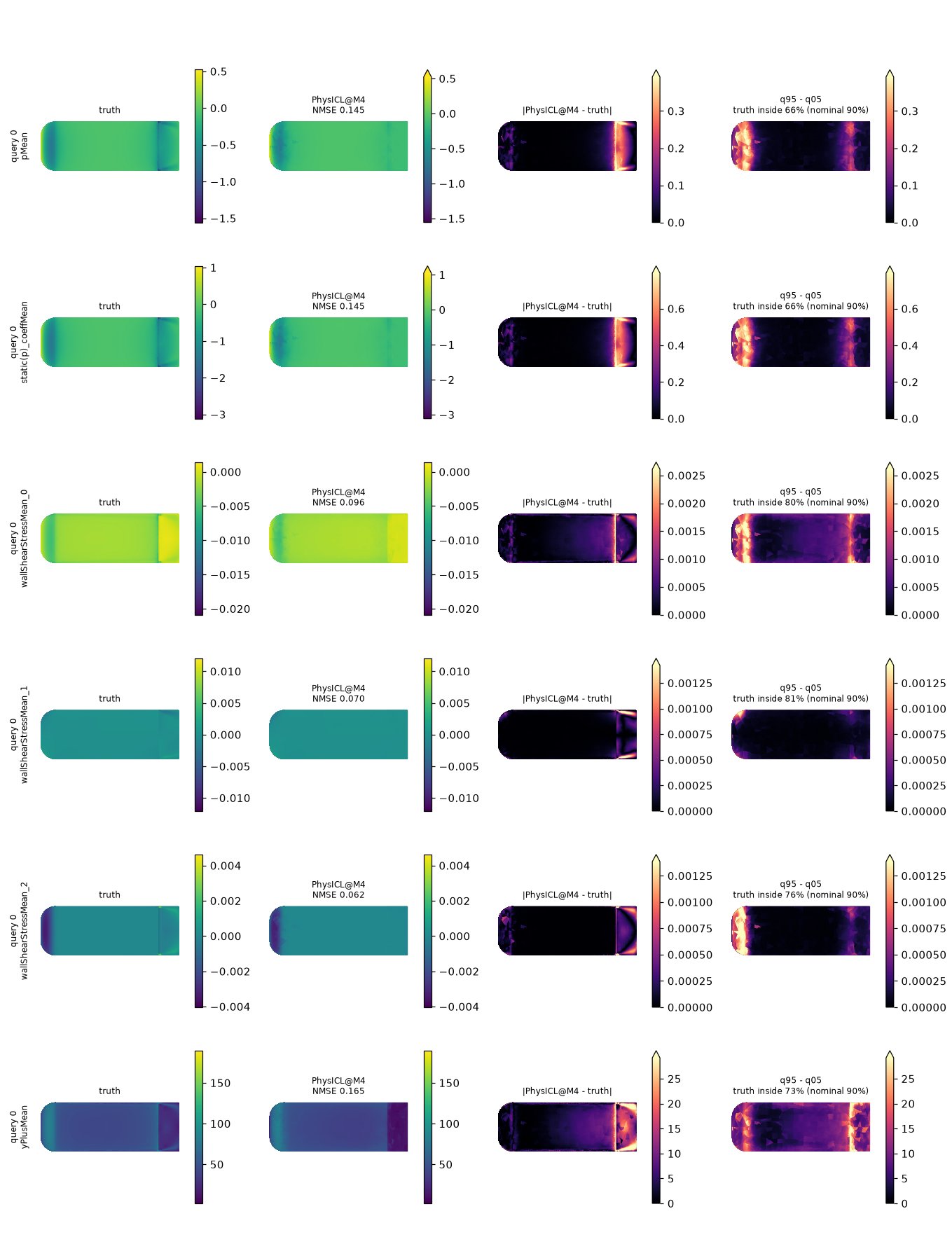}{AhmedML}{}
\begin{figure*}[p]
  \centering
  \IfFileExists{figures/fields/AhmedML_visu_PhysICL_2.png}{\includegraphics[width=\linewidth]{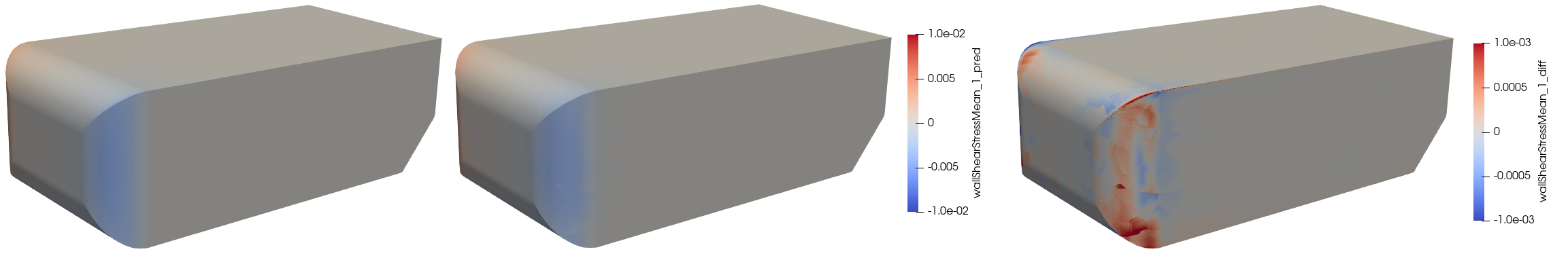}}{\todo{AhmedML 3D view}}
  \caption{\textbf{AhmedML}, second component of the wall shear stress on the front of one query's body, outside its
    context ($K = 64$, $M = 8$, point prediction): the truth (left), \physicl{} (middle), both on one color scale, and
    the prediction minus the truth (right), on a scale ten times finer. The prediction places the band of negative shear
    behind the rounded nose and the peak along its leading edge; the error concentrates on the rounded nose and its edges, where the field varies fastest, and nearly vanishes on the
    flat faces.}
  \label{fig:ahmed-3d}
\end{figure*}
\begin{figure*}[p]
  \centering
  \IfFileExists{figures/fields/2D_MSHE_arms.jpg}{\includegraphics[width=\linewidth, height=0.85\textheight, keepaspectratio]{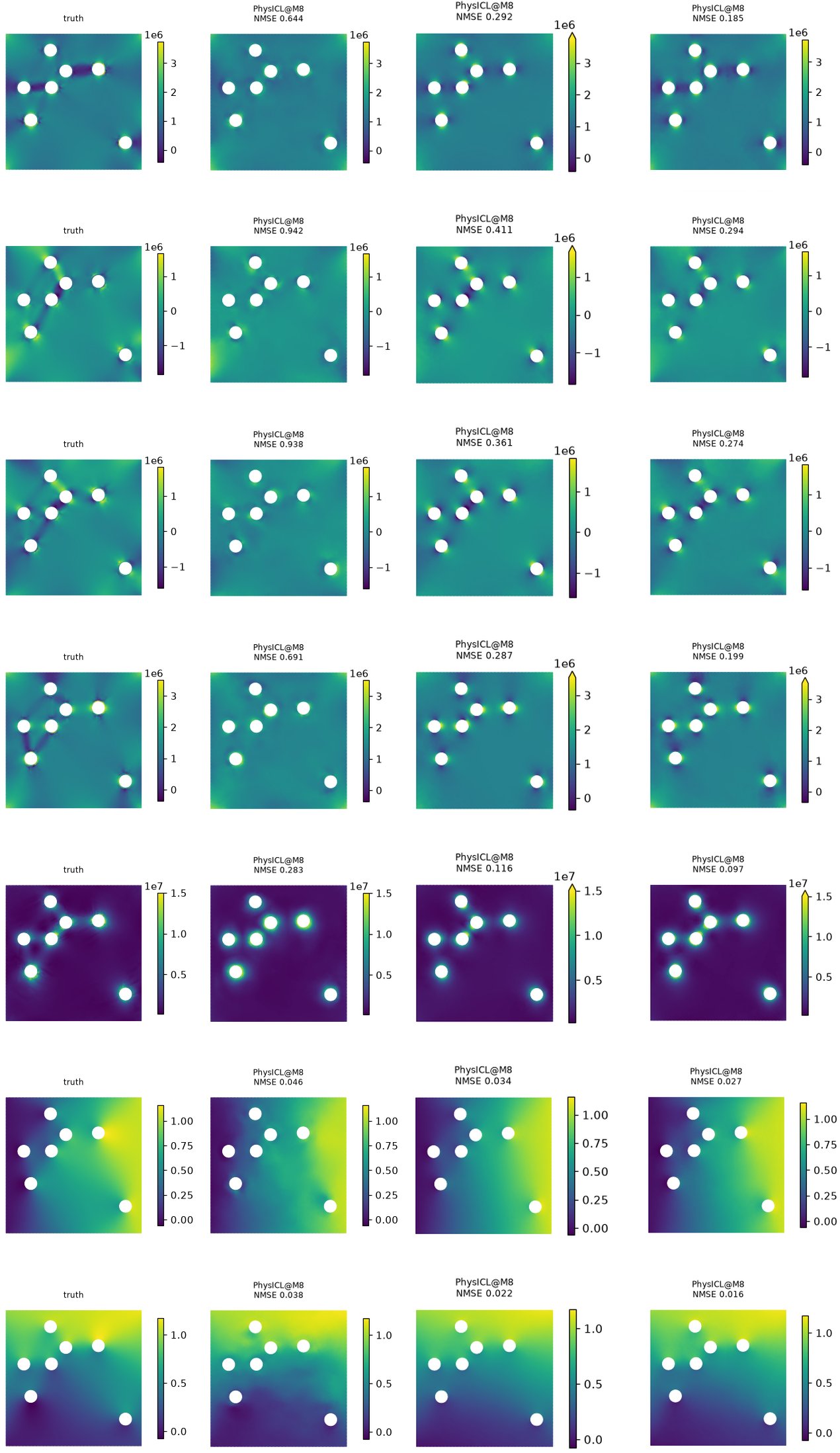}}{\todo{run paper/scripts/field\_figures.py --compare}}
  \caption{\textbf{2D\_MSHE}, every output field of the first query at $K = 64$ (rows: $P_{11}$, $P_{12}$, $P_{21}$,
    $P_{22}$, $\psi$, $u_1$, $u_2$): the truth, then \physicl{} pretrained (a), fine-tuned on the proxy alone (b) and
    on the augmented proxy (b$'$), all at $M = 8$. The pretrained model
    predicts the stresses flat around the holes; both fine-tunings place their peaks there.}
  \label{fig:fields-2D_MSHE-arms}
\end{figure*}

\clearpage  %

\section{Hidden free-stream direction on the pressure coefficient}
\label{app:heading-cp}

\begin{figure*}[t]
  \centering
  \IfFileExists{figures/heading_cp_pilot.pdf}{\includegraphics[width=\linewidth]{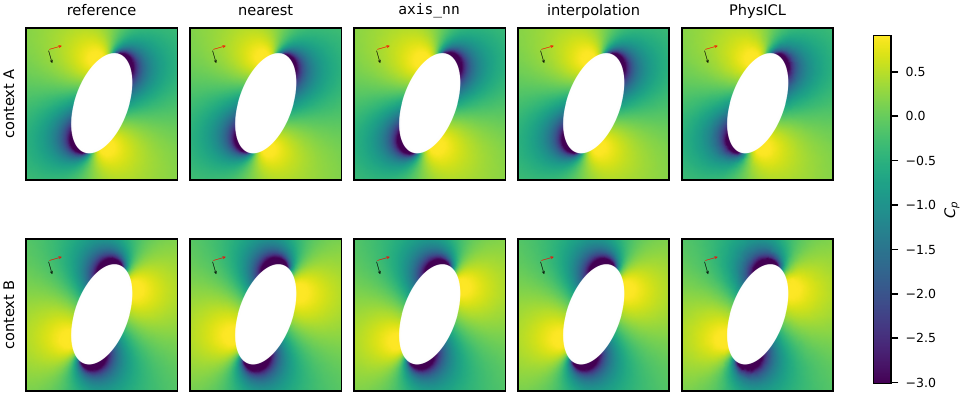}}{\todo{run
  paper/scripts/heading\_swap.py}}
  \caption{\textbf{Hidden free-stream direction on $C_p$}, an incidence episode of the pilot (a template outside the eight of
  Table~\ref{tab:heading-cp}, $K = 16$, \physicl{} at $M = 1$), one query: the reference and each method's prediction with
  the contexts of laws A (top) and B (bottom), one $C_p$ scale. Arrows: the hidden free-stream directions of
  A (black) and B (red).}
  \label{fig:heading-cp}
\end{figure*}

\paragraph{Protocol.} Each episode puts one elliptic body in a box, a \texttt{shapes} exterior domain, under the
inviscid flow model at Mach 0.3, with the pressure coefficient $C_p$ as output and the incidence $\alpha$, over
$-5^\circ$ to $15^\circ$, as the only input. The context simulations lie on both sides of a central interval of
incidence that holds the $Q = 4$ queries. Laws A and B differ only in the free-stream direction, turned by $90^\circ$
and given by no input; every other draw and model input is the same. Eight templates, with two episodes each, were
fixed before any prediction; \physicl{} runs at $M = 1$. Each query has two truths, one per
law: the prediction from the context of law A is scored against truth A, and that from law B against truth B, by
the squared error over the variance of the two truths together; each template gives the median of these errors
per method.
\texttt{axis\_nn} copies the context simulation nearest in $\alpha$, the interpolation blends linearly the two that
bracket the query in $\alpha$, and ``nearest'' in Figure~\ref{fig:heading-cp} is the nearest-context baseline of
Section~\ref{sec:exp-baselines}.

\paragraph{Results.} The scores of Table~\ref{tab:heading-cp} are on $C_p$ (one query in
Figure~\ref{fig:heading-cp}). Every query of every method is associated with the right truth both ways: the
prediction from context A is closer to truth A than the prediction from context B, and conversely. \physicl{} beats
\texttt{axis\_nn} on most templates, irregularly: by an order of magnitude on some, below it on others, and on fewer
templates at $K = 16$ than at $K = 8$. The interpolation between the two bracketing context simulations is more accurate than
\physicl{} on every template, at both $K$. It is a reference specific to this protocol, not a generic method: it needs
a single scalar input and context simulations on both sides of each query, and has no counterpart when several inputs or the
geometry vary.

\begin{table}[t]
  \centering
  \caption{\textbf{Hidden free-stream direction on $C_p$}, per template: the ratios of the median right-context errors over its
  episodes, queries and laws (reference over \physicl{}: above 1, in bold, \physicl{} is the more accurate). Pockets: the episodes with a supersonic wall pocket.}
  \label{tab:heading-cp}
  \small
  \IfFileExists{results/heading_cp_table.tex}{\resizebox{\linewidth}{!}{\begin{tabular}{lccccccc}
\toprule
& & & \multicolumn{2}{c}{\texttt{axis\_nn} / \physicl{}} & \multicolumn{2}{c}{interpolation / \physicl{}} \\
\cmidrule(lr){4-5} \cmidrule(lr){6-7}
template & aspect & pockets & $K = 8$ & $K = 16$ & $K = 8$ & $K = 16$ \\
\midrule
165 & 1.92 & 2/2 & \textbf{2.32} & \textbf{2.08} & 0.24 & 0.24 \\
213 & 1.51 & 0/2 & \textbf{2.07} & \textbf{1.55} & 0.15 & 0.18 \\
233 & 1.55 & 0/2 & \textbf{5.12} & \textbf{3.92} & 0.18 & 0.24 \\
315 & 1.99 & 1/2 & \textbf{1.85} & 0.64 & 0.16 & 0.15 \\
329 & 1.71 & 0/2 & 0.67 & 0.90 & 0.20 & 0.29 \\
335 & 1.57 & 0/2 & \textbf{10.0} & \textbf{6.95} & 0.22 & 0.19 \\
351 & 1.56 & 0/2 & \textbf{13.9} & \textbf{15.5} & 0.36 & 0.43 \\
379 & 1.93 & 1/2 & 0.66 & 0.88 & 0.14 & 0.18 \\
\bottomrule
\end{tabular}
}}{\todo{run
  paper/scripts/heading\_swap.py}}
\end{table}

\end{document}